%% file: main.tex
\DocumentMetadata{ 
	pdfstandard = a-2b,
	pdfversion  = 1.7,
	lang		= en-US,
}
\documentclass[twoside]{mitthesis} %,fontset=libertine, fontset=newtx-sans-text, fontset=heros-stix2, fontset=stix2
\usepackage{listings}%   documentation is here https://ctan.org/pkg/listings

\usepackage[version=4]{mhchem}%   documentation is here https://ctan.org/pkg/mhchem

\usepackage{lipsum}
\IfPackageAtLeastTF{lipsum}{2021/09/20}{\setlipsum{auto-lang=false}}{}

\usepackage[style=ieee,maxbibnames=10,sorting=none]{biblatex}% style=ext-numeric-comp,articlein=false,giveninits=true
	\DefineBibliographyStrings{english}{url= \textsc{url} ,  }% replaces default "[Online]. Available" by "URL"

\usepackage{adjustbox}
\usepackage{algorithm}
\usepackage[noend]{algpseudocode}
\usepackage{booktabs}% better quality tables, https://ctan.org/pkg/booktabs
\usepackage{array}%    additional options for table columns, https://ctan.org/pkg/array

\hypersetup{%
	pdfsubject={Template for writing MIT theses with the mitthesis class},
	pdfkeywords={Massachusetts Institute of Technology, MIT},
	pdfurl={},
	pdfcontactemail={},
	pdfauthortitle={},
}

\begin{document}

%%% edit the following commands to match your thesis %%%%%%%%%%

\title{Computational Fabrication and Assembly for \textit{In Situ} Manufacturing}

% \Author{Martin Nisser}{EECS}[BEng Edinburgh][MSc ETH Zurich]
% Note that third, fourth, fifth, and sixth arguments are optional [] and may be omitted

% note on names: most of the following names are made up; Silas Holman was a physics professor at MIT in the 19th century.

\Author{Martin Eric William Nisser}{Department of Electrical Engineering and Computer Science}[BEng University of Edinburgh (2013)][MSc ETH Zurich (2015)][S.M. Massachusetts Institute of Technology (2019)]
% \Author{Luisa Hernández}{Department of Research}[B.S. Mechanical Engineering, UCLA, 2018][M.S. Stellar Interiors, Vulcan Science Academy, 2020]
% \Author{Thurston Howell III}{Department of Economics}[MBA, Ferengi School of Management, 2022]

% Use once for each degree fulfilled by thesis
% For two degrees from one department, leave the department argument blank for the second degree {}.
% \Degree{Bachelor of Science in Physics}{Department of Physics}
% \Degree{Master of Science in Physics}{}
\Degree{Doctor of Philosophy}{Department of Electrical Engineering and Computer Science}

% If there is more than one supervisor, use the \Supervisor command for each.
\Supervisor{Stefanie Mueller}{Associate Professor of Electrical Engineering and Computer Science}
% \Supervisor{Secunda Castor}{Professor of Research}
% \Supervisor{Quintus Castor}{Professor of Log Dams}

% Professor who formally accepts theses for your department (e.g., the Graduate Officer, Professor Sméagol,...)
% If more than one department, use more than once
% **If you need to reduce vertical space, put the acceptor title in the second argument and leave the third blank {}.**
 \Acceptor{Leslie A. Kolodziejski}{Professor of Electrical Engineering and Computer Science}{Chair, Department Committee on Graduate Students}
% \Acceptor{Tertius Castor}{Professor of Log Dams}{Graduate Officer, Department of Research}
% \Acceptor{Quarta Castor}{Professor of Lodge Building}{Graduate Officer, Department of Mechanical Engineering}

% Usage: \DegreeDate{Month}{year}
% Valid degree months are September, February, or June
\DegreeDate{May}{2024}

% Date that final thesis is submitted to department
\ThesisDate{May 17, 2024}

%%%%%%  Choose whether to have a CREATIVE COMMONS License  %%%%%%%%%%%%%%%%%%%%%%%%%%%%%%%%%%%%%%
%
% If you are using a cc license, put details of your cc license here. 
% Omit this command if you are not using a cc license.
%
\CClicense{CC BY-NC-ND 4.0}{https://creativecommons.org/licenses/by-nc-nd/4.0/}
%

%%%%%%%  Solutions for overflowing titlepage  %%%%%%%%%%%%%%%%%%%%%%%%%%%%%%%%%%%%%%%%%%%%%%%%%%%

% If your title page is overflowing (from too many names, degrees, etc.):
%
% (a) you can reduce the 12pt and 18pt skips between various blocks to 6pt with this command:
%
% \Tighten
%
% (b)  you can scale down the Signature block at the bottom with this command:
%
% \SignatureBlockSize{\small}  %or this one \SignatureBlockSize{\footnotesize}
%
% (c) you can put the acceptor name and title onto two lines, rather than three like this:
%
% \Acceptor{Tertius Castor}{Professor and Graduate Officer, Department of Research}{}
% \Acceptor{Quarta Castor}{Professor and Graduate Officer, Department of Mechanical Engineering}{}
%
% (d) you can change the font size of the the author name[s] with
%
%	\AuthorNameSize{\normalsize}
%
% (e) and you can omit any previous degrees from the title page, instead mentioning them in the Biosketch

% Also, if you prefer to keep the text toward the top of the page with most white space at the bottom, you
% can you this command to squash all of the vertical glue (stretchy space) with this command:
%
% \Squash 
%
% This command is useful when the text has not already reach the bottom of the page, since the glue gets squashed automatically
% when the page is too full.

%%%%%%%%%%%%%%%%%%%%%%%%%%%%%%%%%%%%%%%%%%%%%%%%%%%%%%%%%%%%%%%%%%%%%%%%%%%%%%%%%%%%%%%%%%%%%%%%%

%%% Make titlepage
\maketitle

%%%%%%%%% Contents that you need to write follows %%%%%%%%%%%%%%%%%%%%%%%%%%%%%%%%%%%%%%%%%%%%%%%%

% \includeonly{acknowledgments,biography,chapter1,chapter2,...,appendixa,...} 
%   for usage, see https://latexref.xyz/_005cinclude-_0026-_005cincludeonly.html

%%% Frontmatter (write this material in the mentioned files)  %%%%%%%%%%%%%%%%%%%%%%%%%%%%%%%%%%%%

% The abstract environment creates all the required headings and footers. 
% You only need to the text of the abstract in the file abstract.tex
\begin{abstract}
	\input{abstract.tex}% use \input rather than \include because we're inside an environment
\end{abstract}

\include{acknowledgments}% .tex extension is presumed by \include 

% \include{biography}% optional, see MIT Libraries https://libraries.mit.edu/distinctive-collections/thesis-specs/#format

%%% Table of contents and lists of stuff (delete lists you don't need, e.g., if no tables) %%%%%%%%

\tableofcontents
\listoffigures
\listoftables

%%% Chapters of thesis  %%%%%%%%%%%%%%%%%%%%%%%%%%%%%%%%%%%%%%%%%%%%%%%%%%%%%%%%%%%%%%%%%%%%%%%%%%%

%% If you want to use "double spacing", you should start here...

 \include{Introduction} 
 \include{Relatedwork}
 \include{Laserfactory}

 \include{Electrovoxel}
 \include{Pullup}

 \include{SelectiveMixels}
 \include{Learnings}
 \include{Conclusion}

%%% Appendicies of thesis  %%%%%%%%%%%%%%%%%%%%%%%%%%%%%%%%%%%%%%%%%%%%%%%%%%%%%%%%%%%%%%%%%%%%%%%%

\appendix
% \include{appendixa}

%%% Bibliography  %%%%%%%%%%%%%%%%%%%%%%%%%%%%%%%%%%%%%%%%%%%%%%%%%%%%%%%%%%%%%%%%%%%%%%%%%%%%%%%%%

\printbibliography[title={References},heading=bibintoc]

% biblatex also supports chapter-by-chapter bibliography, https://tex.stackexchange.com/a/296502/119566
% see the biblatex manual, section 3.14.3

%%%% Option for natbib %%%%%%%%%%%%%

%%   use an appropriate style (.bst) and your own .bib file[s]

%\bibliographystyle{plainnat}
%\bibliography{mitthesis-sample.bib}

\end{document}

%% file: abstract.tex
% From mitthesis package
% Version: 1.01, 2023/06/19
% Documentation: https://ctan.org/pkg/mitthesis
%
% The abstract environment creates all the required headers and footnote. 
% You only need to add the text of the abstract itself.
%
% Approximately 500 words or less; try not to use formulas or special characters
% If you don't want an initial indentation, do \noindent at the start of the abstract

Fabrication today relies on disparate, large machines spread across industrial facilities. These are operated by domain experts to construct and assemble artefacts in sequential steps from large numbers of parts. This traditional, centralized mass manufacturing paradigm is characterized by large capital costs and inflexibility to changing needs, complex global supply chains hinged on economic and political stability, and waste and over-manufacturing of uniform artefacts that fail to meet the technical and personal needs of today's diverse individuals and use cases. Today, these challenges are particularly severe at points of need, such as the space environment. The space environment is remote and unpredictable, and the ability to manufacture in situ offers unique opportunities to address new challenges as they arise. However, the challenges faced in space are often mirrored on Earth. In hospitals, disaster zones, low resource environments and laboratories, the ability to manufacture customized artefacts at points of need can significantly enhance our ability to respond rapidly to unforeseen events. In this thesis, I introduce digital fabrication platforms with co-developed hardware and software that draw on tools from robotics and human-computer interaction to automate manufacturing of customized artefacts at the point of need. Highlighting three research themes across fabrication machines, modular assembly, and programmable materials, the thesis will cover a digital fabrication platform for producing functional robots, a modular robotic platform for in-space assembly deployed in microgravity, and a method for programming magnetic material to selectively assemble.

%% file: acknowledgments.tex
%% acknowledgments.tex

% From mitthesis package
% Version: 1.01, 2023/10/16
% Documentation: https://ctan.org/pkg/mitthesis

\chapter*{Acknowledgments}
\addcontentsline{toc}{chapter}{Acknowledgments}

I have been advised and supported by a great number of people throughout my PhD. I would like to first thank my advisor, Stefanie Mueller, for giving me the unbounded freedom to pursue the research that was most meaningful to me. My committee members, John Hart and Daniela Rus, also provided invaluable support along the way. My thanks go to John for his insights and our discussions around hybrid manufacturing which helped anchor the ideas contained in this thesis in practical terms. Thank you Daniela for being a sounding board on topics around modular robotics and for your support during my time in CSAIL. I also have to thank my fellow lab members in the HCIE Lab, in whose company the lab became a second home. I have been fortunate to have had the chance to collaborate with incredible people from across CSAIL, MIT and beyond, whose expertise and friendship made this research not just exciting, but\textemdash perhaps more importantly\textemdash fun as well. 
Beyond the research introduced below, my thanks also go to my mentors, friends, and colleagues at the The Educational Justice Institute and Brave Behind Bars. Teaching behind the wall~\cite{BBB} has been one of the most meaningful experiences of my PhD.

%% file: Introduction.tex
\chapter{Introduction}
\label{sec:Introduction}

Manufacturing used to be personalized. Prior to industrialization, manufacturing globally was artisanal, designed to produce low-volume products customized to individuals needs and interests. As the world industrialized, first textiles and then a windfall of other goods were for the first time available as cheap, mass-produced goods as standardization gave rise to economies of scale. From washing machines to cars, this led to widespread increases in standard of living as the prices of these products dropped and their availability grew. However, over a hundred years have passed since the first Model T rolled off Ford's assembly plant in Detroit, and besides the organizational innovations of lean manufacturing pioneered by Toyota in the 1950s~\cite{womack2007machine}, manufacturing at the turn of the 21st century remained in many ways unchanged from a century ago. However, with the spread of digital manufacturing over the past two decades, a suite of new technologies has reinstated the ability for modern production tools to produce personalized artefacts. The operation of Moore's Law since 1965 has collapsed the prices and form factors of computational fabrication machines. Widespread adoption of Computer Aided Design (CAD) tools and advances in generative models, fueled by Artificial Intelligence (AI), are lowering the barrier to production for organizations and laypeople in ways that are fundamentally disrupting not just what can be made, but by whom. With these two parallel developments\textemdash the miniaturization and cost-effectiveness of fabrication capabilities on the one hand, and the increasing ease with which non-experts can design for them on the other\textemdash new opportunities are rising for \textit{in situ} manufacturing.

Manufacturing infrastructure in 2023 was still largely characterized by the same features that described Ford's automotive plants in 1923. Large capital costs for hyper-specialized machines are designed to mass-manufacture huge quantities of identical parts to generate the economics of scale required for profitability. As mass manufacturing spread across industries and plants metastasized across the globe, the accumulating costs of both mass manufacturing and mass consumerism have come more clearly into view. Conditioned for waste via overproduction and non-reyclability, costly warehousing leading to product degradation and obsoletion, and unabsorbed environmental externalities, the World Economic Forum reported in 2022 that manufacturing is now responsible for a fifth of global carbon emissions and over half of global energy consumption~\cite{WEF}. A single factory represents just one node in a global supply chain network of manufacturers and assembly plants, and the Environmental Protection Agency reports that transportation now contributes over a quarter of global carbon emissions~\cite{EPA}. The yoking of production to these supply chains has hinged delivery of crucial supplies of microchips and medical devices to global economic and political stability. The vulnerability of this network was made plain when in 2021, a 6-day blockage of the Ever Given in the Suez Canal caused up to 60-day shipping delays in \$60 billion of trade and grounded 5\% of the world's container fleet. Moreover, once delivered, the standardized artefacts produced by this network remain in many instances ill equipped to meet the modern needs and interests of particular use-cases and individuals. Companies developing hardware that push the boundaries of innovation have turned to fabricating their technologies in-house, and individual users that stray from bodily norms such as able-bodiedness and handedness find themselves unaccommodated for by the demands that economies of scale place on the mass manufacturing paradigm.  

But over the past two decades, digital manufacturing and robotics have set manufacturing on a radically new path. Characterized by digital design, automation and customization, wrapped in a hardware package that's inexpensive compared to traditional manufacturing infrastructure, digital manufacturing platforms like 3D printing, CNC milling and laser sintering have disrupted manufacturing industries by affording users the ability to rapidly and inexpensively produce customized parts of almost arbitrary geometries from a single design file. By 2015, production of end-use parts represented 51\% of additive manufacturing services~\cite{quinlan2017industrial}. New Balance, BMW and Invisalign are just few of a growing list of established mass manufacturers that have turned to additive manufacturing to produce flagship products, from shoe soles to orthodontics, at mass volume. In 2014, Made In Space 3D-printed a ratchet aboard the International Space Station from a file uplinked moments before~\cite{werkheiser20143d}. And until recently, the fuel injector for the European Space Agency's flagship Ariane 6 rocket was painstakingly assembled from 248 individually machined components; now it's printed in one go as a monolithic piece of nickel-based alloy. In the medical device industry, 3D printing is actively used to fabricate personalized implants, prosthetics, and most recently, drugs. In 2015, the US Food and Drug Administration approved the use of the first 3D-printed drug, and researchers have since printed tablets that for the first time offer patients their required drug dosages with personalized timed-release profiles~\cite{genina2012tailoring}. However, a focus on the advancements of individual fabrication technologies belies the exponential growth of the digital manufacturing sector as a whole. It is the form of exponential growth that led Gordon Moore to conjecture in 1965\textemdash to much public skepticism\textemdash the future appearance of "such wonders as home computers". Known today as Moore's law, he observed that over five years, a doubling had been occurring in the capacity to integrate components into a fixed size of integrated circuit, and suggested it could happen for another ten years. 

He was wrong; it happened for another fifty years. The trajectory of digital manufacturing appears to be following this trend. From the mid 2000s, Wohler Associates has for example reported sustained exponential growth in 3D printers sold globally~\cite{kianian2017wohlers}. Sales had grown 10-fold to 20,000 by 2011, to over 200,000 by 2015, and to over 2 million annual sales in 2021, estimated by Forbes at a market valuation of \$10.6 billion and counting~\cite{Forbes3dp}. This is fueled in part by the continued expiration of key patents, starting with Stratasys' patent on Fused Deposition Modeling in 2009~\cite{crump1992apparatus} and continuing with patent expirations related to performance materials, particularly metals, notable Selective Laser Sintering~\cite{beaman1991selective} in 2014 and Selective Laser Melting~\cite{meiners1998shaped} in 2016. In the late 2000s, the cheapest 3D printer commercially available cost \$10,000; a desktop printer can today be bought for \$100. This hundred-fold drop in both price and size is for a desktop machine capable of creating parts more geometrically complex than many production machines today. Parallel to the ability to fabricate customized hardware automatically, the ability to design them automatically is emerging in tandem. While generative models in AI have been widely popularized by the appearance of Deepfakes in video creation, related technologies are already employed for hardware creation. Researchers are today developing the tools to create 3D-printable designs based on textual user prompts~\cite{faruqi2023style2fab}. AutoDesk, one of the worlds leading CAD developers, has employed generative design in its flagship product since 2017 to suggest and complete users' CAD designs based on functional needs like load requirements. That same year, Autodesk published a tool equipping users with similar abilities for design and assembly of electronic circuits~\cite{anderson2017trigger}. ChatGPT, AI's most recent phenomenon, can generate printable files for objects based on natural language alone.

The emergence of miniaturized, affordable fabrication capabilities, together with new tools with which non-experts can design for them, creates a new opportunity for \textit{in situ} manufacturing at points of need. The space environment provides a vital case in point. Anything launched to orbit today must be packaged into a launch fairing, which it typically no larger than 3 meters in diameter. Inside the fairing, the artefacts being launched have also to have been designed to withstand the rigors of a launch, which can induce loads of up to 9g. And once these artefacts reach orbit, they have to be assembled, maintained, and repaired in one of the harshest environments we know of, by teams of people who don't necessarily have any manufacturing expertise. But these challenges are mirrored on Earth. Hospitals, laboratories, disaster zones, and even community maker spaces are all environments that share a demand for unique hardware solutions with speed and customization that a centralized mass manufacturing paradigm rarely affords. This is because centralized mass manufacturing places hard constraints on our ability to customize artefacts for individual needs and applications, to deliver them to the right place at times of need, and to transport them across both global and\textemdash increasingly\textemdash extra-planetary supply chains, that are subject not just to volume constraints, but also to global economic and political stability. Creating new manufacturing and assembly capabilities that can be deployed at points of need requires addressing two key challenges. First, we need portable hardware systems that can operate autonomously outside of traditional manufacturing ecosystems. Second, we need software tools that allow customization of the artefacts that these hardware systems can produce. But in order to achieve automated manufacturing at points of need, these hardware-software systems must be developed across a hierarchy of assembly levels from which artefacts are made today: at the machine level, at the part level, and at the material level. To address these challenges, this Thesis will introduce co-developed hardware and software to enable in situ assembly using multi-process manufacturing machines, modular assembly platforms, and programmable materials.

For a machine to manufacture a product in a modern sense, it must be able not only create its geometry, but also to assemble it with the electronics that imbue it with \textit{function}. Automating the manufacturing of functional artefacts of this kind requires \textbf{multi-process manufacturing machines}. Companies like Voltera and BotFactory today sell desktop machines capable of printing functional printed circuit boards (PCBs) in the same form factor and price point as a commercial 3D printer. These are being used in both industry and academia to rapidly build circuit boards for devices ranging from kitchen appliances and wearables to robots. Practitioners have begun to try to consolidate printers and electronic manufacturing processes into singular machines to print artefacts from robots to flexible electronics~\cite{valentine2017hybrid}. Other researchers have developed new fabrication pipelines lower the barrier to protoyping 3D artefacts with embedded electronics~\cite{zhu2020curveboards}. However, the ability to manufacture custom electromechanical artefacts, like robots, at the push of a button, has remained out of reach. This requires the consolidation of manufacturing processes that span geometry creation, circuit fabrication, and parts assembly. This thesis will address these challenges, and demonstrate a method to manufacture integrated devices and robots in a singular machines without manual intervention.  

The digitization of fabrication technologies, coupled with the fall in both the cost and size of electronics, has also spurred advancements in the notional fabrication paradigm known as self-assembly. Known sometimes as programmable matter, bottom-up self-assembly shares the vision of a singular platform for manufacturing. However, instead of accomplishing assembly in a top-down manner using an external agent or manipulator, self-assembly is a parallel process for bottom-up fabrication using information and actuation embedded in individual parts or materials themselves. This paradigm is actively explored across the scientific community, spanning fields of modular robotics, architecture, biology and materials science. At the molecular scale, self-assembly is effectively used in nature to assemble complex biological structures \cite{whitesides1991molecular}, and formal models evaluating the power of self-assembly have proven it to be Turing-universal. Seeking to harness its capabilities, engineers have used self-assembly across scales to create microstructured materials with designed optical and mechanical properties at the $\mu$m-scale \cite{grzybowski2003electrostatic}, to self-assembled structures at the mesoscale~\cite{nisser2016feedback} up to aerial self-assembly of structures at the m-scale \cite{papadopoulou2017self}. As an alternative to automating the assembly of robots in a top-down manner~\cite{nisser2021laserfactory}, the reconfigurable robotics community is exploring bottom-up self-assembly in two using both active modules and passive materials. In the short term, it is seen as particularly important for addressing the need for in situ assembly in logistically challenging environments like space, and for enabling assembly of structures at the nano and micro scales, such as molecular machines for medical interventions. This thesis will explore self-assembly using both active agents and passive materials, and demonstrate their applicability with physical hardware. To instantiate the idea of active agents, we will introduce \textbf{modular self-assembly platforms} as a method to self-reconfigure parts of a structure into a target shape. To instantiate the idea of passive assembly, we will introduce \textbf{programmable materials} as platform for programming materials themselves to assemble into target shapes. We will introduce a coding-theoretic framework by which materials can be programmed for selective self assembly, and whose generalizability across materials and scales will be demonstrated by instantiating these programs both magnetically at the mesoscale and chemically at the nanoscale, using DNA. Enabling the fabrication of custom hardware in situ requires a parallel effort to create software tools. As traditional manufacturing systems are operated by experts, these new tools must allow new end users to effectively design the artefacts that these new fabrication systems can produce. A final contribution of this thesis will be to introduce software tools that empower users to generate designs, manufacturable fabrication files, and automated assembly procedures suited to consolidated assembly platforms introduced herein, via both assembly and self-assembly.

\begin{figure}[ht]
  \includegraphics[width=\columnwidth]{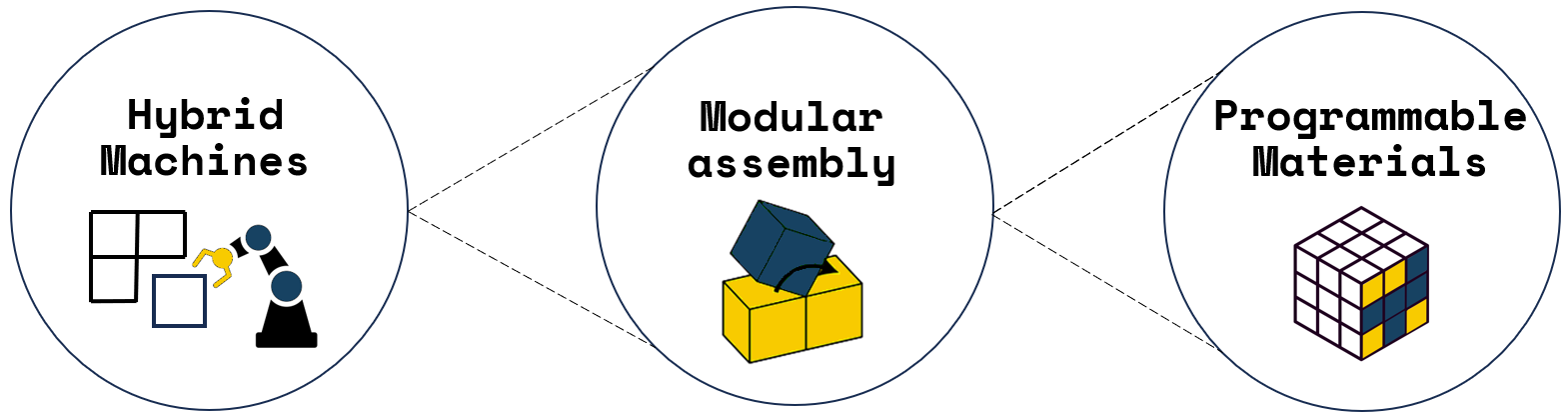}
  \caption{Automating fabrication and assembly at three levels: at the Machine level, at the Part level, and at the Material level. This thesis will outline methods that demonstrate these hierarchical assembly techniques using three platforms: (1) multi-process manufacturing machines, (2) modular assembly platforms, and (3) programmable materials.}
  \label{fig:hm-ma-pm}
\end{figure}

We have just introduced the motivations behind the problems this thesis seeks to address, and the key work carried out in its service. In summary, this thesis will build on recent advancements in digital fabrication and hardware miniaturization to develop consolidated manufacturing techniques capable of producing functional artefacts in singular platforms. I will demonstrate opportunities to achieve this via both top down assembly methods, using hybrid manufacturing workcells, and bottom-up self-assembly, using modular systems and programmable materials. Accompanying the hardware platforms, I will introduce software tools to help users to design and define assembly protocols. I will use these platforms to demonstrate fabrication of advanced, functional artefacts across both scales and use cases, including robots, space structures and molecular machines. Next, we will summarize the motivation and challenges associated with these three methods for automating fabrication and assembly in situ: (1) multi-process manufacturing machines, (2) modular assembly platforms, and (3) programmable materials. In these subsections, we will also enumerate the thesis' primary academic contributions, which span contributions to design, experimental work, algorithms, and applications. For clarity, in in keeping with the structure of this thesis, these contributions are categorized according to their relevance to machines, modules, or materials. Together, these introduce methods to automate fabrication in situ at three different levels: at the machine level, the part level, and at the material level.

\section{Machines}

To automate fabrication at the machine level, multi-process manufacturing machines are required that consolidate traditionally disparate manufacturing and assembly processes. 

To fabricate a fully functional device\textemdash a 3D structure with integrated electronics and actuators\textemdash requires three steps; (1) fabricating the device geometry, (2) creating conductive circuit traces, and (3) populating the structure with electronic components. Researchers have in recent decades developed numerous personal fabrication methods that democratize the creation of such functional devices. These methods typically focus on extending the abilities of existing fabrication machines designed to create geometry to also permit creating conductive traces, for example by generating hollow internal pipes for 3D prints that can be filled with conductive material~\cite{schmitz2015capricate}, or using laser cutters to selectively ablate conductive films that can be populated with components~\cite{groeger2019lasec}. Printable Hydraulics~\cite{maccurdy2016printable} fabricated a robot complete with Hydraulic actuators that could be used for walking, but required electronic traces and components to be affixed separately. Peng et al~\cite{peng20163d} fitted a coil winding mechanism onto a 3D printer head to print electromagnets by feeding conductive wire in tandem with traditional filament, but artefacts were limited to plastic devices with solenoids, and required external power and circuitry. Voxel8~\cite{Voxel819} deposited silver circuit traces in tandem with traditional 3d printing, but lacked the component assembly and selective curing required to complete fabrication of functional devices end-to-end. An approach by Valentine et al.~\cite{valentine2017hybrid} uses a silver dispenser to create circuit traces and a pick-and-place mechanism to automatically assemble components onto a commercial 3-axis motion control stage. However, without adding a dedicated curing mechanism to their additive direct-writing fabrication process, it lacks the capability to cure the traces during fabrication. Thus, following fabrication, the resulting devices from Valentine et al. must be manually removed and separately cured at elevated temperatures in an oven for an additional two hours to be made functional. This can cause problems as existing work has shown that when electronic components are placed into an oven to cure silver, the temperature can damage the parts (Shrinky Circuits~\cite{lo2014shrinkycircuits}). In addition, subjecting the entire material substrate to high temperature may cause warping or degradation of many additively printed polymers. As yet, personal fabrication machines do not have the ability to integrate all three steps, and instead rely on human intervention to complete the fabrication procedure. As such, the fabrication of functional devices continues to demand engineering expertise, for example, to etch PCBs or solder components. In addition, repeatability and precision in the construction is compromised, as techniques based on manual intervention impede the ability for engineered devices to be made to specification.

A system we developed in 2021 called LaserFactory~\cite{nisser2021laserfactory} introduced a vertically integrated machine capable of manufacturing fully-functioning robots and customized devices, including quadrotors and electronic health wearables, within minutes. The machine consists of a \$150 add-on to a lasercutter which extends the laser's subtractive manufacturing capabilities with a conductive extruder, to deposit circuit traces, and a pick-and-place mechanism, to place components. In an accompanying software tool, users can drag-and-drop components and wires from a parts library onto a canvas to design a device, and then print it with a click. In 10 minutes, the machines manufactures a fully functional quadrotor that flies directly out of the platform; in 8 minutes, it prints a sensorized electronic health wristband tailored to the users's wrist, ready for use. Researchers have since begun to replicate this hybrid manufacturing architecture to fabricate quadrotors in larger form factors using a larger workspace navigated by a robotic arm~\cite{kosmal2022hybrid} 

\subsection{Contributions}

In summary, this thesis makes the following contributions to the development of direct hybrid manufacturing platforms:

\begin{itemize}
    \item A hardware add-on consisting of a silver dispenser and a pick-and-place mechanism to augment an existing laser cutter's functionality with the ability to create circuit traces and place components.
    
    \item $\mathrm{CO_2}$ laser sintering, a technique for curing dispensed silver traces using the attenuated power of a $\mathrm{CO_2}$ laser cutter to create highly conductive circuit traces. 
    
    \item A motion-based signaling technique for augmenting an existing fabrication platform without the need to interface with its underlying software.
    
    \item A software toolkit consisting of three elements: a tool that supports the concurrent design of device geometry, circuit traces and electronic component layouts; a tool that visualizes the resulting 3D geometry and steps of the fabrication process; and a translation pipeline that converts the design into machine instructions for the fabrication device and hardware add-on.

    \item Demonstration applications using our multi-process manufacturing machine, including the first end-to-end fabrication of a functional quadrotor with no manual intervention.
    
\end{itemize}

\section{Modules}

To automate assembly at the part level, modular assembly platforms conceive of partitioning a target structure into parts ("modules"), and embedding the actuation and information required for assembly into the modules themselves. This form of assembly typically takes one of two approaches: discretizing a structure into disjoint 3D parts which are made to reconfigure, or discretizing a structure into kinematically linked 2D surfaces which are made to fold.

Seeing reconfiguration as an alternative to the traditional, top-down method of assembly, roboticists in particular have explored bottom-up approaches to assembly by embedding actuation into individual parts. The result has been a vision of modular self-reconfigurable robots that has been pursued for over 30 years~\cite{stoy2010self,gilpin2008miche,yim2000polybot}. Exhibiting unique benefits in adaptability, scalability, and robustness, modular self-reconfigurable robots (MSRR) promise application domains that include space exploration~\cite{yim2003modular,baca2014modred, haghighat2022approach}, reconfigurable environments~\cite{sprowitz2014roombots,neubert2016soldercubes}, search and rescue~\cite{daudelin2018integrated}, and shape-changing user interfaces~\cite{roudaut2016cubimorph, nisser2021programmable}. Roboticists have typically built MSRR via individually actuated modules connected by temporary joints~\cite{rus2001crystalline,sprowitz2014roombots,gilpin2010robot}. MSRR based on cubic modules have moreover achieved self-reconfigurability in two dimensions via sliding~\cite{an2008cube} and disassembly~\cite{gilpin2011making}, as well as in three dimensions via pivoting~\cite{romanishin20153d,romanishin2013m}. However, a grand challenge facing self-reconfigurable structures and robots is their scalability~\cite{yim2007modular}---existing designs often require separate mechanisms for actuation and attachment~\cite{sprowitz2014roombots, romanishin20153d}, and in addition, require mechanical components such as motors, gears, and transmissions. However, these components are often bulky, complex, and expensive, hindering their miniaturization and scalability. 

One promising approach to addressing these challenges is to leverage electromagnetic forces to connect and actuate modules at the same time with a single component. Being solid-state\textemdash that is, having no moving parts itself\textemdash it is also easy to maintain and manufacture for a large-scale system. In one paper~\cite{nisser2017electromagnetically} I proposed and simulated the use of inexpensive electromagnets embedded into cube edges to actuate pivots between adjacent modules via repulsion while creating temporary hinges via attraction. Unlike traditional hinges that require mechanical attachments between two elements, this approach requires no dedicated physical mechanism and can be formed between any electromagnet pair dynamically. In Electrovoxel~\cite{nisser2022electrovoxel}, we introduced a cube-based reconfigurable robot that utilizes an electromagnet-based actuation framework to reconfigure in three dimensions via pivoting. While a variety of actuation mechanisms for self-reconfigurable robots have been explored, they often suffer from cost, complexity, assembly and sizing requirements that prevent scaled production of such robots. To address this challenge, we use an actuation mechanism based on electromagnets embedded into the edges of each cube to interchangeably create identically or oppositely polarized electromagnet pairs, resulting in repulsive or attractive forces, respectively. By leveraging attraction for hinge formation, and repulsion to drive pivoting maneuvers, we can reconfigure the robot by voxelising it and actuating its constituent modules\textemdash termed Electrovoxels\textemdash via electromagnetically actuated pivoting. To demonstrate this, we developed fully untethered, three-dimensional self-reconfigurable robots and demonstrate 2D and 3D self-reconfiguration using pivot and traversal maneuvers on an air-table and in microgravity on a parabolic flight. This paper described the hardware design of our robots, its pivoting framework, our reconfiguration planning software, and an evaluation of the dynamical and electrical characteristics of our system to inform the design of scalable self-reconfigurable robots.

Identifying folding as a compelling alternative to reconfiguration due to its speed, researchers have also previously leveraged embedded actuation to fold low-profile sheets into 3D structures using materials such as shape memory polymers~\cite{nisser2016feedback,felton2014method} and shape memory alloys~\cite{firouzeh2015robogami} as well as using external activation via uniform heating~\cite{tolley2014self} and selective laser heating~\cite{nisser2021laserfactory}. Methods for unfolding 3D polyhedral meshes into 2D sheets have also been studied across mathematics, physics, mechanical engineering and computer graphics \cite{takahashi2011optimized,demaine2005survey}. However, existing methods require dedicated machinery such as lasers, ovens or local joule-heating to activate material actuators. These actuators can also be challenging to use as they must be sufficiently low-profile to be embedded in the sheet, reducing their strength and often limiting candidates to material-based actuators such as shape memory materials which are challenging to control and often cannot be activated bidirectionally. They are typically redundant after initial use, and the embedding of these actuators results in a thickening and weighing down of the laminate itself, compromising both foldability and portability.

In PullupStructs~\cite{niu2023pullupstructs}, we introduce a method to rapidly create 3D geometries from 2D sheets using pull-up nets: a string routed through the planar faces which can be pulled by a user to fold the sheet into its target 3D structure. This provides a way to fold a sheet into its target shape using common string or nylon, using just \textit{a single actuated degree of freedom} controlled by a user. Our string-based design is highly inexpensive, requires no dedicated machinery, can be actuated powerfully via external means, and can be removed following fabrication. While prior work has explored pull-up nets to fold 3D geometries~\cite{meenan2008pull}, this was restricted to manual fabrication designs for the 5 platonic solids; in our work, we provide a tool to automatically compute these designs for all admissible geometries and introduce a digital fabrication pipeline to manufacture them. Given a 3D target structure, this process unfolds its 3D mesh into a planar 2D sheet populated with cutlines and throughholes. After laser-cutting the sheet and feeding thread through these throughholes to form a pull-up net, a user pulls on the thread to fold the sheet into the 3D structure. We introduce the fabrication process and build several prototypes demonstrating the method's ability to rapidly create a breadth of geometries suitable for low-fidelity prototyping across a wide range of applications. 

\subsection{Contributions}

In summary, this thesis makes the following contributions to the development of modular assembly and folding techniques:

\begin{itemize}

    \item The design and fabrication of an untethered, modular robotic unit capable of pivoting in 3D using an electromagnetically actuated pivoting technique.
    
    \item A parameterized version of Amp\`ere’s Force law that relates custom electromagnet designs to the forces they produce, illustrating the design space of electromagnetic actuators for reconfiguration.
    
    \item Experimental results that empirically validate our our force model relating electromagnet parameters to the forces produced.

    \item Demonstrations of the reconfigurable system using a simulated microgravity environment with an air table, followed by deployment on a parabolic flight to demonstrate untethered three-dimensional reconfigurability in 3D space without the kinematic constraints of a ground plane. 

    \item A reconfiguration planning software that allows users to specify a high level reconfiguration sequence, and that outputs the underlying electromagnet assignments required to achieve the specified input.

    \item A pull-up net technique for rapidly cutting and folding sheets into 3D structures, using an algorithm to unfold a 3D mesh into a planar 2D sheet populated with cutlines and throughholes.

    \item A web-based user interface, equipped with both a pre-existing set of 141 polyhedral meshes and a feature to accept custom 3D meshes, that outputs manufacturable 2D sheets for use on a laser cutter.
  
    \item A digital fabrication pipeline centering on a laser cutter to rapidly fabricate structures based on the outputs of the user interface.

    \item A range of fabricated geometries that demonstrate our pull-up net method's ability to rapidly create both aesthetic and functional geometries spanning a range of applications in low-fidelity rapid prototyping.
    
\end{itemize}

\section{Materials}

To automate fabrication at the material level, programmable materials conceive of embedding the actuation and information required for assembly into materials themselves.

As introduced above, self-assembly by modular self-reconfigurable robots (MSRR) involves modules that modulate their behavior online in order to locate, position and bond themselves to their neighbors, for which each module requires embedding with computation, sensing and actuation. While active assembly has been used to successfully reconfigure a variety of robotic systems, it is these actuators that are typically the most significant challenge to scaling systems up in number and down in size due to the cost and complexity of embedding them into individual modules \cite{zykov2007experiment}.

In contrast, passive self-assembly engendered by material programming obviates the need for active actuation and control. Instead, system actuation is outsourced to an external excitation, and in the case of \textit{stochastic} self-assembly~\cite{tibbits2012self}, this excitation requires no local control and is governed instead only by global parameters such as excitation magnitude. Stochastic assembly sacrifices efficiency and predictability for advantages in cost, complexity and scale; by enabling the environment to actuate reconfiguration, it trades off deterministic assembly times of individual modules for statistical assembly rates of the collective. To assemble stochastically, modules require pre-programming to enforce correct mating during random collisions with their intended mate. This programmed specificity between pairs of mating faces is typically achieved via minimization of free surface energy via topology\cite{hacohen2015meshing}, wettability \cite{bowden1997self}, magnetic forces \cite{lu2021enumeration} or electrostatic \cite{grzybowski2003electrostatic} interaction. While the wider scientific community has often been interested in constraining the self-assembly problem to 2D, for instance by using a shaker table \cite{jilek2020centimeter}, roboticists have leveraged liquid tanks to study assembly in 3D. Fluidic assembly at the mesoscale has become a particularly widely studied problem in robotics \cite{tolley2008dynamically,tolley2010fluidic,tolley2011programmable,krishnan2008increased,kalontarov2010hydrodynamically, zykov2007experiment}. Existing stochastically self-assembling modules typically include two features to enable assembly: first, embedded magnets that generate near-field forces to bring modules close, and second, selective geometry on module faces that encodes the specificity to only permit bonds between mating pairs \cite{jilek2021towards,hacohen2015meshing,jilek2020centimeter, tsutsumi2007multistate}. However, three key challenges remain for the development of stochastic self-assembling systems: (1) \textit{scalability} that shows how modules can be made both numerous and small; (2) \textit{selectivity guarantees} that help bound module misassembly; and (3) \textit{reconfigurability} that let modules acquire different target shapes. 

In a paper titled Selective Self-Assembly using Re-Programmable Magnetic Pixels~\cite{nisser2022selective}, and its extension, Mixels~\cite{nisser2022mixels, nisser2021stochastic, nisser2022demonstration}, we address these concerns. This paper introduces a method to generate highly selective encodings that can be magnetically "programmed" onto physical modules to enable them to self-assemble in chosen configurations. We generate these encodings based on Hadamard matrices, and show how to design the faces of modules to be maximally attractive to their intended mate, while remaining maximally agnostic to other faces. We derive guarantees on these bounds, and verify their attraction and agnosticism experimentally. Using cubic modules whose faces have been covered in soft magnetic material, we show how inexpensive, passive modules with planar faces can be used to selectively self-assemble into target shapes without geometric guides. We show that these modules can be easily re-programmed for new target shapes using a CNC-based magnetic plotter, and demonstrate self-assembly of 8 cubes in a water tank. Finally, we also show how to the selective encodings we generate can be applied to other materials, and highlight initial results from using our technique to self-assemble DNA. For prior work on programmming materials to disassemble via dissolution, please see Nisser et al.~\cite{nisser20193d}.

\subsection{Contributions}

In summary, this thesis makes the following contributions to the development of selectively programmed materials for assembly:

\begin{itemize}

    \item An algorithm to generate 2D arrays, based on Hadamard matrices, that exhibit low pair-wise cross-correlations for both translated and rotated arrays.

    \item A method to instantiate these arrays as magnetic signatures that form selectively attractive interfaces, using a custom built magnetic plotter to program arrays into soft magnetic material.

    \item An evaluation of the number of selectively attractive interfaces that can be generated given the quality of the cross-correlation between the underlying arrays. 
    
    \item An evaluation of our ability to make predictions with regard to the attraction and agnosticism between magnetically programmed interfaces, verified against experimental results. 
    
    \item A demonstration of 8 soft magnetic modules, programmed with our magnetic signatures, that can self-assemble into a target shape under stochastic agitation in a fluidic environment.

\end{itemize}

%% file: RelatedWork.tex
\chapter{Related Work}
\label{sec:Relatedwork}

This Thesis builds on research across a range of disciplines and communities, primarily across robotics, fabrication, and human-computer interaction. The work on multi-process manufacturing machines is related to research that augments existing fabrication machines as well as research that investigates how to make functional electromechanical devices by fabricating circuit traces and assembling electronic components. The work on modular assembly is related primarily to research in modular robotics. Finally, the work on programmable materials is related chiefly to work in robotics, materials, and tangible user interfaces.

\subsection{Fabricating Circuit Traces for Electromechanical Devices}

Researchers have developed a variety of methods in pursuit of functional device fabrication. One of the first approaches was to augment passive 3D printed objects with circuit functionality by adding circuit traces to them after fabrication. For instance, in the Un-Toolkit~\cite{mellis2013microcontrollers}, users sketch circuits using a 2D drawing pen loaded with silver ink. Similarly, in Midas~\cite{savage2012midas}, users attach conductive tape to existing objects to enhance them with circuit functionality. Extending this work, robotics researchers printed joule-heating circuits onto copper-clad kapton which were etched in ferric chloride and attached to shape memory polymers to stimulate self-folding of the underlying structure~\cite{nisser2016feedback}. Researchers have also developed computational materials that integrate circuit traces and discrete components directly into laminate sheets to create functional structures~\cite{yang2023compumat}. PipeDream~\cite{savage2014series} 3D prints objects with internal pipes and then asks users to fill the pipes with copper paint after fabrication to make them conductive. Similarly, Curveboards~\cite{zhu2020curveboards} first 3D prints a housing before requiring channels to be manually filled with conductive silicone. Extending this work, SurfCuit~\cite{umetani2017surfcuit} creates channels on the surface of a 3D printed object that users affix copper tape to and solder the components on. 

More recently, researchers have investigated how to automate the circuit creation process. For instance, a laminate sheet (base material and conductive copper layer) can be used in a laser cutter to create circuit traces on the sheet’s surface via cutting (Foldtronics~\cite{yamaoka2019foldtronics}) or ablation (LASEC~\cite{groeger2019lasec}). Another technology, inkjet printing of silver ink, can print circuit traces directly onto existing 2D substrates (Instant Inkjet Circuits~\cite{kawahara2013instant}, Building Functional Circuits~\cite{kawahara2014building}). With the availability of conductive filament for 3D printing, researchers have also started to co-create object geometry and circuit traces in one integrated process. PrintPut~\cite{burstyn2015printput}, Capricate~\cite{schmitz2015capricate}, ./Trilaterate~\cite{schmitz2019trilaterate}, and Flexibles~\cite{schmitz2017flexibles} all use carbon-based conductive filament alongside regular filament to fabricate capacitive touch sensors and wires onto 3D geometries. Carbon-based conductive filament, however, has a high trace resistance and does not support all forms of electronics. Newer approaches such as 3D printing with silver ink (Voxel8~\cite{Voxel819}) and carbon fiber (FiberWire~\cite{swaminathan2019fiberwire}) produce lower resistance connections and therefore increase the range of electronic components and circuits that are practical.

\subsection{Automatically Integrating Electronic Components}

None of the techniques mentioned above automatically integrate electronic components as part of the fabrication process, which is a task that must be completed after fabrication, usually manually. Hodges et al. demonstrate how to facilitate this process by attaching electronic components to stickers, which can then be added to the circuit (Circuit Stickers~\cite{hodges2014circuitstickers}). 
% Most closely related to our work is a
An approach by Valentine et al.~\cite{valentine2017hybrid} uses a silver dispenser to create circuit traces and a pick-and-place mechanism to automatically assemble components onto a commercial 3-axis motion control stage. However, without adding a dedicated curing mechanism to their additive direct-writing fabrication process, it lacks the capability to cure the traces during fabrication. Thus, following fabrication, the resulting devices from Valentine et al. must be manually removed and separately cured at elevated temperatures in an oven for an additional two hours to be made functional. This can cause problems as existing work has shown that when electronic components are placed into an oven to cure silver, the temperature can damage the parts (Shrinky Circuits~\cite{lo2014shrinkycircuits}). In addition, subjecting the entire material substrate to high temperature may cause warping or degradation of many additively printed polymers. To address this issue, we build our hardware augmentation onto an existing laser cutter and use the heat of the laser to both cure traces made from silver paste and also solder components into place. For this, we build on work from Lambrichts et al.~\cite{lambrichts2020diy}, who used a laser to cure solder paste. Solder paste is different from silver paste in that it can be used to solder components but not to create circuit traces as it beads up when soldered due to surface tension. Researchers have previously used laser sintering of inkjet-printed silver nanoparticle inks to fabricate traces on paper substrates, however these traces were pre-baked in an oven at 110$^\circ$C for 20 minutes, and required components to be added separately via traditional soldering~\cite{balliu2018selective}. In contrast, we demonstrate that laser soldering can be used to both cure silver paste without pre-baking the silver compound, and to simultaneously bond components, yielding a method that can quickly create and cure circuit traces and thus can make fully functional devices in just a few minutes. As laser soldering addresses specific traces selectively, it also removes the issue of subjecting the material and electronic components to thermal stress across the entire sheet.

Rather than adding existing electronic components to the circuit, several researchers have investigated how to fabricate electronic components from scratch. Lewis et al.~\cite{lewis2015device}, for instance, pioneered 3D printing an LED. A 3D printer for electromagnetic devices~\cite{peng20163d} showed how motors and solenoids can be created by winding coils around 3D geometry during the printing process. However, since most electronic components cannot yet be created with personal fabrication tools, the fabricated objects still need to have commercially available electronic components, such as a microcontroller, added to their circuit after fabrication.

\subsection{Augmenting Existing Fabrication Machines}

Several research projects have augmented existing fabrication machines with additional hardware to perform new tasks. One of the first extensions of fabrication machines was to augment laser cutters with a camera to be able to detect the material on the fabrication bed (LaserCooking~\cite{fukuchi2012laser}, PacCam~\cite{saakes2013paccam}). More recently, researchers created hardware add-ons to fabricate new types of geometries, for instance, by adding a mill head to an existing 3D printer (ReForm~\cite{weichel2015reform}). A mill can also be used to undo previous fabrication steps, by removing formerly created geometry (Patching Physical Objects~\cite{teibrich2015patching}, Scotty~\cite{mueller2015scotty}). Researchers have also modified the existing stationary build plate of 3D printers and replaced it with a rotating platform to allow 3D printing onto objects at an angle (Patching Physical Objects~\cite{teibrich2015patching}, Revomaker~\cite{gao2015revomaker}). To assemble 3D printed parts into more complex mechanisms, researchers created an extension to a 3D printer's head that can pick up parts from the platform and place the parts into a new position (3D Printer Head as a Robotic Manipulator~\cite{katakura20193d}). In Fabrobotics~\cite{bhattacharya2024fabrobotics}, researchers fused a 3D printer with mobile robots by embedding robot commands into Gcode, allowing both the robots to manipulate 3D printed parts, and allowing the printer to print directly onto the robots to augment their capabilities. Finally, researchers added custom hardware add-ons to existing fabrication devices to investigate how to fabricate electronic components from scratch. A 3D Printer for Interactive Electromagnetic Devices~\cite{peng20163d}, for instance, uses a wire delivery mechanism to wind coils thereby creating a first set of 3D printed electromagnetic devices.

Jubilee~\cite{vasquez2020jubilee} highlights that building fabrication machines is a significant obstacle to creating novel fabrication processes facing researchers. While this motivates augmenting existing platforms rather than building them from scratch, this introduces the challenge of how to synchronize the add-on's functionality with that of the existing platform. Xprint ~\cite{wang2016xprint} underlines that hardware constraints and non-open source firmware remain major challenges to integrating new fabrication processes into commercial platforms. Existing hardware add-ons, such as those mentioned in the previous section, have thus mostly relied on open-source hardware: A 3D Printer for Interactive Electromagnetic Devices~\cite{peng20163d}, for instance, modified a 3D printer's software control files to accept their custom G-code for controlling their wire-coiling add-on. Similarly, Patching Physical Objects~\cite{teibrich2015patching} used an open-source implementation of the MakerBot 3D Printer firmware to control the timing of the added mill, pump, camera and 5-axis rotating platform. However, relying solely on open-source implementations significantly limits the number of fabrication devices that can be augmented. 

% To be able to augment existing fabrication devices without the need to modify the underlying firmware, we developed a new motion-based signaling technique. Our technique embeds signals into the fabrication file that result in specific movements of the fabrication head, which can then be sensed with a motion sensor (accelerometer) attached to the fabrication head. Such signals can inform the hardware add-on when to start and stop its operation since they represent which part of the fabrication file is currently executed. 

\subsection{Modular Self-assembly via Reconfiguration}

Research in modular self-assembly envisions partitioning a structure into parts, or modules, and using embedded actuation to reconfigure these modules into a target structure. To this end, modular and reconfigurable robots have received widespread attention from the robotics community over several decades. These systems are typically comprised of modular, homogeneous robotic agents that use electromechanical actuators to bond, pivot, or linearly traverse each other. As Chennareddy et al.~\cite{chennareddy2017modular} show, modular self-reconfigurable robots present wide and unique solutions for increasing demands in a diverse set of fields including automation, space exploration and consumer products. One key way in which these robot systems differ centers on their reconfiguration mode; the most common among these are pivoting, sliding, and disassembly.

Reconfiguration via pivoting has seen considerable attention in the research community. 3D M blocks~\cite{romanishin20153d}, and its precursor M blocks~\cite{romanishin2013m}, use momentum driven flywheels to generate pivoting maneuvers between neighboring cube-shaped blocks. However, individual blocks are expensive and complex machines that are difficult to assemble, impeding scalability. Nisser et al.~\cite{nisser2017electromagnetically} introduced a framework for how addressable electromagnets embedded in cubic modules can be used to create pivoting maneuvers by polarizing electromagnets to create temporary hinges and actuating forces. Building on this, Kubits~\cite{hauser2020kubits} demonstrated reconfiguration of cube-shaped modules embedded with programmable magnets in their edges, however their significant power demands require tethering to house electronics and power off-board. Demonstrating the promising range of morphologies that pivoting cubes enable, Sung et al.~\cite{sung2015reconfiguration} developed algorithms to achieve reconfiguration between two distinct configurations using pivoting cubes that are provably correct in under O(n2) moves, barring three inadmissible sub-configurations. Others have also turned to eletromagnetic actuators, particularly for fluidic and spaced-based applications. In Lily~\cite{haghighat2016fluid}, researchers embedded electropermanent magnets into modules to demonstrate fluid-mediated programmable stochastic self assembly in 2D. Ekblaw~\cite{ekblaw2018tesserae} leveraged electropermanent magnets to develop a self-assembly framework for architectural scale space structures. 

Other researchers have investigated how to reconfigure modules via sliding. EM-Cube~\cite{osborne1997cube} showed how a group of tessellating cubes can be reconfigured via sliding by embedding cubes with an array of permanent magnets and electromagnets in parallel. Benoit et al.~\cite{piranda2013new} developed a set of sliding cubes fitted with an array of conveyor-enabling actuators on their top surface to form reconfigurable conveyor belts for small parts. In addition, StickyBricks~\cite{schweikardt2007stickybricks} introduced a method for reconfiguring cubes using an actuated adhesive belt, though this was only demonstrated in 2D. 

Roboticists have also explored multiple ways to connect stationary modular robots using temporary joints~\cite{rus2001crystalline,sprowitz2014roombots,gilpin2010robot}, and how to disconnect them programmatically for disassembly. Pandey et al.~\cite{pandey2016assembly} folded 2D integrated circuits into 3D  blocks to make cellular computational devices. Miche~\cite{gilpin2008miche} showed how a set of cubic modules temporarily aggregated using electromagnets can acquire a shape via disassembly. Pebbles~\cite{gilpin2010robot} built on this work by developing algorithms that further allowed similar cubic modules to dublicate, or "scan", objects using said cubic modules, and recreate its shape in turn. Turning again to magnetism, Softcubes~\cite{yim2014softcubes} introduced a system comprised of stretchable, serially connected cubes capable of self-recovery using embedded magnets.

\subsection{Modular Self-assembly via Folding}

A range of prior works, particularly in HCI, has explored ways to enable the fabrication of 3D structures by folding laser-cut 2D sheets. Flaticulation~\cite{10.1145/3526113.3545695} used a set of cut-in-place joint patterns to enable users to fold laser-cut sheets into target shapes via articulated angles. crdbrd~\cite{hildebrand2012crdbrd} used a laser cutter to fabricate mutually intersecting planar cut-outs from 3D shapes which can be assembled to form low fidelity 3D prototypes. LaserOrigami~\cite{mueller2013laserorigami} and LaserFactory~\cite{nisser2021laserfactory} fabricated 2.5D objects in a laser cutter by attenuating the laser power to heat an acrylic substrate until it becomes compliant enough to fold due to gravity. LaserStacker~\cite{umapathi2015laserstacker} similarly used attenuated laser power to weld sheets of acrylic together to form low-fidelity 3D prototypes. HingeCore~\cite{abdullah2022hingecore} used sandwiched materials laser-cut with finger joints to a partial depth in order to permit a single sheet to be assembled via folding without the need of glue or tabs. InfOrigami~\cite{tao2021inforigami} unfolds 3D meshes into laser-cuttable and color-coded 2D sheets which can be assembled by users with the help of glue.  

However, methods to date may be difficult to scale to high-resolution structures with hundreds or thousands of folds. Structures manufactured using methods that consolidate the folding procedures within the laser cutter platform can be constrained by occlusion between target hinges and the laser. Similarly, user-folded methods may become burdensome and unwieldy by requiring users to manually fold each hinge individually. In this work, we propose laser-cut structures that can be folded in a single step, by actuating a single degree of freedom transmitted via a pull-up string.

The problem of unfolding a polyhedron into a non-overlapping net has challenged researchers for decades, and virtually no guarantees exist for unfolding general polyhedra \cite{demaine2007geometric,demaine2005survey}. The computational effort needed to search all possible unfoldings suffers from combinatorial explosion as the size of the polyhedron increases. Even more challenging is the solving the full ``blooming'' problem, which seeks to avoid collisions between faces as the polyhedron is unfolded \cite{song2004motion,biedl2005can}. Nevertheless, for simple polyhedral meshes, and for meshes with small dihedral angles, heuristic methods offer promising results \cite{schlickenrieder1997nets,chen2013edge}. Although few guarantees exist for finding non-overlapping nets, existing heuristic algorithms have succeeded on thousands of polyhedra searched so far, including shapes with thousands of vertices and faces \cite{demaine2007geometric,schlickenrieder1997nets}. Certain guarantees exist for finding unfoldings, including restriction to convex polyhedra where faces may also be cut \cite{aronov1991nonoverlap}, but the general problem of unfolding along edges remains open. In this work, we introduce a set of heuristic algorithms for constructing an optimal net and string path for the unfolded sheet of a given 3D mesh.

Several actuation mechanisms in HCI, robotics, and art make use of actuators to connect disjoint faces and fold them along specific paths. Rivera et al.~\cite{rivera2017stretching} combined 3D-printing with textile-embedded, string-actuated mechanical arms to fold flat structures into 3D shapes. Self-shaping Curved Folding~\cite{tahouni2020self} 3D-printed hygroscopic materials in flat sheets that form curved 3D structures on exposure to moisture. Work on self-folding robots has led to the development of composite laminate materials, fabricated by bonding together layers of different materials that include structural layers, embedded electronics, actuators and flexure hinges, where actuators have ranged from shape memory polymers~\cite{nisser2016feedback} and shape memory alloys~\cite{firouzeh2015robogami} to pneumatically activated polymer “pouches”~\cite{niiyama2015pouch}. These laminates are composed entirely of thin layers with embedded actuators, whose 2D layouts are made using CAD (Computer Aided Design) software and bonded together, with the exception of discrete electronic components that are soldered on to circuit layers that are part of the composite. In work closely related to our own, Kilian et al~\cite{kilian2017string,meenan2008pull} computed and fabricated string actuation networks for a range of well known but limited crease patterns for folded surfaces. Despite the many advances in using origami-inspired folding to acquire target structures and robot geometries, the embedded actuators have to date been difficult to control, and added significantly to the material thickness and complexity; in parallel, unfolding techniques for string-actuated structures have been shown for only few target shapes.

\subsection{Stochastic Self-assembly}

Rather than control assembly procedures using dedicated machines or electronics, researchers have explored building tessellating parts or programmed materials with selectively paired edges, and leveraging the stochastic energy available in the environment to perturb these to find their mating counterparts. To assemble stochastically, modules require pre-programming to enforce correct mating during random collisions with their intended mate. This programmed specificity between pairs of mating faces has been achieved via minimization of free surface energy via topology\cite{hacohen2015meshing}, wettability \cite{bowden1997self}, magnetic forces \cite{lu2021enumeration} or electrostatic \cite{grzybowski2003electrostatic} interaction. In \textit{PullupStructs}, others used an approach based on computational origami that unpacks a 3D mesh into a 2D sheet, before routing a string through laser-cut holes in the sheet, allowing an external, unguided actuator to pull the 2D sheet into a 3D target structure~\cite{niu2023pullupstructs}. 

Self-assembly of discrete parts have been demonstrated both in 2D using shaker tables~\cite{jilek2020centimeter}, as well as in 3D using liquid tanks. Known as fluidic assembly, the assembly problem at the mesoscale has become a particularly widely studied problem in robotics \cite{tolley2008dynamically,tolley2010fluidic,tolley2011programmable,krishnan2008increased,kalontarov2010hydrodynamically, zykov2007experiment}. Existing stochastically self-assembling modules typically include two features to enable assembly: first, embedded magnets that generate near-field forces to bring modules close, and second, selective geometry on module faces that encodes the specificity to only permit bonds between mating pairs \cite{jilek2021towards,hacohen2015meshing,jilek2020centimeter, tsutsumi2007multistate}. However, three key challenges remain for the development of stochastic self-assembling systems: (1) \textit{scalability} that shows how modules can be made both numerous and small; (2) \textit{selectivity guarantees} that help bound module misassembly; and (3) \textit{reconfigurability} that let modules acquire different target shapes. 

\textit{(1) Scalability:} To assemble arbitrarily complex geometries, encodings for 3D modules must support selectivity great enough to permit uniquely mating pairs of modules in the hundreds or even thousands. In addition, modules must be inexpensive and simple enough to be fabricated in these quantities. Due to this dual problem, a significant corpus of previous research demonstrates the stochastic self-assembly for tiled 2D arrays, such as chessboards\cite{grzybowski2003electrostatic,jilek2021towards,miyashita2009influence}, with only two module types where each module is selective to entirely half of all modules in the set. On the other hand, the individual fabrication of heterogeneous module topologies with manually embedded permanent magnets poses a significant challenge to scalable fabrication. 

\textit{(2) Selectivity guarantees:} Because magnet arrangements typically used to generate near-field forces are poorly discriminating to each other, this framework often leads to misassemblies, because near field forces between both mating and non-mating face magnets are equally strong. In addition, protruding geometrical features used for selectivity can lead to "jamming" by obstructing assembly paths \cite{jilek2021towards}, and bounds on the expected misassembly rate between geometrically dissimilar modules may be difficult to compute.

\textit{(3) Reconfigurability:} To date, structures self-assembled at the mesoscale are not reconfigurable. Because module selectivity is achieved by fabricating individualized geometries, any set of fabricated modules encode only a single target shape (or single set, for non-deterministic encodings). Such modules are therefore unable to be "re-programmed" to self-assemble new target shapes: new shapes require a unique batch of modules to be fabricated from scratch, inhibiting their utility and increasing their potential unit cost.

\subsection{Programmable Magnetic Materials}

Advances in digital fabrication tools have enabled researchers to fabricate objects with a wide range of properties by modifying physical parameters, such as the color ~\cite{programmablefilament2020}, surface texture~\cite{hapticprint2015}, compliance~\cite{mueller2013laserorigami,nisser2021laserfactory}, and dissolution rates~\cite{nisser2019sequential} of objects. More recently, digital fabrication tools have also been used to fabricate objects with other functional properties, such as custom acoustic~\cite{printedspeakers2014} and optical~\cite{lenticular2021} behaviors.

Researchers have also investigated how to program magnets and magnetic sheet material for targeted applications. A programmable magnetic sheet is a soft magnetic sheet that can be programmed into a desired magnetic pattern. HCI researchers have demonstrated the great potential of this approach to create custom  tactile, haptic, and tangible interfaces.
For example, \textit{Magnetic Plotter}~\cite{yasu2017magnetic} explored how to fabricate programmable magnetic sheets that can generate various different tactile sensations for haptic interaction. To program the sheet, \textit{Magnetic Plotter} uses a neodymium magnet that is stronger than the sheet, which allows greater flexibility than just using a non-programmed approach such as \textit{Bump Ahead}~\cite{bumpahead2015}.
\textit{FluxPaper}\cite{fluxpaper2015} also explores a magnetically programmable paper, which allows physical movement and dynamic actuation. Moving to more complex patterns, \textit{MagneLayer}~\cite{yasu2020magnelayer} introduces a layered approach, which can create more complex 2D patterns by combining different programmed patterns into a single sheet.
By leveraging these capabilities, \textit{Magnetact}~\cite{magnetact2019} and \textit{Magnetact Animals}~\cite{magnetactanimals2021} demonstrate various applications in kinetic objects or interactive haptic interfaces for touchscreen devices. Finally, Polymagnets~\cite{polymagnets} is a commercial product that leverages dedicated machinery to program individual pixels, creating magnets with unique applications such as non-contact attachment and rotational locking between two objects. However, researchers have not yet investigated how to create a design and fabrication pipeline for programmable magnetic pixels that supports selectively attractive behaviors between \textit{multiple} objects or that is inexpensive and easily reproducible. Since Polymagnets is a commercial product, the fabrication device is not available to researchers and it is unclear how users are supported in creating magnetic pixel layouts of desired behaviors. 

%% file: Laserfactory.tex
\chapter{Multi-process Manufacturing Machines}
\label{sec:Laserfactory}

\begin{figure} [ht]
  \centering
  \includegraphics[width=\textwidth]{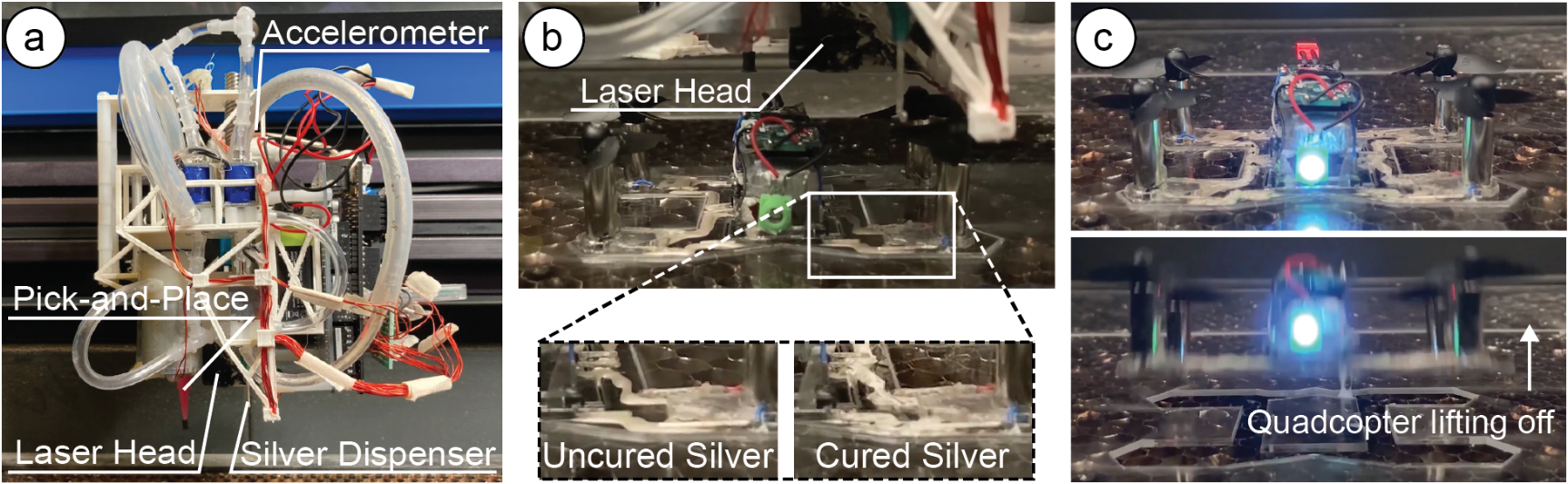}
  \caption{LaserFactory is an integrated fabrication process that creates fully functional devices. (a) Our hardware add-on to an existing laser cutter consists of a silver dispenser and pick-and-place mechanism and allows the machine to not only cut geometry, but also create circuit traces and assemble electronic components. Our accelerometer-based motion classifier enables the add-on to interface with the laser cutter without the need to change the underlying firmware. (b) To cure the deposited silver traces, we developed a laser sintering method that uses the heat of the defocused laser to make the traces conductive. (c) After laser sintering, the fabricated device is fully functional.}
  \label{fig:laserfactory-contributions}
\end{figure}

\begin{figure}[ht]
  \centering
  \includegraphics[width=\textwidth]{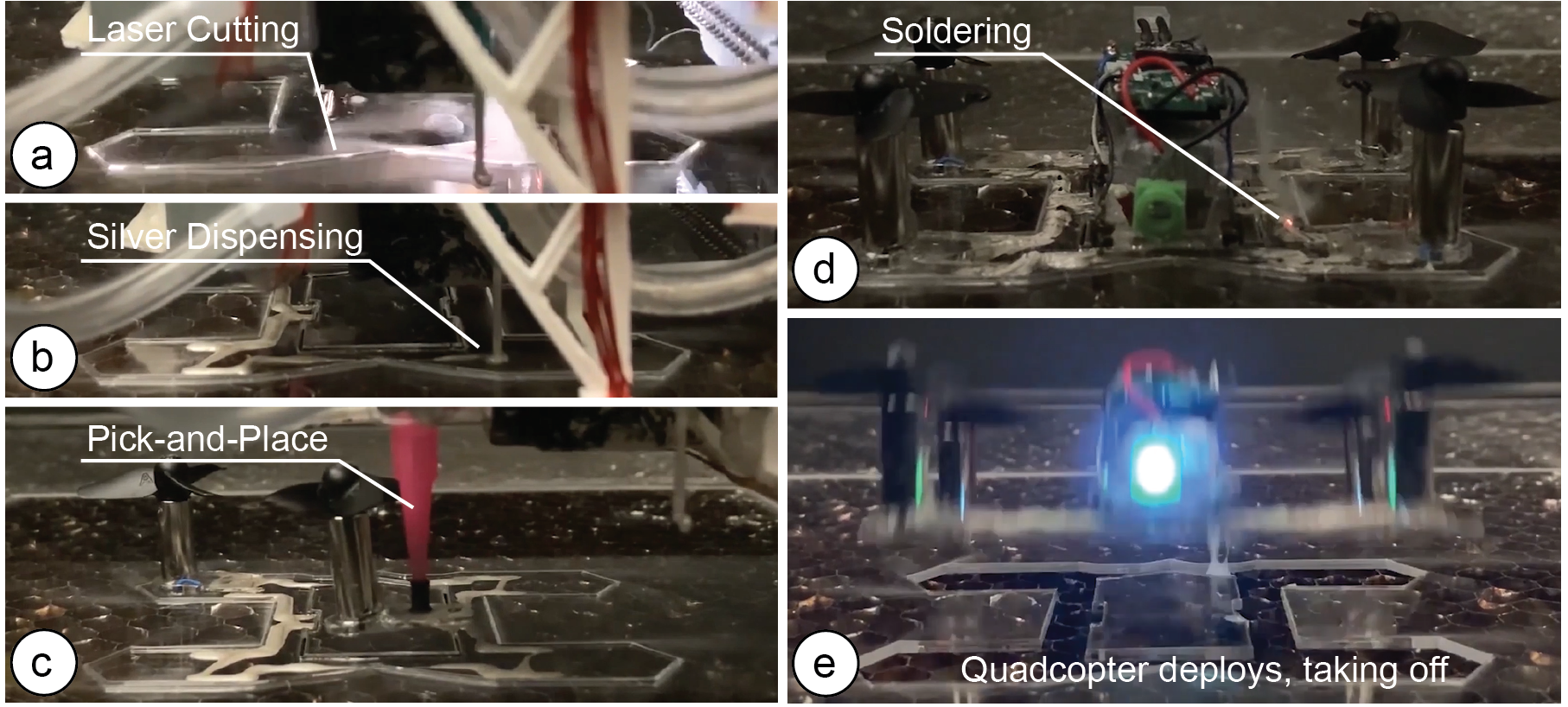}
  \caption{Making a Device with LaserFactory: (a)~Cutting the device geometry, (b)~dispensing silver to form the circuit traces, (c)~picking-and-placing the components, here a quadcopter’s rotor, and (d)~curing the uncured silver traces. (e)~When the last trace is cured, the device is fully functional; here, the quadcopter lifts directly off from the platform.} 
  \label{fig:laserfactory-overview}
\end{figure}

LaserFactory is an integrated fabrication process based on a commercially available laser cutter that has been augmented with a silver dispenser and pick-and-place mechanism to support the manufacture of fully functioning electronic devices without human intervention. Figure~\ref{fig:laserfactory-overview} illustrates the LaserFactory fabrication process: After users download the design file, load it into the standard laser cutter software, and initiate the laser cutter job, LaserFactory first cuts out the device geometry using the regular functionality of the laser cutter (Figure~\ref{fig:laserfactory-overview}a). Next, LaserFactory's silver dispenser deposits silver for the circuit traces (Figure~\ref{fig:laserfactory-overview}b). Following this, LaserFactory's pick-and-place head picks up components from the component storage inside the laser cutter and assembles them onto the circuit traces (Figure~\ref{fig:laserfactory-overview}c). Finally, the laser cures the traces which rigidly connects the components to the circuit (Figure~\ref{fig:laserfactory-overview}d). Once LaserFactory completes these steps, the device is fully functional; here, the quadcopter readily flies off the fabrication platform (Figure~\ref{fig:laserfactory-overview}e). 

In the following sections we describe each of the four main components of this work. We start by describing the hardware add-on that incorporates a silver dispenser and a pick-and-place mechanism. Next we explain the laser sintering technique. After this, we present our motion-based signaling technique and finally we illustrate an end-to-end design and fabrication workflow based on our design and visualization tool coupled with the translation pipeline that converts the design into machine instructions.

\section{Hardware Add-on}

\begin{figure}[ht]
  \centering
  \includegraphics[width=0.9\linewidth]{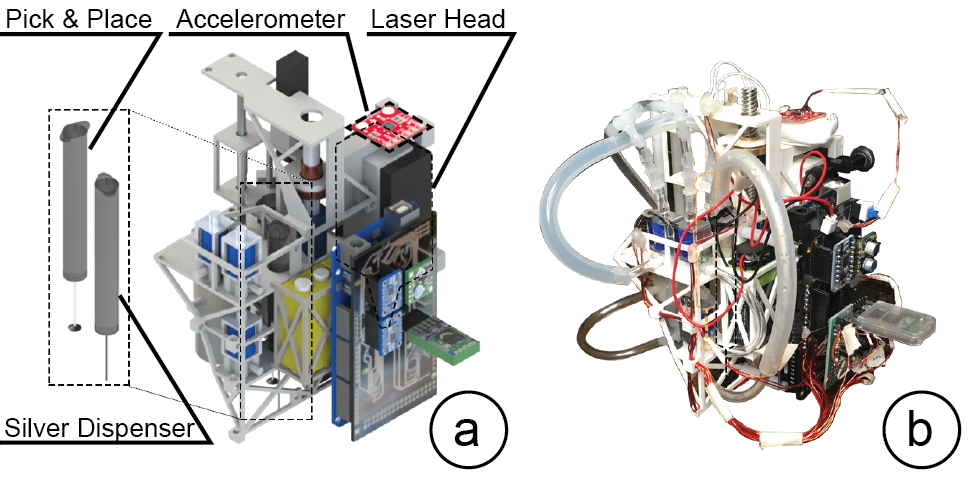}
  \caption{Our hardware add-on shown as (a) a CAD rendering and (b) a photo of the physical device. The add-on consists of a silver dispenser to create circuit traces and a pick-and-place mechanism to assemble electronic components. It attaches to the laser head, which is used to create the device's geometry and to cure the circuit traces. The image also shows the accelerometer used for detecting the motion signaling when the add-on should start/stop its operation.} 
  \label{fig:hardware-addon}
\end{figure}

We developed an add-on for a laser cutter that allows it to fabricate functional electromechanical devices. Our hardware add-on complements the laser cutter's native abilities to create the geometry of a device with circuit creation and assembly capabilities. We accomplish this by adding a silver dispenser and a pick-and-place mechanism to the laser cutter head (Figure~\ref{fig:hardware-addon}).

Our hardware add-on together with the laser cutter's native functionality creates functional devices in the following way: (1) To create the \textit{geometry}, our system uses the regular laser cutter head to cut 2D geometries and optionally bend them into 3D. (2) To add \textit{circuit traces}, our silver dispenser in the form of a syringe extrudes silver paste. It can create highly conductive circuit traces down to (3.2{$\Omega$}/m) sufficient for most electronic components. (3) To \textit{assemble components}, a pick-and-place mechanism consisting of a second syringe on the add-on enables moving components inside the laser cutter. Components are picked from a storage area within the laser cutter platform. Our add-on accommodates lifting components up to a mass of 65g and small SMD components down to size 2010. (4) Finally, we use laser \textit{sintering} (explained below) to cure the traces and make them conductive using the existing laser. The housing of the add-on was 3D-printed using an Ultimaker 3. With all onboard parts, the add-on weights 550g and costs \$150 in parts.  

\subsection{Hardware Implementation}

\textit{Pneumatic Actuation for Dispensing Silver and Picking/Placing Components:} The add-on's silver extrusion and pick-and-place mechanisms are both actuated pneumatically. The silver extruder consists of a 3ml syringe tipped with a 18-gauge needle filled with silver paste. The pick-and-place includes a second 3ml syringe connected to a rubber-tipped 14-gauge needle. For both dispensing and pick-and-place, we use a single pneumatic pump (TCS Electrical, 6V-rated) that supports both mechanisms by regulating air supply via a network of three identical 3-port, 2-position solenoid valves (Phoncoo, 6V-rated).

\textit{Vertical Actuation for Picking-and-Placing Components from the Fabrication Bed:} The pick-and-place syringe must travel vertically during its operation, descending to the fabrication bed to pick/place components, and ascending during traveling to avoid collisions with other components. The pick-and-place syringe is actuated vertically by a NEMA 14 stepper motor (StepperOnline) via a lead screw and rigid coupling, to which it is connected in serial via a spring-loaded linear potentiometer (Sensata 9600). This affords the pick-and-place mechanism the compliance required to avoid damage while allowing the controller software to evaluate the contact forces during component handling. It is also rigidly attached to a second linear potentiometer (BOURNS) which allows sensing of its height and thus distance to the material sheet in real-time. 

\textit{Control Logic of Add-on:} To enable the control logic for the add-on, we mounted a micro-controller (Arduino Mega 2560) onto the add-on and shielded it with a custom PCB that houses the supporting electronics required for the add-on. The supporting electronics include one motor driver (Adafruit DRV8833), one 6V buck regulator (Pololu D24V25F6), and the accelerometer (part of a Sparkfun LSM9DS1 IMU). To switch the pump between the silver dispensing syringe and the pick-and-place syringe, three parallel NPN transistors (2N2222) are used for overcurrent protection, each fitted with 1.5Ω series ballast resistors to avoid thermal runaway.

\subsection{Technical Evaluation of Hardware Add-on}
\label{subsection:HardwareEval}

We evaluated both the silver deposition and pick-and-place capabilities to provide insight into which type of circuits can be created with LaserFactory.  

\textit{Silver deposition trace characteristics:} Since the needle gauge has a significant impact on the minimum trace width, we ran an experiment to determine the smallest needle gauge through which our 6V-rated pump could dispense silver. Smaller needle gauges require higher pressures to dispense the highly viscous silver. We found that by testing needle gauges from 14 to 21, that 18 was the highest gauge through which our 6V-rated pump was able to dispense silver. We then tested depositing this silver two ways. First, by depositing directly onto a material substrate, which facilitates rapid device fabrication but allows silver to spread across the substrate over time, increasing trace widths up to 3mm. As a second slower alternative, we engraved 0.75mm channels using a defocused laser and dispensed silver into these, yielding a minimum trace width of 0.75mm with our current setup. For the future, this trace width could be further reduced by using a stronger pump and narrower nozzle; Valentine et al. \cite{valentine2017hybrid} for instance, use a 0.2mm nozzle with their platform. 

After determining the minimum trace width, we also determined the minimum distance between two adjacent traces, which is important since it determines the pin spacing and thus the size of the components that can be used. To find the minimum trace distance between adjacent traces, we deposited and sintered parallel traces spaced between 0.4mm and 1.8mm apart in 0.2 mm increments. We found that the minimum distance was 0.8mm between traces for silver that had been appropriately refrigerated before use, as the silver's high viscosity and surface tension prevents it from spreading after it has been dispensed. However, for silver left at room temperature, 1.5mm was the minimum distance achievable with our chosen needle gauge. Smaller distances may cause short-circuits between adjacent traces due to the silver spreading across the substrate or ballooning slightly during laser curing. In addition, while our silver dispenser negatively pressurizes to de-ooze before moving between traces, it is not actuated in the z-axis and can therefore not retract from the substrate, leading to silver occasionally being spread between traces. This, however, does not cause short circuits or impact the device functionally since silver only becomes conductive where addressed by the laser which only targets the actual traces, thus the remaining silver can be wiped off. For future work, we plan to make the dispenser retractable to avoid additional spread. Based on these results, we conclude that we can currently create traces fine enough to work with SMD components down to a 1.55 mm pin spacing in ideal conditions where the silver was refrigerated before use, and 2.3mm when silver is at room temperature, which can be used for components with standard pin spacings of 2.54mm.

\textit{Pick and place component criteria:} We evaluated the criteria under which LaserFactory can pick and place components. First, components require a minimum flat surface area of 3x3mm above the component’s center of mass for the suction tip's 2.5mm diameter nozzle to form a good contact. We found this could be increased or decreased by increasing or decreasing the diameter of the pick-and-place nozzle, respectively. Second, components must have a mass smaller than 65g for the pump suction to lift them. This threshold allows lifting large components, such as an Arduino Mega micro-controller (37g). A stronger pump could further raise this threshold. Third, their height cannot exceed 27 mm, currently limited by the maximum extension of the linear potentiometer used to assess picking height. Given these constraints, LaserFactory can pick large, heavy components, such as typical 3V batteries; tall components such as rotors; and a variety of small components, such as SMD resistors down to size 2010. For future work, these constraints can be further lifted by using a smaller nozzle, larger pump, and longer lead screw.

\textit{Use over time:} During the course of fabrication, the peak power draw measured from the two on-board parallel 9V lithium ion batteries was 13.5W. Between no-load cases of cutting/folding and variable-load cases of PCB-making, this is sufficient for approximately 4 hours of continuous device creation. We detected no irregularities in laser carriage motion that would indicate undue stress on the laser cutter during use.

\section{Laser sintering: Curing Circuit Traces using a CO2 laser}

When both the silver paste has been deposited and the components have been placed onto a substrate in the laser cutter, the silver is not yet conductive, nor does it rigidly bond components to the substrate. For this, the silver requires curing via heating, traditionally achieved via baking the created device in an oven before it becomes functional. However, this has several drawbacks, such as that the heat in the oven can damage the material substrate and the electronic components due to the thermal stresses. To solve this issue, we developed a technique that uses the laser to cure the circuit traces and connect components electrically. A benefit of using the laser is that it allows for a fully integrated fabrication process with no manual work required. In addition, heat is only applied locally to traces and thus affects the material sheet only in the locations of the trace. Finally, using the laser is fast, approximately 5 minutes in contrast to the 2 hours achieved via uniform heating in an oven ~\cite{valentine2017hybrid}. Once the last circuit trace is cured, the device is functional right off the fabrication bed.

\subsection{Fabrication Settings for Laser Sintering}

We ran a set of experiments to determine the best power, speed, and z-axis settings for curing silver traces. For the experiments, we used a Universal Laser Systems ULS.PLS.150D laser cutter equipped with a 2.0 inch lens. For the silver, we used the silver paste from SunChemical (C2180423D2), which has a composition of 56.69\% 24K silver and 43.31\% a blend of resin and solvents.

\textit{Power/Speed/Height Settings:} To find admissible laser cutter settings that successfully cured the silver, we dispensed a series of 100mm silver traces, then sintered them with different parameters. We first calibrated the vertical offset that produced a defocused laser spot diameter equal to trace width of silver. We then irradiated the silver using permutations of power and speed settings on the interval 0.2\% to 90\%, recording the resistance of the traces using a Greenlee DM-820A Multimeter and recording whether it was viscous (uncured), solid (cured), or burnt (overcured) to touch. We found numerous admissible power-speed combinations producing rigid, conductive traces. These are bounded by the combination of 0.2\% speed with 7\% power on the lower bound, and the combination of 7\% speed with 90\% power on the upper bound. Linear interpolation between these lower and upper bounds yield further admissible settings, showing that lower powers require commensurately lower speeds to effectively cure the silver. Power/speed ratios greater than this produce charred, brittle traces, while power/speed ratios lower than this leave the silver poorly conductive and uncured. In addition, high speeds above 7\% produce traces that remain subcutaneously uncured regardless of power, while powers below 1\% do not elevate temperatures sufficiently to cure traces at all.

\subsection{Technical Evaluation of Laser Sintering}

To determine the conductivity of traces cured via laser sintering and the sintering connection quality between traces and components, we ran the following technical evaluation. 

\textit{Conductivity:} To evaluate the conductivity that laser sintering achieves in a single pass of the laser, we fabricated 8 traces of length 0.5m and with a minimum trace width of 0.8mm. we then measured the resistance between the traces’ ends using a Multimeter (Greenlee DM-820A). The resistance was normally distributed about 3.2{$\Omega$}/m with a standard deviation of 0.0005. LaserFactory thus supports a range of electronic components, actuators, I2C communication, PWM, and medium to high-power applications.

\textit{Power and frequency:} We used a current-limited power supply to pass DC current through five 100mm traces at 1A, 2A and 3A power without observing arcing or warming. In addition, we used a frequency generator to deliver an AC signal from 0Hz to 100kHz in logarithmic increments. Investigating the signal with an oscilloscope showed there was no significant signal attenuation. 

\textit{Sintering connection quality of traces and components:}
To test how strong the cured connections between the traces and the components are, we evaluated their mechanical rigidity. We used a two-pad component and hung weights to the face of the component to apply normal forces between 25g and 300g in increments of 25g. We found that a normal force of 225g was required to disconnect the component from the trace. The strength of the curing connection is considerably less than that of a traditional sinter connection. While this can be a drawback for devices that undergo high mechanical strain, it also has the benefit that components can be re-used, as users can pull them off with reasonable force. 

\textit{Warping:} Warping is a form of material distortion that can occur during laser cutting, and can arise from laser sintering too. This occurs due to rapid local temperature changes from selective heating of areas being targeted. The amount of warping is affected by both the choice of material and the part size, as heat cannot dissipate out of smaller parts via conduction. 
To explore this, we tested its effect on three materials: extruded acrylic, and two cast acrylics (unbranded plexiglass). All were of equal transparency and thickness (1.5mm). While we observed no warping for the cast acrylic, we found that extruded acrylic did exhibit warping, and we calibrated our laser cutting settings to minimize its effects. Since some of our devices are made from extruded acrylic, this warping can be observed as a result of the laser sintering process in the accompanying video. While this distortion can impact both the visual appeal of devices as well as their operation if their function relies on precise planar surfaces, the effects can be largely avoided by using one of the alternative materials. In addition, we found that warping can be further reduced by cutting the device out in the final stage in the fabrication process instead of at the beginning, allowing the surrounding material to hold it in place during sintering. Alternatively, connector tabs can be left on the cut outline to achieve the same effect, i.e. keeping the device attached to the sheet during sintering, and cutting these tabs to release the device at the end.

\textit{Materials:} We tested LaserFactory's compatibility with a range of materials, including extruded acrylic, cast acrylic, leather, cardboard, hardwood (oak) and plywood. Extruded and cast acrylics were compatible with the full LaserFactory suite, including the engraving of trace channels (\ref{subsection:HardwareEval}) and bending of the substrate (\ref{subsection:Laser-non-planar}) to create 2.5D geometries; however, extruded acrylic was prone to warping. The woods and and cardboard were also tested successfully, although these char visibly during sintering if the laser contacts the material directly. Furthermore, while not susceptible to warping, they cannot support bending or the engraving of trace channels. Lastly, leather and other soft textiles were not able to support the silver traces due to the traces cracking upon being bent after curing. This, however, can be addressed by attaching a firm backing to the soft material. 

In summary, we conclude that laser sintering using a CO2 laser is suitable for creating highly conductive traces, that the sintering connection holds components in place while allowing components to be removed using manual force, and that warping due to laser-induced local heating only affects certain materials significantly.

\subsection{Creating Non-planar Geometries} 
\label{subsection:Laser-non-planar}

In this section, we report on two proof-of-concepts that showcase different methods of creating non-planar geometries using LaserFactory.

\begin{figure}[ht]
\centering
\includegraphics[width=0.95\linewidth]{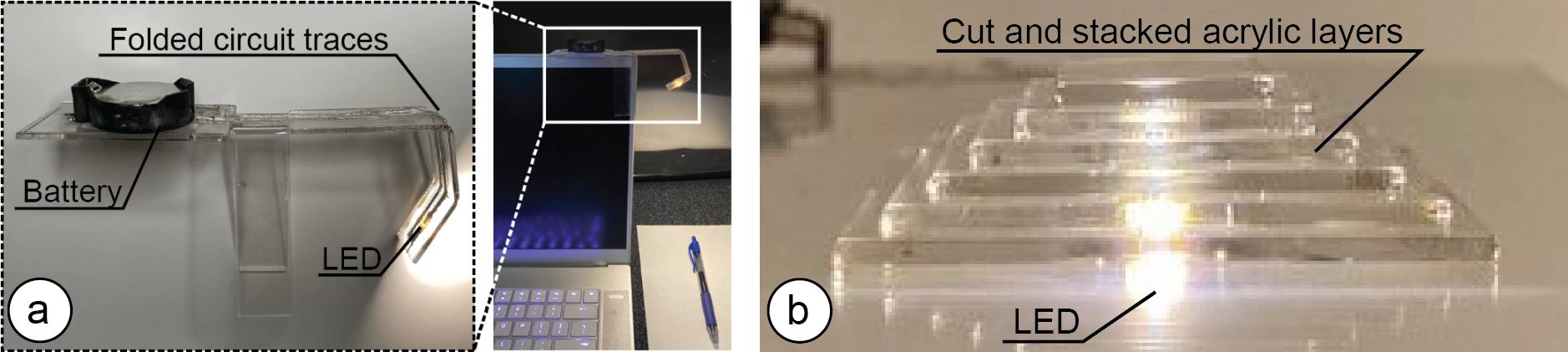}
\caption{LaserFactory can be used to create (a) 2.5D geometries via folding, (b) as well as discretized 3D geometries via cutting and pick-and-placing the material substrate itself.}
\label{fig:geometry}
\end{figure}

\textit{Laser Sintering Folded 2.5D Circuit Traces:} Since our technique is built based on a laser cutter, which has been shown to not only be able to create 2D geometries but also 3D geometries via folding (LaserOrigami~\cite{mueller2013laserorigami}), we also investigated if we can dispense and cure circuit traces that are positioned across a fold, which would allow LaserFactory to fabricate 2.5D geometries with embedded traces and electronics (Figure~\ref{fig:geometry}a). Our early experiments showed that by using the laser to slowly heat traces across a hinge, the silver remains viscous and uncured, which allows the substrate to become compliant and fold under its own gravity. By continuing the heating process after folding, the silver gradually cures and hardens, producing 2.5D geometries with embedded traces. While we did not evaluate the different power/speed settings in a controlled experiment, we did not observe a difference in resistance between folded and planar traces. We include the associated laser settings in Table~\ref{tab:laser-settings}.

\textit{Cutting and stacking layerwise 3D structures:} In addition to folding 2.5D traces, we report on a second method of creating 3D geometries that leverages the unique combination of our pick-and-place mechanism with a laser cutter. In addition to placing electrical components on a device, we can also pick-and-place pieces of the material substrate. To harvest these pieces, shapes are first cut from the substrate and then the pick-and-place is used to stack them to create 3D structures (Figure~\ref{fig:geometry}b). By discretizing a structure into layers, these layers can be cut using a focused laser, stacked via pick-and-place, and welded using a defocused laser~\cite{umapathi2015laserstacker} to produce rigid 3D structures without manual intervention.

\section{Augmenting a Laser Cutter Through Motion-Based Signaling}

Augmenting existing fabrication devices reduces engineering effort by piggy-backing existing platforms rather than building hardware/software from scratch, and by making the new functionality available to a wide range of users who already own or operate the platform. However, interfacing with a commercial fabrication platform that is not built to accommodate an expansion of its functionality is challenging~\cite{valentine2017hybrid}. In building our hardware augmentation to a laser head, we investigated solutions that do not require access to the proprietary software of the fabrication platform and that therefore could potentially be platform-agnostic to permit augmenting a range of different fabrication devices. To this end, our method involves two primary stages. First, our method embeds additional lines in the fabrication file. These include additional paths such as trajectories for navigating between electrical components, and also specific motion patterns that signal instructions to the add-on. Second, our method detects these motion patterns using a motion-based classifer based on accelerometer data running locally on the add-on. Our method to transmit fabrication instructions thereby relies only on the fabrication platform's ability to execute motion in the X-Y plane; the add-on has no explicit knowledge of the contents of the fabrication file itself. In the next section, we provide more details on how we create the additional lines to produce carriage motion of the laser cutter and how we signal the transition between different fabrication stages using the motion patterns in the design file.

\subsection{Creating Carriage Motion by Embedding Additional Lines in the Design File}

\begin{figure}[ht]
  \centering
  \includegraphics[width=0.95\linewidth]{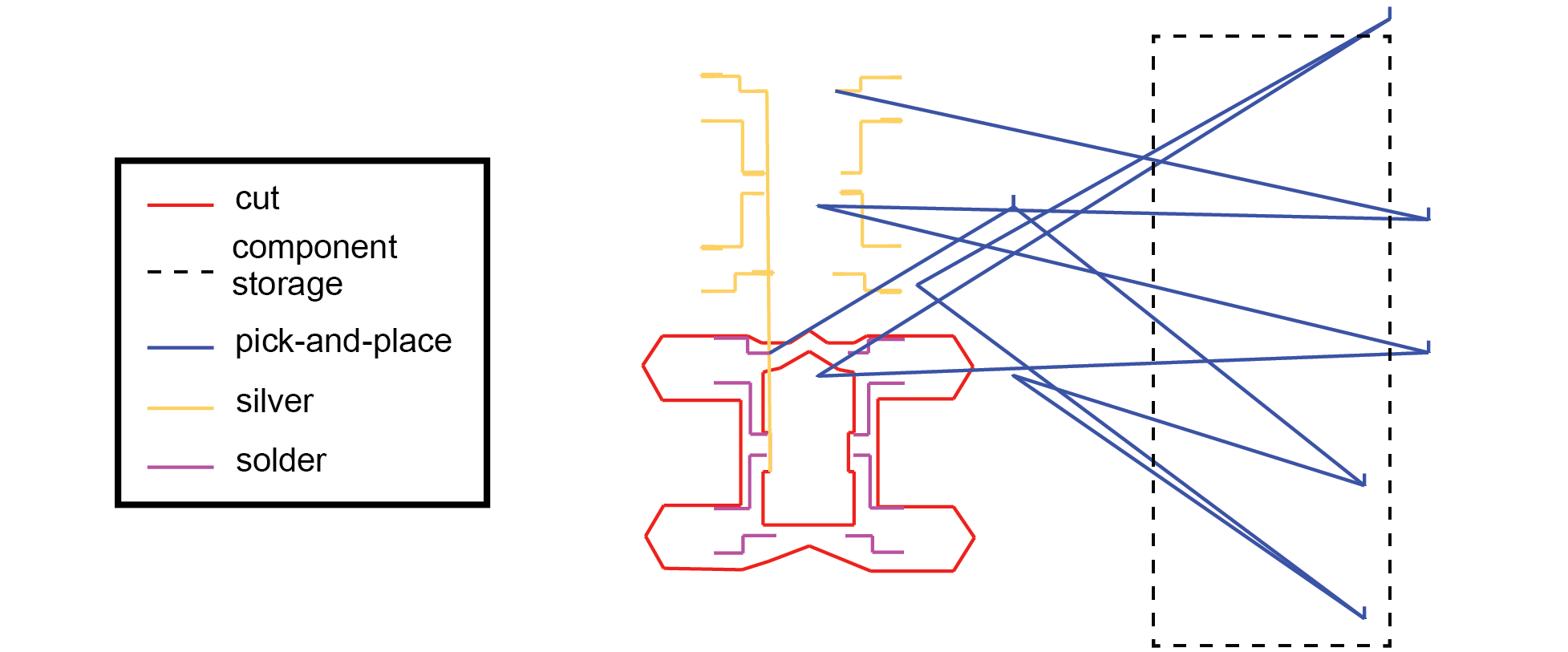}
  \caption{Adding lines to the design file to create additional carriage motion for pick-and-place operations, silver dispensing, and sintering. Lines are offset to compensate for each add-on part's position with respect to the laser head.} 
  \label{fig:motion-lines-1}
\end{figure}

To generate carriage motion without accessing the laser cutter's firmware, we add extra lines into the design file that is sent to the laser cutter. Figure~\ref{fig:motion-lines-1} illustrates this using a quadcopter design file that has additional lines embedded for dispensing silver, pick and place, and sintering. The lines are offset in the design file to compensate for the add-on's position with respect to the laser head. The original design file as created by a user can be seen in Figure~\ref{fig:design-tool}g.

\subsection{Signaling Fabrication Instructions using Motion Patterns in the Design File}

After additional lines have been embedded to create trajectories that guide the laser head to execute silver dispensing and pick-and-place manoeuvres, a separate step involves relaying fabrication instructions to the add-on (when to start and stop silver dispensing and pick-and-place maneuvers) without communication with the laser cutter. This involves first embedding additional lines that encode these fabrication instructions, and then interpreting these using sensors. In choosing sensors, we disqualified candidates that could not be entirely integrated into the add-on, such as external position tracking. We also disqualified candidates that may introduce platform-specific dependencies, such as onboard cameras or position sensors that measure the inside volume of the fabrication platform. 

On the other hand, an accelerometer affixed to the laser head measures only its inertial motion in terms of acceleration, which makes it possible to recreate on other platforms. By appending a unique line pattern to the beginning and end of paths in the fabrication file, we elicit a specific motion from the laser head, which the onboard accelerometer interprets in order to start or stop a fabrication procedure (silver deposition or pick-and-place).

\textit{Motion Pattern Shape:} To select a motion pattern for our signal, we investigated the motion of different shapes. In particular, we considered the motion of straight lines, squares and circles in order to test independent, sequential, and simultaneous motions in the X-Y plane, respectively. While all candidates yielded a uniquely classifiable signal, we found that squares and circles were unsuitable as performing their motions may cause extrusion of silver outside of a trace or collision with other components. A one-dimensional line that programmatically tracks backward along a previously deposited trace, in contrast, provides a trajectory that keeps the tip and any dispensed silver above traces.

\begin{figure}[ht]
  \centering
  \includegraphics[width=0.99\linewidth]{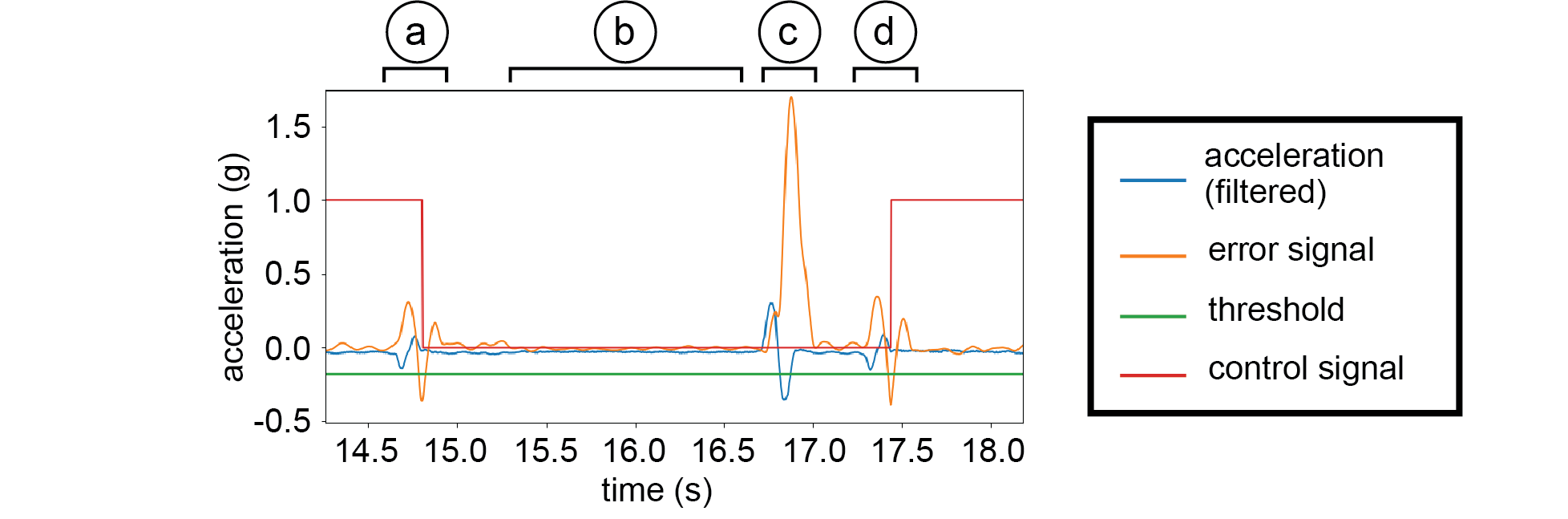}
  \caption{Filtered accelerometer data during execution of the embedded pattern. When the pattern is executed, the (blue) acceleration produces oppositely signed spikes in rapid succession. (a,d) This characterized template causes the (orange) error signal to drop below the (green) threshold which flips the control signal to start/stop a command; here, starting/stopping the silver extrusion. (b) Periods of constant velocity are marked by accelerations around 0; (c) volatile maneuvers such as moving between contiguous traces causes large errors and thus can be differentiated from actual motion-signal instructions embedded in the fabrication file.}
  \label{fig:imu-signal}
\end{figure}

\textit{Duration of Line Pattern:} The motion pattern's duration, and therefore physical length, must be small enough to be quickly evaluated by the classifier to allow rapid transmission of fabrication instructions yet long enough for the digital sampling to produce sufficiently many data points for the classifier. We tested the execution of straight lines of lengths between 0.5mm - 6mm in 0.5mm increments, measuring the duration of each. For our processor's maximum sampling speed of 300 Hz, 3mm lines constituted the fastest signal (150ms) that had a sufficient number of data points (45 samples) to encode a uniquely identifiable signal. By accelerating along the trace for 3mm, then decelerating to stop, the signal elicits two consecutive equal but opposite spikes on the accelerometer. This creates a unique signal, as no other maneuver requires re-tracing the exact same path, that can be reliably classified using the accelerometer (Figure~\ref{fig:imu-signal}). 
% (Figure~\ref{fig:motion-lines}).

\begin{table*}[ht]
\caption{Fabrication settings. Columns, from left to right, show (a) process order (b) line manipulations, (c) laser cutter settings, (d) add-on settings for the silver dispenser and pick-and-place mechanism. A tilde indicates varying a value.}
\label{tab:laser-settings}
\begin{adjustbox}{width=\textwidth}
\begingroup
\fontsize{7pt}{7pt}\selectfont
\begin{tabular}{|ccc|cc|cccc|ccccc|}
\toprule
\multicolumn{3}{|c|}{\textbf{Fabrication   Process}} & \multicolumn{2}{c|}{\textbf{Line Manipulation}} & \multicolumn{4}{c|}{\textbf{Laser Cutter Settings}} & \multicolumn{5}{c|}{\textbf{LaserFactory Hardware Settings}} \\
\midrule
Step              & Primitive                      & Mandatory? & Offset (X/Y mm)            & Passes            & Color    & Power (\%)  & Speed (\%)  & Height (mm) & Pump    & Stepper    & Valve 1    & Valve 2    & Valve 3    \\
\midrule
\#1a              & Cut outline                    & Yes        & (0,0)                      & 1                 & Red      & 80          & 30          & 5           & OFF     & OFF        & OFF        & OFF        & OFF        \\
\#1b              & Raster footprints              & No         & (0,0)                      & 1                 & Black    & 30          & 30          & 5           & OFF     & OFF        & OFF        & OFF        & OFF        \\
\#1c              & Engrave channels               & No         & (0.4,0.7)                  & 1                 & Green    & 70          & 25          & 22          & OFF     & OFF        & OFF        & OFF        & OFF        \\
\#2a              & Dispense silver                & Yes        & (-1.8,-68)                 & 1                 & Yellow   & 0           & 1.5         & 0.8         & ON      & OFF        & OFF        & ON         & ON         \\
\#2b              & Signal/De-ooze                 & Yes        & (-1.8,-68)                 & 1                 & Yellow   & 0           & 1.5         & 0.8         & ON      & OFF        & ON         & ON         & OFF        \\
\#3a              & Pick                           & Yes        & (1.5,3)                    & 1                 & Blue     & 0           & 1.5         & $\sim$      & ON      & $\sim$     & OFF        & OFF        & ON         \\
\#3b              & Carry component                & Yes        & (1.5,3)                    & 1                 & Blue     & 0           & 1.5         & 28          & ON      & OFF        & OFF        & OFF        & ON         \\
\#3c              & Place                          & Yes        & (1.5,3)                    & 1                 & Blue     & 0           & 1.5         & $\sim$      & ON      & $\sim$     & ON         & OFF        & OFF        \\
\#4               & Cure silver                    & Yes        & (1.5,3)                    & 1                 & Magenta  & 7           & 1           & 68          & OFF     & OFF        & OFF        & OFF        & OFF        \\
\#5a              & Fold hinge                     & No         & (0,0)                      & 60                & Cyan   & 30          & 100         & 68          & OFF     & OFF        & OFF        & OFF        & OFF        \\
\#5b              & Cut 2nd outline                & No         & (0,0)                      & 1                 & Orange     & 30          & 100         & 5           & OFF     & OFF        & OFF        & OFF        & OFF       \\
\bottomrule
\end{tabular}
\endgroup
\end{adjustbox}
\end{table*}

\subsection{One-time signal calibration and Real-Time Classification}

To build the classifier, we perform a one-time calibration of the 3mm line signal described above. We perform this calibration by executing the pattern and measuring the corresponding linear accelerations from the accelerometer in the X-Y plane. We sample the pattern at 300 Hz and low-pass filter it using a 24-point averaging filter to remove noise to construct the signal template.

\begin{figure}[ht]
  \centering
  \includegraphics[width=0.99\linewidth]{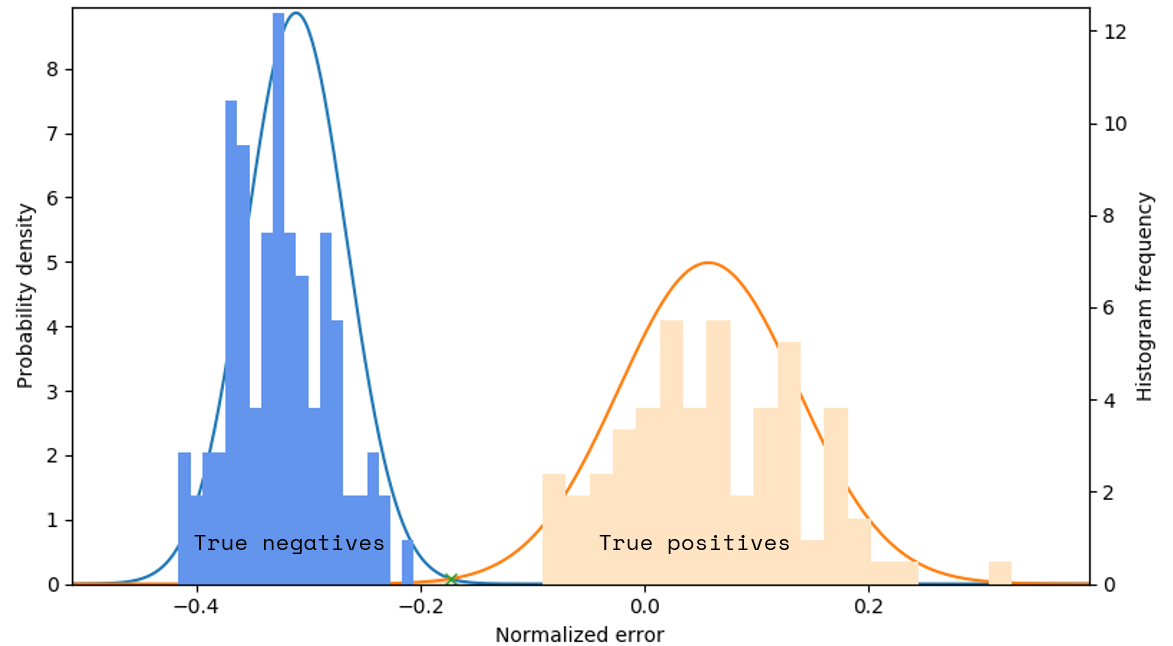}
  \caption{Threshold characterization for the signal classifier. Signals are sampled to build a dataset of true positives, and other accelerations are sampled to build a dataset of true negatives. Modeling these as normal distributions, we place a threshold at their intersection to differentiate between signal and noise.
}
  \label{fig:roc}
\end{figure}

To choose the error signal threshold to identify a signal (Fig. \ref{fig:roc}), we built a dataset of true positives by sampling the signal pattern 100 times, recording the maximum negative error produced by the matched filter for each. Equivalently, we built a dataset of true negatives, compiling a further 100 measurements taken during manoeuvres not associated with a signal. Modeling each as a normal distribution $\mathcal{N}(\mu,\sigma^2)$, we found the mean and standard deviation for the true positives (X) and true negatives (Z) were $X\sim \mathcal{N}$(-0.311,0.002) and Z$\sim \mathcal{N}$(0.057,0.008)  respectively and choose our threshold at the intersection of these curves (-0.18) in order to minimize misfirings (false positives and negatives).

After the one-time calibration, our add-on is ready for real-time detection of the motion signals embedded in the design file. For the real-time detection, we low-pass filter the raw acceleration data using a 24-point averaging filter and save it into a 45-point circular buffer. We then build a matched filter by evaluating this buffer against the signal template using the L2 norm at 300 Hz. This produces an error signal of the low-pass filtered data that our detection algorithm normalizes around 0. We then continuously evaluate the output of the matched filter and signal an instruction to the silver extruder or pick-and-place mechanism when the error signal surpasses a characterized threshold.

\subsection{Technical Evaluation of the Motion-Based Classifier}

We evaluate both how reliable our motion-based classifier can detect the motion-signals and if the motion-based classification approach transfers across different fabrication machines. 

\textit{Performance:} First, we evaluated the accuracy with which our motion-based classifier detects signals embedded in the fabrication file. To determine this, we mounted the add-on to the laser head, then started fabrication of a design consisting of 5 horizontal lines and 5 vertical lines, each 50mm in length, totalling a combined 10 start signals and 10 stop signals (20 signals in total). For each signal, we noted whether detection was successful, and after all lines were executed, we dismounted the add-on. We repeated this procedure 10 times, for a total of 200 signals. One signal was undetected, yielding a classification accuracy of 99.5\% for this test, with 0 false positives, and 1 false negative. We then executed fabrication of our three application examples; the quadcopter, wristband, and PCB, five times each. These consist of a total of 465 signals, including 305 silver deposition sequences between 7mm and 90mm in length, and 115 pick-and-place sequences between 140 and 190mm in length. No signal misfirings were registered during these trials. By integrating the cumulative distribution functions of our normal distributions to evaluate the Receiver Operating Characteristic (ROC) curve for our characterized classification threshold, we find that our model predicts a false positive rate of approximately 0.2\% and false negative rate of 0.1\%, which agrees with these results. This result shows that our procedure produces a reliable classification due to our ability to freely design a unique signal for detection. However, in case the detection misfires, LaserFactory currently has no way to detect it. We plan to explore such detection as part of future work by embedding additional lines that function as interrupts.

\textit{Transferability:} We designed the hardware add-on to be fully self-contained and the fabrication signaling to be motion-based in order to demonstrate an add-on design that can be used to augment different fabrication machines. To test if our add-on can indeed work on different fabrication machines, we tested our approach on the 3D printer Ultimaker 3. To conduct this test, we wrote a custom G-code script that mimicks silver deposition by executing first a vertical then a horizontal 30mm line, with our 3mm signal pattern appended at every line end. We use the G0 command to execute the pattern, representing a co-ordinated movement at rapid rate without material extrusion with feed rate set to F15000. We deployed this file 10 times on the Ultimaker 3 and achieved a 100\% signal classification accuracy. While additional tests on other fabrication devices are needed before full transferability can be claimed, the results show that motion-based signaling has the potential to work across different fabrication devices. Before motion-based signaling can be used on a new machine, the motion-signal must first be characterized via the one-time calibration and the add-on's physical clips need to be adjusted to the differently shaped tool head.

\section{Designing a Device for Use with LaserFactory}

\begin{figure*}[ht]
  \centering
  \includegraphics[width=\textwidth]{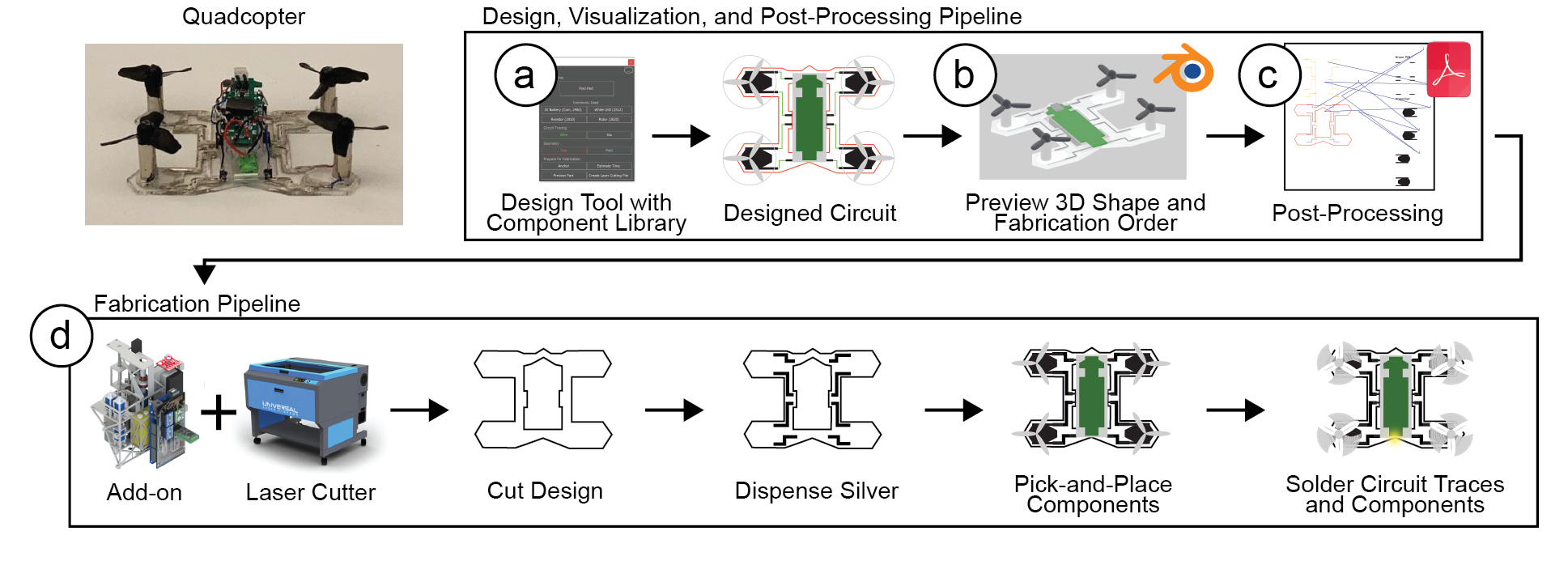}
  \caption{LaserFactory Pipeline: (a) Design Tool: Users place components, and draw geometry and circuit traces. (b) Visualization Tool: Users can preview the design in 3D and visualize the fabrication steps for debugging the design. (c) Post-Processing: On export, the design file is converted into machine instructions for the augmented laser cutter. (d) Fabrication: The augmented laser cutter then executes the post-processed file, i.e. cuts the geometry, dispenses silver for circuit traces, pick-and-places components, and then cures the silver to make the traces conductive and sinters the components. Once fabricated, the device is fully functional.}
  \label{fig:pipeline}
\end{figure*}

To support the creation of devices with LaserFactory, we provide the end-to-end fabrication pipeline shown in Figure~\ref{fig:pipeline}. It consists of a design and visualization tool as well as a post-processing script that converts the design file into a set of machine instructions for fabricating the device with LaserFactory on the augmented laser cutter. The design tool lets users create 2D designs using drawing tools for the geometry (‘cut’, ‘fold’), the circuit traces (‘wire’), and for placing electronic components (‘place part’). This allows users to concurrently design the geometry and the electronic circuit. The accompanying visualization tool renders both the final 2D or 3D design and allows users to animate each step of the fabrication process. This assists users who may find it difficult to visualize the abstract 2D drawing. In addition, it allows users to inspect the order of each step in the fabrication process. On export, our post-processing script automatically translates the design file into machine instructions. Users do not have to add any explicit instructions for the add-on.

\subsection{Design Tool}

\begin{figure}[ht]
\centering
\includegraphics[width=0.95\linewidth]{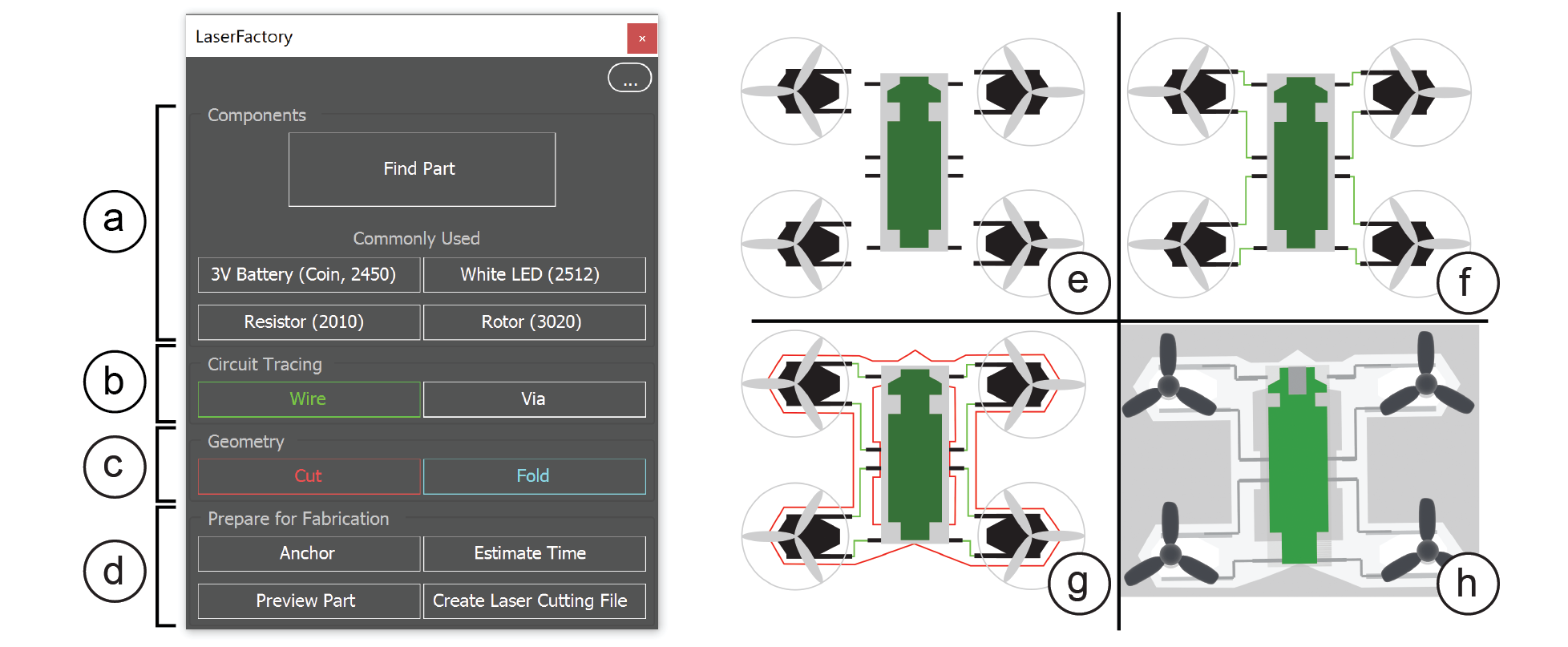}
\caption{Users use the design toolbar to design a device such as a quadcopter. The steps involved (a,e) placing components from a part library, (b,f) routing circuit traces, (c,g) drawing lines for the geometry, here showing cut lines and (d,h) using export tools, here showing rendering the output of the visualization tool.}
\label{fig:design-tool}
\end{figure}

Our design tool (developed as a plugin to the 2D editor Adobe Illustrator) supports users in placing components, drawing circuit traces and designing the geometry of their devices. Figure~\ref{fig:design-tool}a shows the toolbar of our design tool, which has the following functionality.

\textit{Electronic component library:} To facilitate circuit design, users can select electronic components from a library of parts. Common components can be accessed directly from the toolbar and others can be found using the 'find part' button that opens the component library (Figure~\ref{fig:design-tool}a). After selecting a component, it is loaded as a 2D representation onto the canvas (Figure~\ref{fig:design-tool}b). Components appear as their physical footprint superimposed with their electronic symbol. Seeing the electronic components allows users to appraise the size of the main geometry in their design and adjust the design as needed.  

\textit{Wire drawing tool:} LaserFactory offers a wire drawing tool (Figure~\ref{fig:design-tool}b), which is used to connect electronic components with traces. While free-form traces are possible, we enforce the wire tool to snap to 0/90° in order to align signaling motion with trace geometry for cleaner fabrication. Traces can also snap to the electrodes of components on approach to them, ensuring good electrical connections during fabrication. In the drawing, circuit traces are differentiated from other types of lines by color coding them in green (Figure~\ref{fig:design-tool}g).

\textit{Geometry drawing tools:} The LaserFactory user interface offers two drawing tools to define the geometry of the object (Figure~\ref{fig:design-tool}c): the cut drawing tool and the bend drawing tool. In the drawing, cut lines are visualized in red and bend lines are visualized in cyan. For bending, users only have to indicate where to bend, but not in which order, which is taken care of by our post-processing script. For bending, users must use the 'anchor' tool to assign a section as the anchor plane which constrains this section as the one that remains planar in the event of folding. 

\textit{Export tools:} Finally, the interface includes a set of tools for use on completion of the design (Figure~\ref{fig:design-tool}d). The "estimate time" tool parses the file to estimate the total fabrication time that is then displayed in the view. The 'Preview Part' tool opens the visualization tool, which we describe in more detail in the next section. Finally, the 'create laser cutting file' tool post-processes the design for fabrication and generates the final fabrication file.

\subsection{Visualization Tool}

\begin{figure}[ht]
  \centering
  \includegraphics[width=0.85\linewidth]{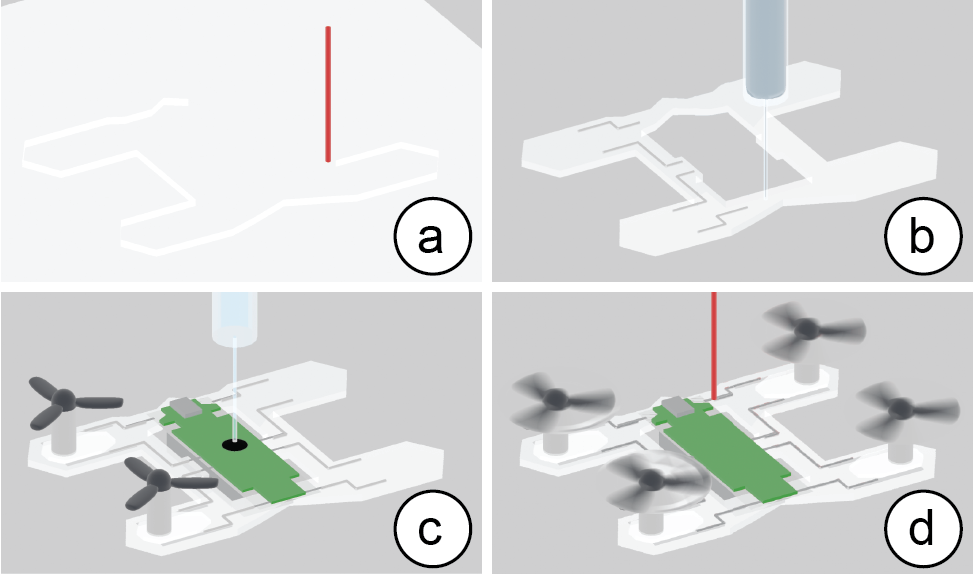}
  \caption{Visualization showing the resulting 3D geometry and order of fabrication steps: (a) cutting, (b) silver deposition, (c) component placement, (d) sintering and deployment.}
  \label{fig:visualization-tool}
\end{figure}

Since the 2D drawing is an abstract representation of the resulting design, we developed a 3D visualization tool (developed as a plugin to the 3D editor Blender) that users can launch at any time in the design process. Figure~\ref{fig:visualization-tool} shows the visualization tool, which has the following functionality.

\textit{Rendering the 3D design:} The visualization tool shows the design as a solid 3D geometry, including any folded parts. This allows the user to see if the folds are placed correctly and result in the desired 3D geometry. In addition, all electronic component footprints are replaced with their respective 3D models, providing a preview of the final device. 

\textit{Order of fabrication steps:} Besides showing the design as a 3D shape, the visualization tool also contains a video playback that animates each step of the fabrication process. This allow users to double check if the post-processed design file contains the correct fabrication order. For instance, folded geometry must be released with a cut first before any folding can occur, which can be observed in the animation. 

\textit{Visualizing the movement of the tool head:} Finally, the visualization tool also renders the tool used in each fabrication step: the laser during cutting and curing (and any folding), the silver dispenser during trace creation, and the pick-and-place nozzle during component placement. This allows users to assess potential collision in their design, such as unwanted intersections between the laser beam and placed components to prevent damage.

\subsection{Post-processing of Design File}

When exporting the design file, a number of post-processing steps are applied to the design file to make it work with the LaserFactory add-on. The user’s exported design contains only the cut and bend lines, the circuit traces, and the locations of components on the device. In a first post-processing step, additional motion lines are added for picking-and-placing components and curing the traces. Next, all lines that belong to the silver dispenser and pick-and-place head are offset by each add-on's distance to the laser head. In addition, all curing lines are offset by a small amount to account for the laser beams offset at larger heights. After this, the pre-designed motion pattern that encodes the start/stop signal for the silver dispenser and the pick-and-place mechanism are added to the drawing. Finally, all lines are color-coded to ensure proper fabrication order. All line-manipulations are summarized in Table~\ref{tab:laser-settings}.

\textit{Adding lines for sintering and pick-and-place trajectories:} The original design file does not contain lines for picking up components from the component storage on the laser cutter's periphery and placing them onto the device geometry. We thus generate additional lines to guide the laser head between the pick-up locations of each component, and the target locations on the device as specified in the design file. To generate the lines, we use a simple path-planning procedure that allows for obstacle-avoidance of all other components on the canvas. While the user only draws circuit traces once, the path of each circuit trace must be tracked twice: once for dispensing the silver with the silver dispenser and once for curing with the defocused laser. We thus create a second copy of the circuit traces to be used for curing. Finally, we append the motion patterns to signal to the silver dispenser and pick-and-place mechanism when to start/stop their operation.

\textit{Offsetting lines in the X-Y plane:} As part of the post-processing, we apply offsets in the X-Y plane to the different fabrication steps. First the circuit traces and pick-and-place paths are offset to account for the physical offset between the laser head and the silver dispenser nozzle and the pick-and-place nozzle, respectively. Next, we apply a second round of offsets for curing and folding which involve the defocused laser at different heights. This is done to account for a slight misalignment between our laser and the z-axis, and requires a one-time calibration.

\textit{Order of execution:} Most laser cutters, including the ULS.PLS.150D used here, order the execution of lines by color and allow assigning each color to a different power/speed/z-axis setting. To enforce the correct ordering of steps in our fabrication pipeline, we assign each step a designated color. Because lines of the same color are by default executed in the order in which they are drawn and written to the design file (.svg), we re-order lines of the same color, for instance, to encode which bend lines should be executed first in the event of serial folds. Our system automatically post-processes the design file with the modifications detailed above, producing a fabrication file ready for the laser cutter.

\textit{Loading onto the laser cutter:} Following post-processing, users can load the output file into the regular laser cutter software to begin fabrication. Before starting fabrication, the user must also load the corresponding laser setting files (.las) that assigns power, speed, and z-axis settings for each color in the drawing. Settings for the pump and other hardware supporting the LaserFactory add-on’s operation are programmed once onto its microcontroller and then valid independent of the design file. All laser cutter settings and hardware add-on settings are summarized in Table~\ref{tab:laser-settings}. Users place material in the bed, mount the LaserFactory add-on onto the laser head using three 3D printed clips, load components into the storage area, and then execute the job.

\section{Applications}

The results from our evaluation indicate that devices created by Laserfactory can be used for applications requiring fine sensing and control, high frequency signals, and actuator driving with minimal loss. We demonstrate several of these capabilities through our example applications: a self-deploying quadcopter, a sensor-equipped wristband, and a printed circuit board (Figure~\ref{fig:app1ications}).

\subsection{Self-deploying quadcopter}

\begin{figure}[ht]
\centering
\includegraphics[width=0.99\linewidth]{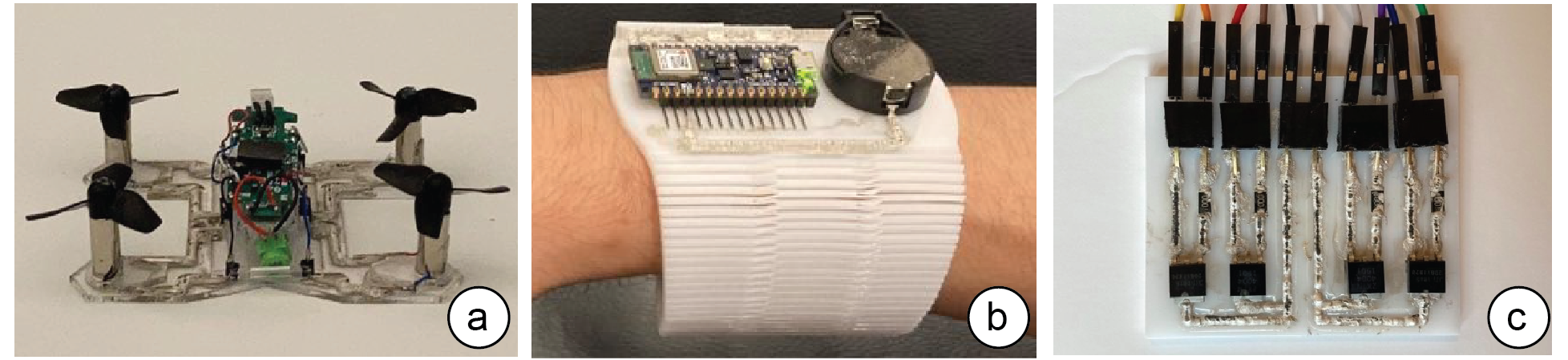}
\caption{Functional Devices made with LaserFactory: (a) a quadcopter that can lift of the fabrication platform, (b) a sensor-equipped wristband capable of gesture recognition, and (c) a printed circuit board, specifically an H bridge.}
\label{fig:app1ications}
\end{figure}

As our first example, we designed and fabricated a quadcopter (Figure \ref{fig:app1ications}a) to showcase three of LaserFactory's capabilities. When designing a quadcopter, one primary concern is minimizing its total mass. By using LaserFactory's ability to create geometry and circuit traces in tandem, we were able to reduce the area of the quadcopter platform to only contain material where circuit traces or components need to be placed, minimizing the overall mass. Second, this example illustrates LaserFactory's capability to fabricate devices that require non-trivial electrical currents, which is only practical with low resistance traces, such as those made by LaserFactory.

Third, this example illustrates LaserFactory's capability to create devices end-to-end without human intervention; in this case the quadcopter can self-deploy directly from the build plate if needed. The total fabrication time of the quadcopter was 11 minutes. The electronic components placed as part of the fabrication process include the four rotors plus an integrated PCB and battery. Note that unlike regular SMD components that can be used as-is, we harvested these components from an existing quadcopter and therefore had to expose and position the pins by fastening them to acrylic substrates prior to placing them on the LaserFactory component storage area in order to be compatible with our technique. 

\subsection{Personalized devices and wearables}

We fabricated a personalized, sensor-equipped wristband (Figure \ref{fig:app1ications}b) capable of sensing gestures that can be transmitted to control a mobile phone via bluetooth. It is comprised of a white sheet of acrylic with a personalized cutting pattern and a living hinge, in addition to an Arduino Sense BLE, battery, and passive components. By gesturing in one of four different directions (forward, backward, left, right), a different command is issued, and an LED changes color. The total fabrication time for this device was 9 minutes, after which the wristband can be picked out of the laser cutter and used immediately. This highlight's LaserFactory's ability to rapidly create personalized electronics on-demand. We also created capacitive sensors from traces themselves, showing how simple touch and humidity sensors as well as antennae can be fabricated without the need to procure dedicated external components.

\subsection{Printed circuit boards}

We fabricated an H bridge motor driver as an individual PCB (Figure \ref{fig:app1ications}c). H bridges are circuits used to switch the polarity of a load, allowing them to directly drive a DC motor bidirectionally. They are frequently used by hobbyists and electronic novices because microcontrollers, such as the Arduino platform, cannot drive motors directly from their pins. The H bridge takes 12 minutes to fabricate and is made up of 13 components: two PNP transistors, two NPN transistors, four resistors and five female header connectors. These connectors are used to easily mate with the four required control signals and a motor. This application demonstrates LaserFactory's capability to create ICs from basic SMD components, such as transistors and resistors. Since transistors and resistors can be used to create a variety of different ICs with different  functionality (e.g., H bridges, Op Amps), this example showcases that LaserFactory can create a diverse set of devices even when only basic components are loaded into the component storage area. Thus, rather than ordering the ICs, LaserFactory can make them on demand. Moreover, as a single fabrication file encodes all fabrication instructions, such files can be easily shared and downloaded in order to fabricate PCBs that other experts have designed.

\section{Discussion}
We next discuss design decisions and their limitations for the LaserFactory system and comment on opportunities for improvements. 

\textit{Transferability between different laser cutters:} Just as power, speed and height settings must be characterized for cutting the same material on two different laser cutters, the same is true for the settings of the LaserFactory add-on. The laser cutter settings in Table~\ref{tab:laser-settings}, 3D printed clips for mounting the add-on, and the motion signal used are all calibrated for use on our specific laser cutter machine, i.e. the ULS.PLS.150D system. Thus, the settings need to be re-calibrated to work with the specific laser power, laser head shape, and carriage motion that is expected to differ between machines. Underlying this is the requirement for the firmware to allow reading an SVG/PDF file line by line, and to not automatically smooth all accelerations. For laser sintering, we tested and confirmed its compatibility with a range of power, speed and height permutations. This suggests that laser cutters within a range of power ratings could potentially be compatible for use with laser sintering as long as they have an adjustable platform height. While outside the scope of our work, it would be an important future work to validate this cross-device functioning. 

\textit{Unidirectional communication}: The platform agnosticism gained by leveraging unidirectional communications comes with the trade-off that LaserFactory is not set up to transmit instructions back to the laser cutter firmware. In its current state, this prevents LaserFactory from being able to communicate interrupts to the laser cutter software to allow for potential interventions. Such interrupts would be useful to allow aborting fabrication in cases of malfunction or an unexpected contact detected by the pick-and-place sensor. One method of enabling bidirectional communication would be to fit the hardware add-on with an onboard transceiver, however interfacing it with laser cutter firmware would necessarily be platform specific.

\textit{Component Loading:} In the current version of LaserFactory, we use a pre-loaded component storage inside the laser cutter that users can re-stock as needed. The storage consists of engraved footprints of electronic components that help the user align them correctly during loading. LaserFactory uses these positions to compute paths for the pick and place operations. The current system thus relies on foreknowledge about what components will be used. For future work, we plan to investigate how we can enable pre-stocking of a larger number of components and how to feed in components on-demand to accommodate real-time design decisions. In addition, LaserFactory's hardware currently does not support electronic component rotation. Thus, the orientation of pre-stocked components must match that of their virtual counterparts. For future work, we plan to add a rotation axis to the pick-and-place tool to extend the degrees of freedom available to designers. We also plan to investigate stocking material sheet in the component storage area, which can be cut and stacked to form multimaterial structures.

\textit{On-Demand Device Programming:} The microcontrollers that are picked-and-placed as part of the fabrication process are currently manually pre-programmed ahead of time, requiring foreknowledge of the device in which they will be used and its context. For future work, we plan to explore methods for flashing programs on-demand to accommodate more versatile deployment; a promising candidate for this could be programming via infrared receivers (Kilobots~\cite{rubenstein2012kilobot}).

\section{Summary}
In this chapter, we presented LaserFactory, an integrated fabrication platform that can rapidly create the geometry of a device, create its circuit traces and assemble components without manual intervention. We demonstrated how we can augment an existing laser cutter with a hardware add-on without interfacing with its underlying firmware by using a motion-based signaling technique that can inform the add-on when to start and stop its operation. We illustrated the two main features of our hardware add-on, a silver dispenser used for circuit trace creation and a pick-and-place mechanism used for assembling electronic components, and showed that the add-on can create high-resolution traces of high conductivity and assemble a range of different electronic components. We then showcased laser sintering, a technique that uses a $\mathrm{CO_2}$ laser to cure dispensed silver paste and discussed which laser cutter settings are most suitable to cure the traces. Finally, we showed our end-to-end design and fabrication pipeline consisting of a design tool, a visualization tool, and a post-processing script that transforms the design file into machine instructions for LaserFactory. We also showed example applications that included a quadcopter with actuators, a sensor-enhanced wristband, and a PCB assembled from basic transistors and resistors. Researchers have in recent decades made significant progress toward the long-term vision of being able to download a device file and have it fabricated at the push of a button. While laypeople can today do so for passive, primarily decorative objects via commercially available laser cutters and 3D printers, the fabrication of fully functional, electromechanical devices demonstrated in this chapter takes a step toward that shared vision.

By developing multi-process manufacturing machines that consolidate different manufacturing processes into a single platform, we can build mesoscale artefacts like robots in a single machine. This is useful for a variety of applications, but there are many instances that require the ability to re-configure our existing infrastructure for new needs, instead of fabricating them from the ground up. One such application is space structures, like optical telescopes. We need to leverage the most advanced of our manufacturing technologies to make these on earth, but the size of all current space-borne telescopes are limited as a result. One way to overcome this is to still manufacture them on earth, but to do so in modular parts, and enable them to assemble in orbit. In the next chapter, we will introduce a modular assembly framework to allow structures to be partitioned into modules, transported compactly, and self-assemble in orbit using embedded electromagnets.

%% file: Electrovoxel.tex
\chapter{Modular Self-assembly via Reconfiguration}
\label{sec:Electrovoxel}

In the previous chapter, we introduced a method to automating fabrication and assembly at the machine level, and identified instances such as the space environment where automating assembly of structures at the part level would be advantageous. To address this, we use a modular robotic framework that conceives of a space structure as being composed of modular parts, and use actuators embedded in these modules to automate assembly on orbit.

\begin{figure}[ht]
  \centering
  \includegraphics[width=0.9\linewidth]{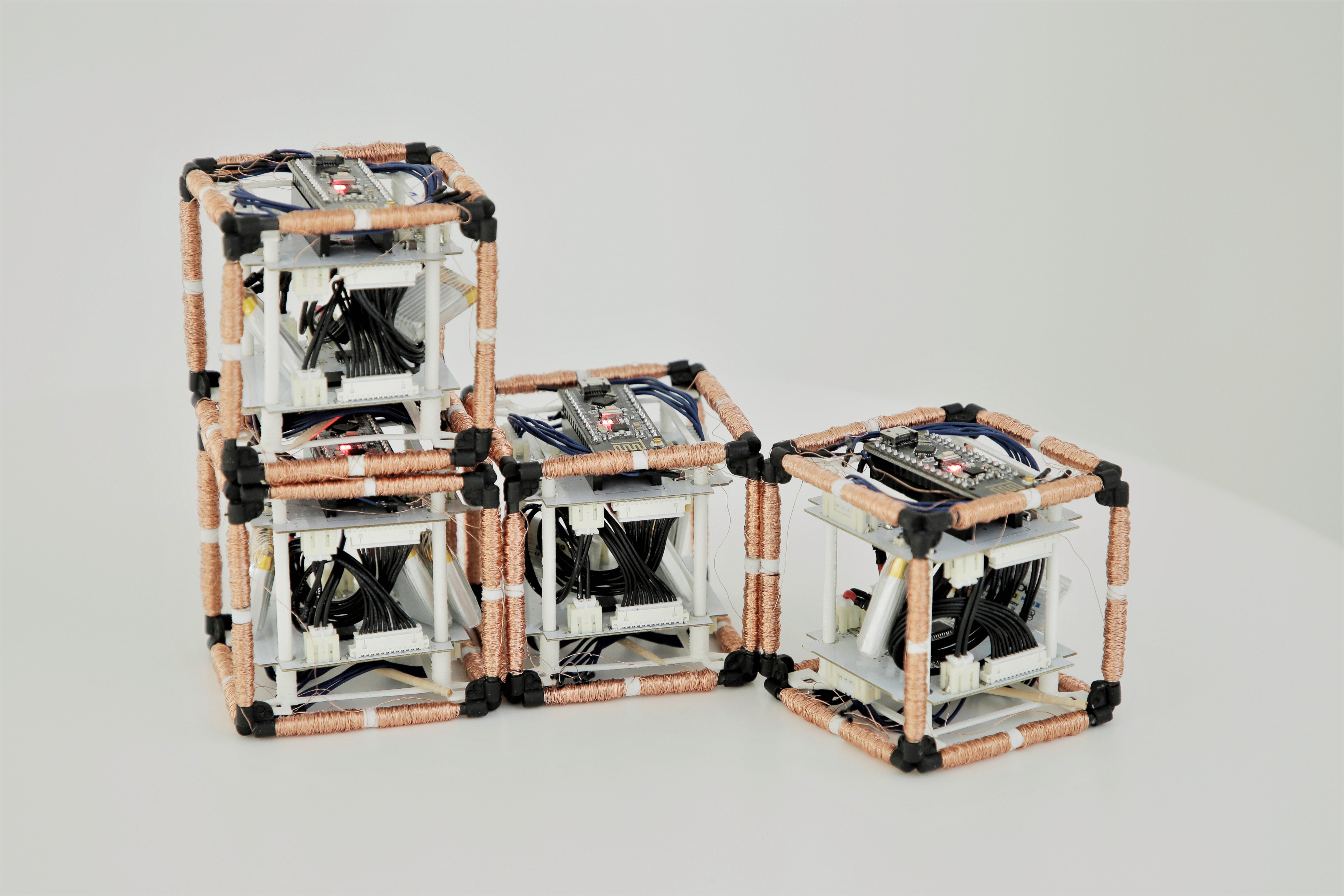}
  \caption{Four Electrovoxels, electromagnetically actuated modules capable of reconfiguring in three dimensions in microgravity using embedded electromagnets.} 
  \label{fig:EV-ensemble}
\end{figure}

In this chapter, we outline the first demonstration of reconfigurable robots leveraging an electromagnetically actuated pivoting framework that are {\it fully untethered}, supported by reconfiguration planning software and electromagnet force predictions verified experimentally. A key goal is to validate these robots' use for microgravity environments to enable near-term space industry applications~\cite{yim2003modular,yim2007modular,nisser2017electromagnetically,hauser2020kubits}, where propellant-free actuation and reconfigurability address many challenges associated with today’s limitations on launch mass and volume, as well as facilitating stowage during launch. Reconfigurable modules can enable the augmentation and replacement of structures over multiple launches, form temporary structures to aid in spacecraft inspection and astronaut assistance, function as self-sorting storage containers, and allow spacecraft to actively change their inertia properties. Microgravity alleviates demands placed on actuation forces, facilitating untethering of the modules by moving electronics onboard, and we chose electromagnet parameters such as winding number, core radius and material to limit current. We also parameterize Amp\`ere’s Force law in terms of these parameters to expand the design space of electromagnetic actuators for future modules' force and mass requirements. We simulate a microgravity environment using an air table, and deploy our modules on a parabolic flight to demonstrate untethered three-dimensional reconfigurability in space.

Relative to existing self-reconfigurable robots~\cite{romanishin2013m, romanishin20153d}, our robots are light (103g), inexpensive (\$68), and easy to fabricate (80 min/cube), promising scalability. In addition, using full assemblies, we demonstrate Sung et al.'s \cite{sung2015reconfiguration} two reconfiguration primitives, the pivot and traversal, demonstrating the electromagnetically actuated pivoting framework's compatibility with algorithms that allow reconfiguring large numbers of cube-based robots between arbitrary 3D shapes. To show how the framework complies with \cite{sung2015reconfiguration} to reconfigure more complex shapes, we constructed a web interface that simulates reconfiguration between user-defined shapes.

\begin{figure*}[ht]
  \centering
  \includegraphics[width=0.99 \textwidth]{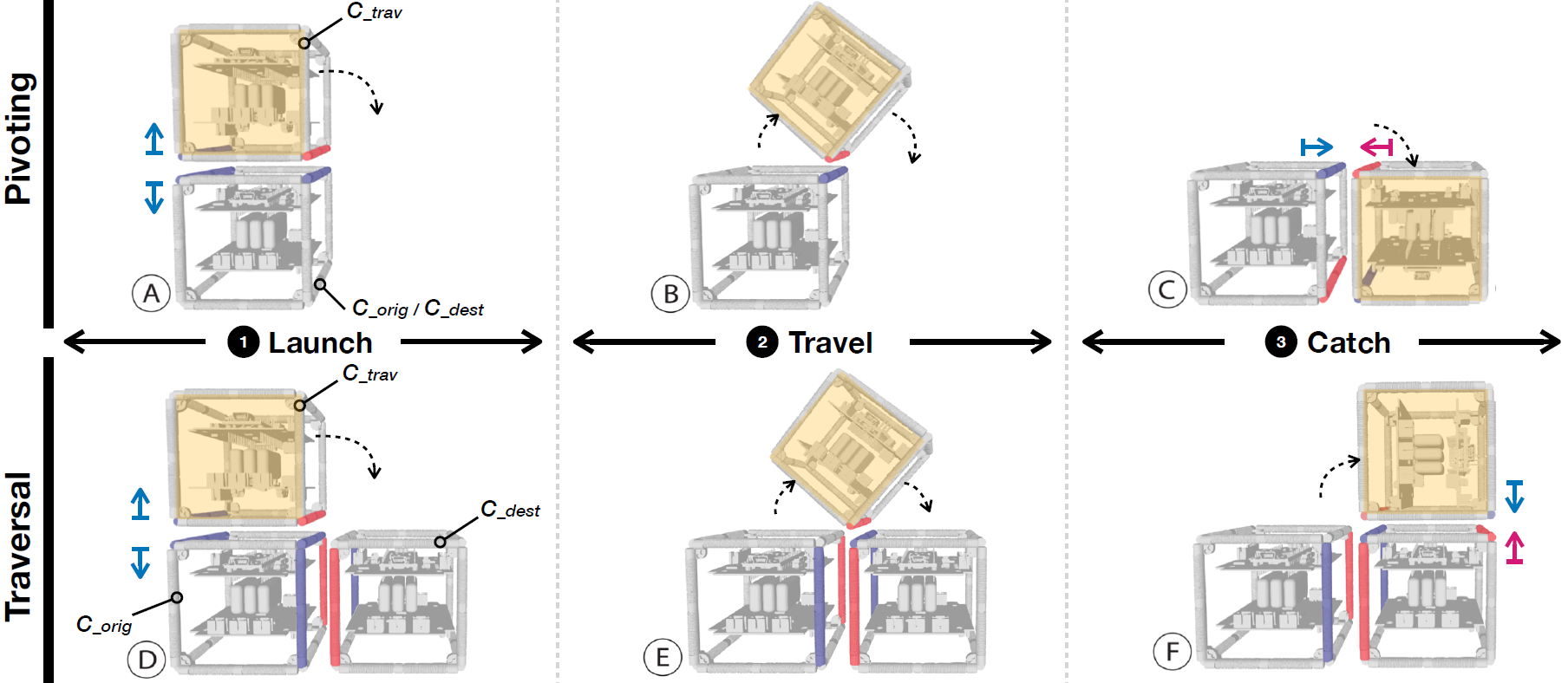}
  \caption{Reconfiguration maneuvers for (above) Pivoting and (below) Traversal. Electromagnets are shaded in red and blue to indicate polarization w.r.t. a global co-ordinate system; like polarizations repel, unlike polarizations attract.} 
  \label{fig:image-for-algo-martin}
% \vspace{-0.6cm}
\end{figure*}

\subsection{Actuation Mechanism} 
\label{sec:alg}
Algorithm \ref{"alg:command-recipe"} describes both the pivot and traversal reconfiguration maneuvers for our electromagnetically actuated robots. There are three steps to the polarization sequence for both the pivot and traversal maneuvers, which we call the \textit{Launch}, the \textit{Travel}, and \textit{Catch} phases (Fig. \ref{fig:image-for-algo-martin}). In each of these phases, three cubes are involved: a \textit{traveling} cube (the cube selected for moving), an \textit{origin} cube (from which the traveling cube launches), and a \textit{destination} cube (which catches the traveling cube). For the pivot (Fig. \ref{fig:image-for-algo-martin}, 1st row), the origin and destination cubes correspond to the same physical cubes; for the Traversal (Fig. \ref{fig:image-for-algo-martin}, 2nd row), they correspond to different cubes. The algorithm inputs are a cube ID, its desired rotation axis and rotation direction. Given these inputs, all electromagnet assignments (repulse, attract, or OFF) are uniquely defined and identified by our software.

% \iffalse
%%%%%%%%%%%%%%%%%%%%%%%%
\begin{algorithm}[t]
\caption{General Reconfiguration Algorithm}
\label{"alg:command-recipe"}
\begin{algorithmic}[1]
\Procedure{PivotCube}{}
\\
\\
\hspace*{\algorithmicindent} \textbf{Input:}
\hspace{4pt} $C_{travID}$ \hspace{14pt} Traveling cube ID \\ 
\hspace{50pt}  $[axis,dir]$ \hspace{5pt} Axis, Direction of pivot \\

\hspace*{\algorithmicindent} \textbf{Action:}
\hspace{1pt}Perform pivot/traversal using electromagnets \\ 
% \hspace{57pt}  $[axis,dir]$ \hspace{5pt} Axis, Direction of pivot 
\\
\State // Create cube objects with correct electromagnets 
\State $C_{trav} \gets getTravelCube(C_{travID}, [axis,dir])$
\State $C_{orig} \gets getOriginCube(C_{trav}, [axis,dir])$
\State $C_{dest} \gets getDestinationCube(C_{trav},[axis,dir])$

% \State // Each cube contains 12 electromagnets. Especially relevant magnets will be referred to here as $M_{launch}$, $M_{hinge}$, and $M_{catch}$ as shown in Fig. \ref{fig:image-for-algorithm}
\\
% \State // (0) 
\If {traversal}
    \State $\textbf{attach}(C_{orig}, C_{dest})$
    % \State $\textbf{pwm}(C_{push}, C_{push} \to M_{sticker1}, d_{02})$
    % \State $\textbf{pwm}(C_{stop}, C_{stop} \to M_{sticker2}, d_{03})$
    % \State $\textbf{pwm}(C_{push}, C_{push} \to M_{sticker2}, d_{04})$
\EndIf
\\
\State // Launch
\State $\textbf{createHinge}(C_{trav}, C_{orig}, [axis,dir])$
\State $\textbf{launchPivot}(C_{trav}, C_{orig}, [axis,dir])$
\State $\textbf{wait}(timeForLaunching)$
\\
\State // Travel
\State $\textbf{endLaunchPivot}(C_{trav}, C_{orig}, [axis,dir])$
\State $\textbf{wait}(timeForTraveling)$
\\
\State // Catch
\State $\textbf{attach}(C_{trav}, C_{dest})$

\EndProcedure
\end{algorithmic}
\end{algorithm}
%%%%%%%%%%%%%%%%%%%%%%%%
% \fi

During the Launch phase (Fig. \ref{fig:image-for-algo-martin} A and D), we polarize one electromagnet pair identically to launch the maneuver, while oppositely polarizing a second pair to form an attractive hinge. For the Traversal (Fig. \ref{fig:image-for-algo-martin}, bottom), we energize two additional pairs of electromagnets to keep the non-traveling cubes attached to one another; for this, we choose electromagnets oriented orthogonally to the launching electromagnets in order to avoid unwanted interactions between these pairs. During the Travel phase (Fig. \ref{fig:image-for-algo-martin} B and E), after a short pulse we switch the launching electromagnets off, while the remaining electromagnet pair remain attractive to maintain the hinge. During the Catch phase (Fig. \ref{fig:image-for-algo-martin} C and F), we energize a new pair of attractive electromagnets to form a stable bond in the newly acquired configuration.

\subsection{Simulation and Control Interface}

\begin{figure}[ht]
  \centering
  \includegraphics[width=0.80\columnwidth]{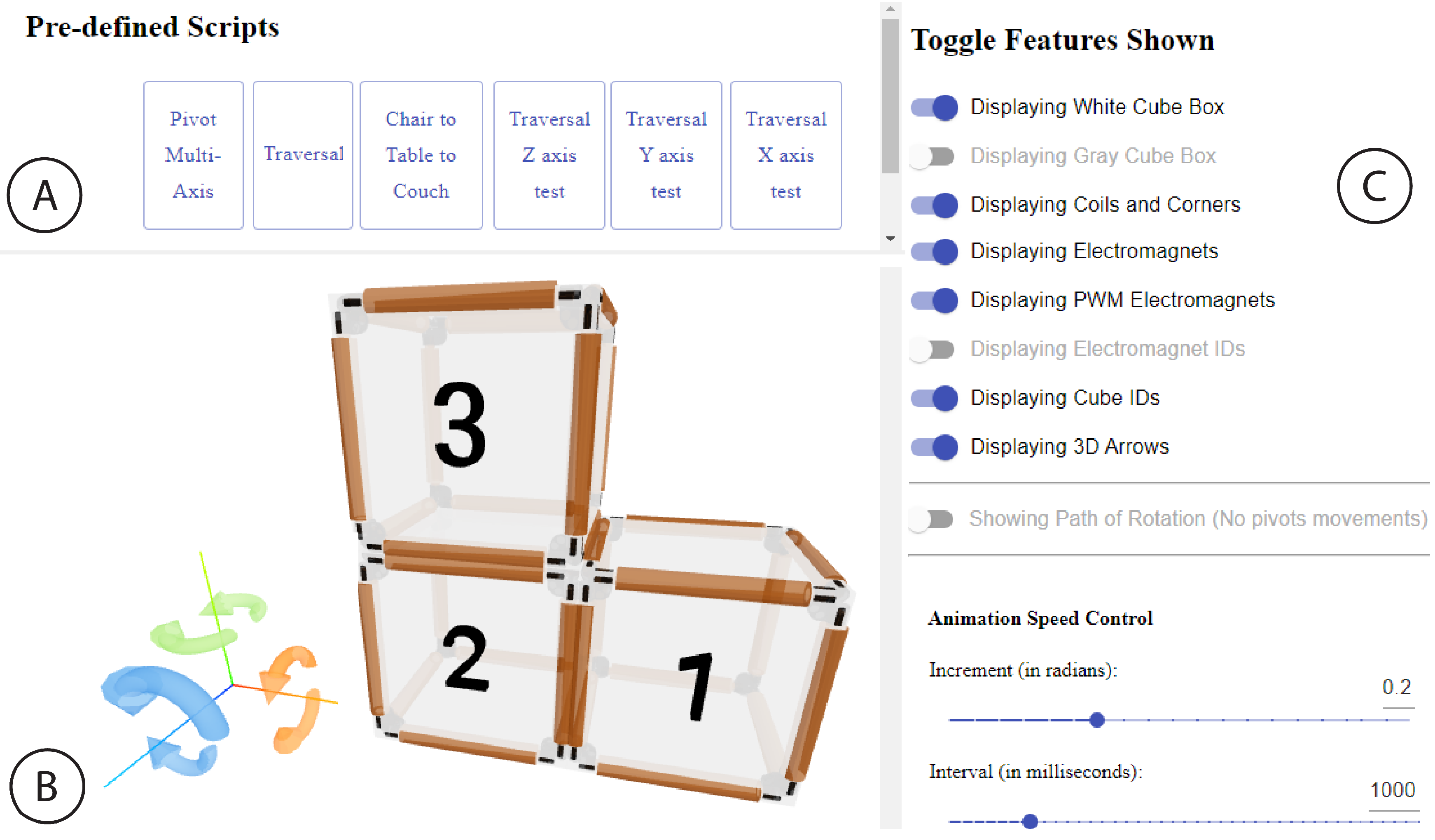}
  \caption{Web simulation for planning reconfigurations and calculating the associated electromagnet commands. (A) Pre-scripted maneuvers. (B) Viewport. (C) Settings.} 
  \label{fig:simulation-screenshot}
% \vspace{-0.6cm}
\end{figure}

Manually planning pivoting maneuvers and their associated electromagnet assignments becomes intractable for more than a few cubes. To let users visualize and plan reconfiguration maneuvers, we developed a simulation (Fig. \ref{fig:simulation-screenshot}) that computes all electromagnet assignments based on desired reconfiguration maneuvers specified by the user. The simulation is browser-based and built using React, TypeScript, and Three.JS. It consists of three parts: (A) different ways to interact with the cubes (via buttons, direct manipulation, or code), (B) a viewport that simulates the cubes and affords their direct manipulation, and (C) a settings panel to toggle simulation features (e.g., different render modes for the cubes). 

% \textit{Interaction} 
We provide three ways to define maneuvers. Users can initiate maneuvers directly by clicking cubes and arrow directions, each resulting in a single pivot. Alternatively, they can launch pre-defined scripts via the buttons that encode multiple consecutive rotations. And finally, new buttons encoding different reconfiguration maneuvers can be added at any time in the underlying Typescript file. To do so, users define the number and locations of starting cubes addressed by (x,y,z) coordinates, and define each subsequent maneuver by specifying the cube number and pivot direction. 

% \textbf{System} 
Cubes in the viewport are affixed to a grid of cubic cells of unit length, with each cube occupying an integer address on coordinates (x,y,z). Rotations are permitted along orthogonal axes X,Y,Z, in clockwise or counter-clockwise directions. Because physical rotations require cubes to form both repulsive and attractive edges, cubes must share a face with another cube to execute a valid pivot. Each cube is assumed to have access to the local occupancy information of its neighboring cells. Prior to executing a pivot, this occupancy is checked in order to determine whether a selected rotation results in a pivot or a traversal. If the maneuver path is obstructed, the viewport returns an error message. Given a pivot direction and unobstructed path, there exists for each cube a unique valid edge about which to form a hinge, but up to two valid edges to repel and actuate the maneuver; in this case, we choose the edge corresponding to the cube with which the hinge is formed. We represent all rotations using quaternions in order to facilitate unordered rotations about multiple axes. 

Finally, the settings panel allows users to set rendering features. These include displaying IDs for cubes, their electromagnets and their polarization values, displaying occupancy requirements of neighboring cells to prevent collisions for desired maneuvers, toggling animation speed, and setting rendering fidelity. Once a sequence of maneuvers is defined, the associated electromagnet assignments can be ported to a transmitting microcontroller for deployment on the hardware.

\section{Electromagnets}

In this section, we first describe the selection of the electromagnet parameters for our system. To support further exploration of electromagnetically driven pivoting cubes, we then provide a force model to allow generating candidate electromagnets by using Amp\`ere’s Force law to compute magnetomotive force for a given electromagnet pair parametrically. We finally apply this force to a preliminary dynamical model of a 2-cube system.

\subsection{Electromagnet Parameters}

The idealized case of Amp\`ere’s law gives that an electromagnet's force is proportional to $(NI)^2\mu$, where $\mu$ is the permeability of its core, $N$ the number of turns and $I$ the current applied. To limit each cube's mass and size, we first selected the narrowest COTS ferrite cores available, at 1.625mm radius R and initial permeability $\mu_0$ of 2000. In addition, larger cores proved difficult to be supported by the rated pressure of our air table on a 60mm-side cube footprint. We next chose the COTS SMD drivers we found capable of delivering the highest current, at 1.2A continuous. Using exploratory experiments, we determined that an $N$ of 800 provided sufficient force to pivot 100g prototype cubes within interactive times of 2s. Finally, choosing a wire gauge of AWG 34 at this $NI$ yielded a coil resistance of $10.5\Omega$, allowing driving the electromagnets, microcontroller and auxiliary electronics from a single untethered power source of 11.1V-12.6V.

\subsection{Electromagnet Force Model}

\begin{figure}[t]
  \centering
  \includegraphics[width=0.90\columnwidth]{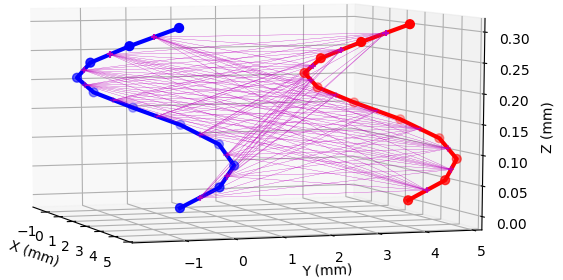}
  \caption{Computing electromagnet forces. Shown here for $D_1,D_2$=10 between 1-turn coils (force vectors in magenta).} 
%   \Description{Simulation}
  \label{fig:elec-model}
% \vspace{-0.6cm}
\end{figure}

The dipole approximation commonly used to find electromagnets' magnetic field strengths is invalid over short separation distances, such as when neighboring cubes are in contact. As such, we use Amp\`ere’s force law (\ref{cont}), which expresses the force $F_{1,2}$ exerted on coil 1 due to coil 2 as a double line integral over each coil's geometry where infinitesmal wire elements $\textup{d}l_1$ and $\textup{d}l_2$ in wires 1 and 2 are energized with currents $I_1$ and $I_2$, respectively, $\hat{r}_{12}$ is a unit vector from each element on wire 1 to those on wire 2 separated by distance $r$, and $\mu(I)$ is the permeability of the electromagnet core as a function of current.
  
\begin{equation}
F_{1,2} = \frac{\mu(I)}{4\pi} \int_1 \int_2 \frac{I_1 \textup{d}l_1 \times (I_2 \textup{d}l_2 \times \hat{r}_{12} )}{\left|\left|r \right|\right|^2}
\label{cont}
\end{equation}

Equation (\ref{cont}) has no known analytical solution and is discretized to give (\ref{eq_disc}), where $D_1$ and $D_2$ represent the number of discretized elements in wires 1 and 2. 
% The system dynamics are assumed to be slow and a quasi-static approximation holds.

\begin{equation}
F_{1,2} = \frac{\mu(I)}{4\pi} \displaystyle\sum_{p\, = \,1}^{D_1}  \displaystyle\sum_{q\, = \,1}^{D_2} \frac{I_p \textup{d}l_p \times (I_q \textup{d}l_q \times \hat{r}_{pq} )}{\left|\left|r \right|\right|^2}
\label{eq_disc}
\end{equation}

We parameterize (\ref{eq_disc}) in terms of radius, length, turns, and pitch, and solve this numerically for our electromagnet parameters, using 8000 elements ($D_1,D_2=8000$) per coil to compute $F_{1,2}$ from the sum of 64,000,000 force vectors over separation distances of 0.5mm to 20mm in 0.5mm increments (85 minutes/increment on a Razer Blade Intel(R) Core(TM) i7-8750H). Fig. \ref{fig:elec-model} illustrates this computation for single-turn electromagnet coils with 10 elements each (100 force vectors).

\subsection{Dynamic Model}

\begin{figure}[t]
  \centering
  \includegraphics[width=0.80\columnwidth]{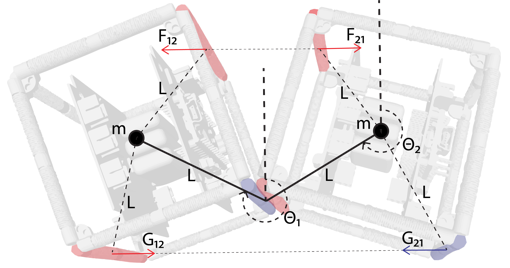}
  \caption{Dynamical model. Massless links of length L (solid lines) connect point masses m. Force F actuates the pivot via repulsion; G attracts electromagnets to form a new stable bond.} 
  \label{fig:dyna}
\end{figure}

We apply these forces to a preliminary model (Fig. \ref{fig:dyna}). We model two 60-mm side length cubes as a 2-link pendulum consisting of two point masses $m$ placed at the distal ends of two massless links (length $L$) connecting the hinge to each cube's center of mass. Electromagnet forces are applied at locations that correspond to the positions of the other two electromagnet pairs associated with the maneuver along vectors that connect them. We derive the equations of motion and solve with Kane's Method using Python's SymPy package.

\section{Hardware}

\subsection{Electronics}

\begin{figure}[ht]
  \centering
  \includegraphics[width=0.99\columnwidth]{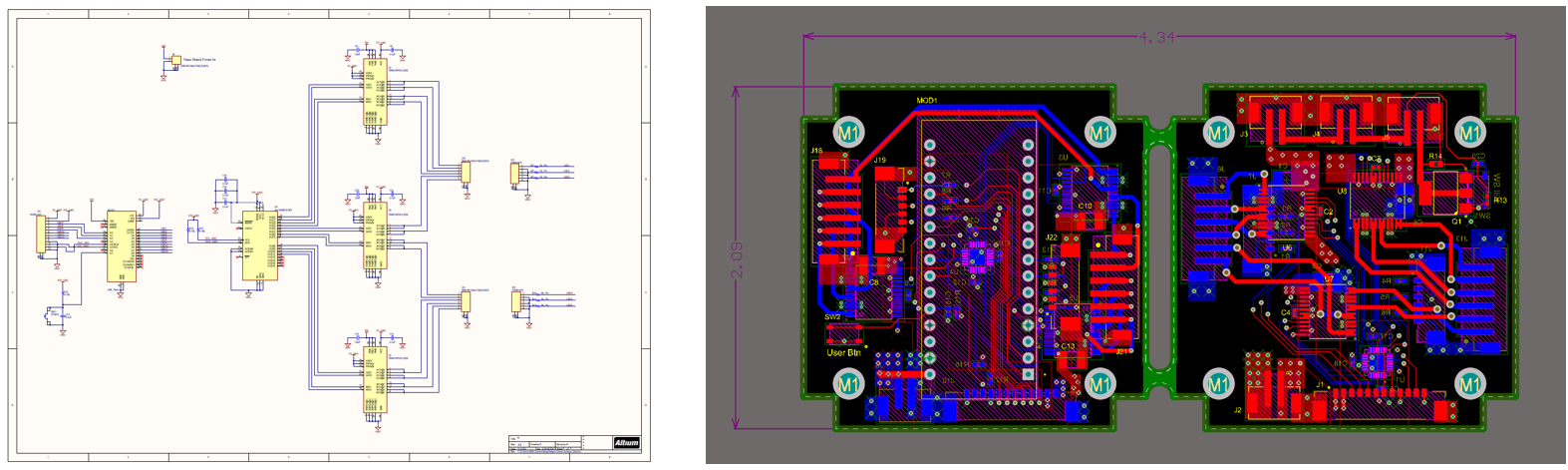}
  \caption{Printed Circuit Board design. (Left) Schematic, page 1 of 2. (Right) Layout, board 1 of 2.} 
  \label{fig:pcb}
\end{figure}

Each electromagnet is comprised of 800 turns of 34 AWG magnet wire wound around a ferromagnetic core (fair-rite 77) of 3.25mm diameter, 55.5mm length and initial permeability ($\mu_i$) of 2000, with average electrical properties characterized by a capacitance of $118.1 \mu F$, an inductance of $21.44mH$, a resistance of $10.65 \Omega$ and a Q factor of $1.265$. Each actuator (core + winding) costs just \$0.66. The circuitry for an untethered cube with 12 electromagnets consists of a microcontroller (Arduino Nano) integrated with a wireless transceiver (nRF24L01), two 16-channel GPIO expanders (Semtech SX1509) and 6 full dual H-bridges (Toshiba TB6612FNG). These are distributed evenly between two double-sided 0.78mm PCBs of square cross-section (side length 42mm) which sandwich three serially connected 3.7V batteries (ENGPOW 3.7v 150mAh Lipo, 4.2V at full charge). Combined, this allows controlling each electromagnet as to enable bidirectional pivoting in three orthogonal axes. A partial view of the schematic and layout for the printed circuit boards are shown in Fig. \ref{fig:pcb}.  

We use an NRF-equipped Arduino Nano as a centralized controller to transmit commands from a laptop to modules via radio. We utilize a simple open-loop bang-bang control scheme. To accommodate addressing N cubes, each with 12 electromagnets, where each electromagnet can be polarized in two directions or turned off, each command consists of a 16-bit signed integer that encodes the cube ID $[1..N]$, electromagnet ID $[1..12]$ and its polarity $[-1,0,1]$. Individual messages are transmitted in 20 milliseconds, and separate commands can be transmitted to configure individual cubes to drive selected electromagnets using PWM at a chosen duty cycle $[0..255]$.

\subsection{Mechanical Design}

\begin{figure}[ht]
  \centering
  \includegraphics[width=0.97\columnwidth]{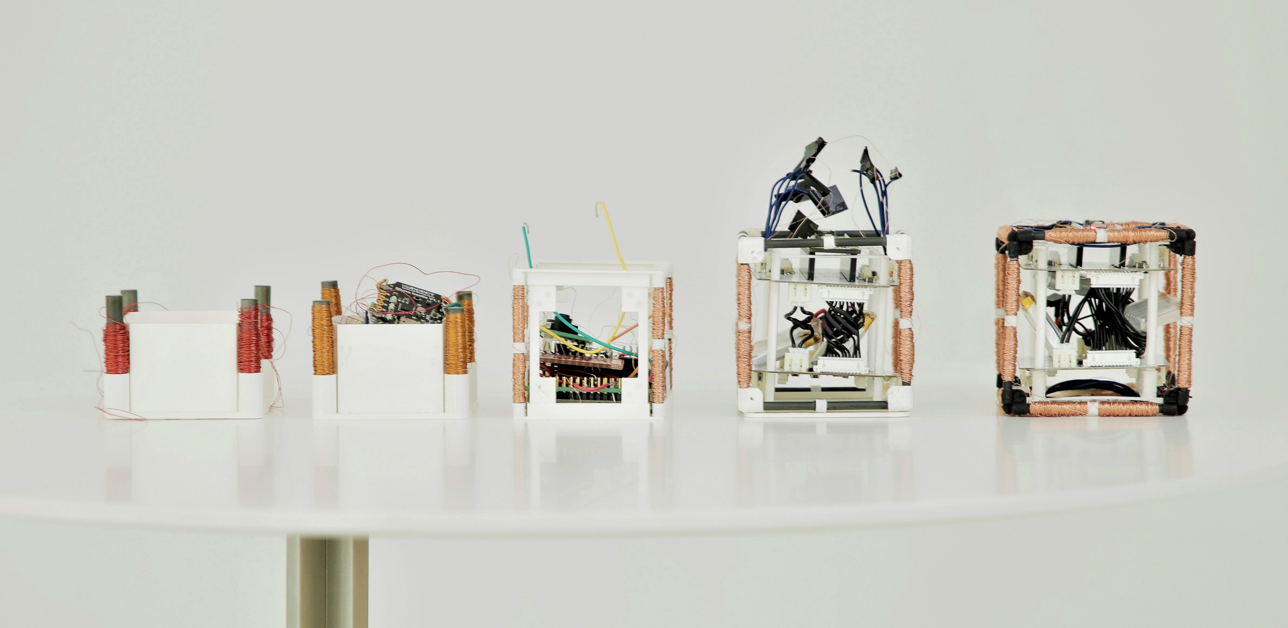}
  \caption{Iterations on module design, from (left) first to (right) final prototype.}
  \label{fig:iterations}
\end{figure}

\begin{figure}[ht]
\centering
\includegraphics[width=0.85\columnwidth]{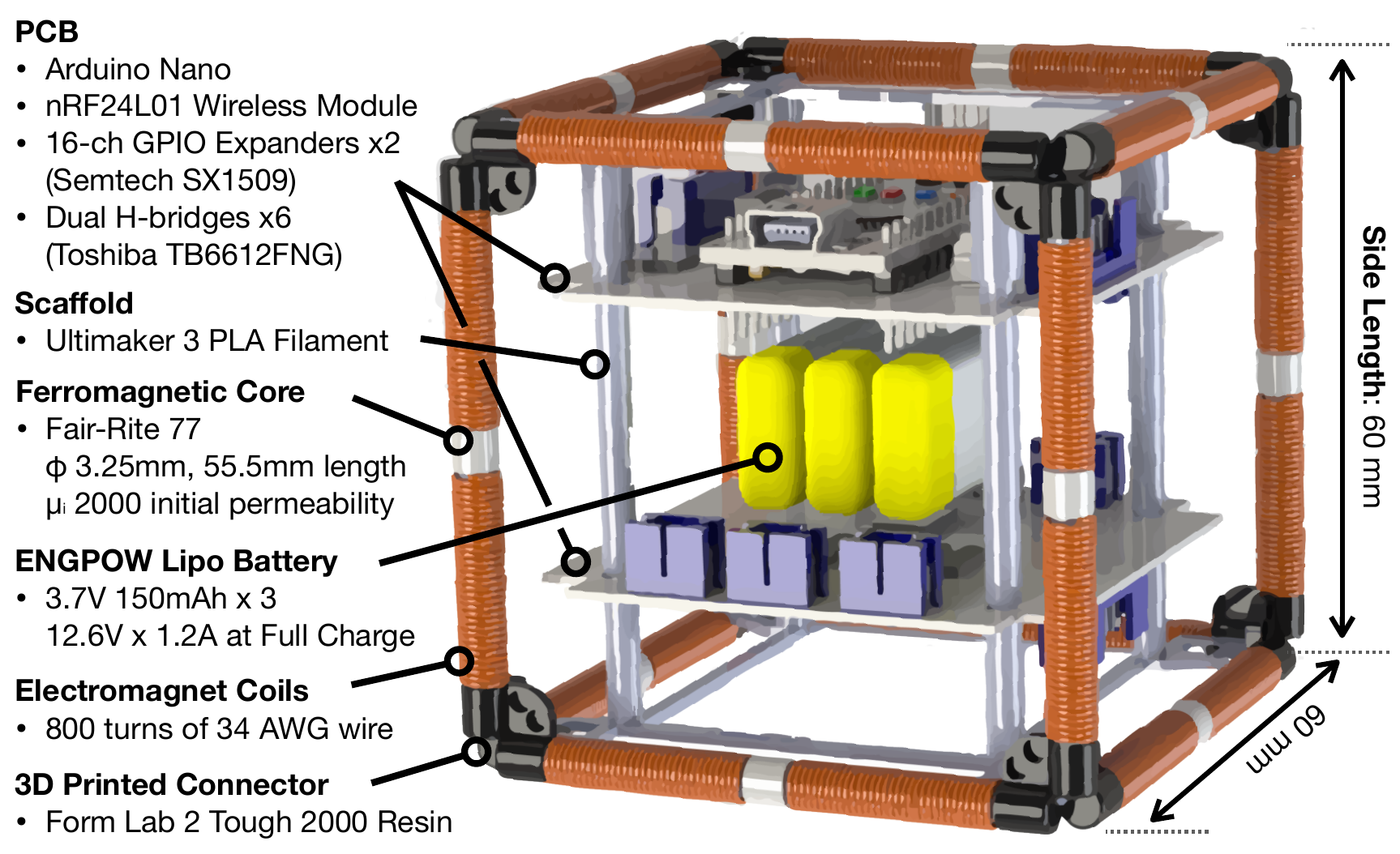}
\caption{CAD model of a cube module.}
\label{CAD}
\vspace{-0.6cm}
\end{figure}

We undertook a number of design iterations to reach the final prototype architecture (Fig. \ref{fig:iterations}). Each module in the final design (Figures \ref{fig:EV-ensemble} and \ref{CAD}) is a 60mm side length cube and can be described as a unit cell of a primitive cubic Bravais lattice, with edges representing electromagnets that connect to vertices representing corner connectors. In the middle of the cube, two PCBs sandwich three batteries, centering the system's mass to limit moment of inertia, and are fixated by a scaffold that interfaces via struts to all 8 corner connectors. The corner connectors are 3D printed on a Formlabs 2 using Tough 2000 resin and the scaffold from PLA using an Ultimaker 3. Table \ref{tab1} details the cost and mass breakdown of a cube; mass and cost are for total in each row, and structure costs are based on raw material pricing.

\begin{table}[ht]
\caption{Module Cost/Mass Breakdown}
\begin{center}
\begin{tabular}{|c|c|c|c|c|}
\hline
% \textbf{Table}&\multicolumn{3}{|c|}{\textbf{Table Column Head}} \\
% \cline{2-3} 
\textbf{Section} & \textbf{Part}& \textbf{Number}& \textbf{Mass (g)}& \textbf{Cost (\$)}\\
\hline
Electromagnet & Ferrite core     & 12 & 22.7 & 6.2 \\
Electromagnet & Coil winding     & 12 & 24.1 & 1.8 \\
\hline
PCB           & MCU              & 1  & 5.6  & 8.9 \\
PCB           & Cabling          & 36 & 9.2  & 11.4 \\
PCB           & Boards \& ICs    & 2  & 20.4 & 31.6 \\
PCB           & Batteries        & 3  & 15.6 & 8.2 \\
\hline
Structure     & Corners          & 8  & 2.7  & 0.4 \\
Structure     & Scaffold         & 2  & 2.8  & 0.2 \\
\hline
\textbf{Total}&     /            &  / & \textbf{103.1}& \textbf{68.7}\\
\hline
% \multicolumn{4}{l}{$^{\mathrm{a}}$Sample of a Table footnote.}
\end{tabular}
\label{tab1}
\end{center}
\end{table}

Two Molex cables connect the upper and lower PCBs, and four cables of 6 wires each (2 leads/electromagnet) harness triads of three electromagnets to the PCBs. These four triads consolidate all electromagnet wirings to two pairs of diametrically opposed corners to simplify wire routing; each triad connects three orthogonally positioned electromagnets to the PCBs in order of axis to ensure symmetry between all cubes such that they respond identically for a given command.      

Each cube took 80 minutes to assemble, discounting time to reflow-solder PCBs and manually wind electromagnets.

\section{Experiments \& Results}

\subsection{Simulation} The simulation renders pivoting maneuvers and outputs correct electromagnet IDs and polarity assignments for any valid reconfiguration; assignments were verified on hardware by reconfigurations across all dimensions and electromagnets. The simulation supports real-time interaction with up to 200 modules while rendering the associated CAD files (1.1Mb .STL), and replacing these with low resolution proxy cubes permits interaction with 1000 modules. Fig. \ref{fig:chair-to-table-to-couch} (see supplementary video) illustrates the viewport simulating and computing electromagnet assignments for reconfiguring 19 cubes from a chair to a table (via 22 maneuvers) to a couch (40 maneuvers). 

\begin{figure}[ht]
  \centering
  \includegraphics[width=0.99\columnwidth]{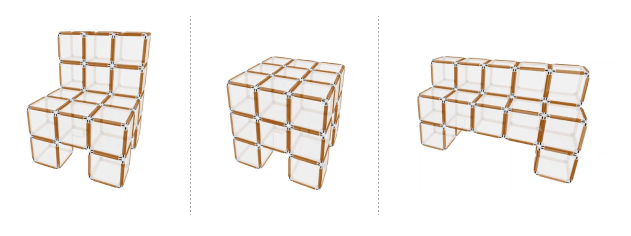}
  \caption{Reconfiguring between a chair, table and couch}
  \label{fig:chair-to-table-to-couch}
\end{figure}

\subsection{2D Experiments: Air Table}

The modules were deployed on an air table (ScienceFirst \#12000) and programmed to perform the two reconfiguration primitives; a pivot and a traversal (see supplemental video). Each electromagnet drew 11.7W (nominally 11.1V x 1.06A) to 15.1W (fully charged, 12.6V x 1.2A) for the duration of the maneuver, which was 1.53s for pivots and 1.03s for traversals.  
%and programmed to perform pivoting maneuvers sequentially such that each electromagnet was tested as both a hinge and as an actuator in turn.

52 pivots and 30 traversal maneuvers were performed, with a success rate of 100\% and 94\%, respectively. To yield this success rate, the electromagnets of the manually assembled cubes required careful positioning; small misalignments resulted in failures of the cubes to generate sufficient attractive forces to catch stably, and traversal maneuvers showed a higher likelihood of failure due to involving more electromagnets.

\subsection{3D Experiments: Parabolic Flight}

\begin{figure}[ht]
  \centering
  \includegraphics[width=0.99\columnwidth]{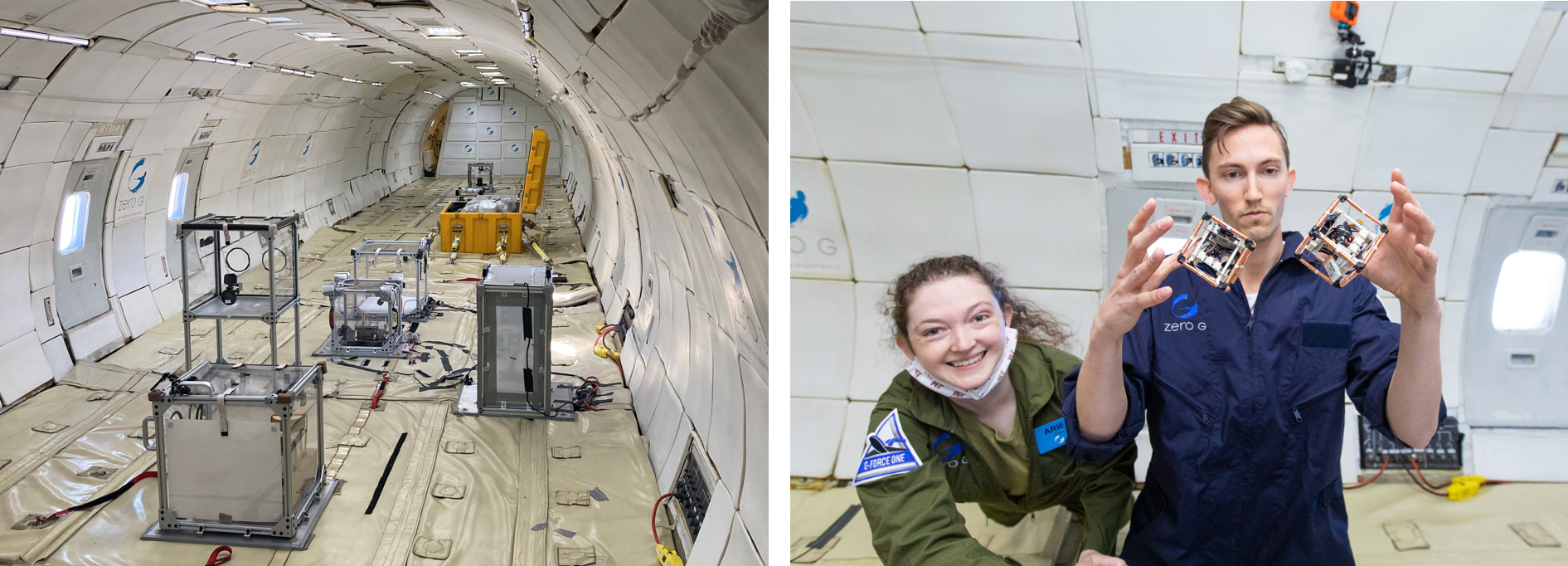}
  \caption{Parabolic flight deployment. (Left) experiments installed on the aircraft. (Right) free-floating Electrovoxels during a parabola.} 
  \label{fig:flight}
\end{figure}

\begin{figure}[ht]
\centering
\begin{subfigure}{0.5\columnwidth}
\centering
    \includegraphics[width=0.90\columnwidth]{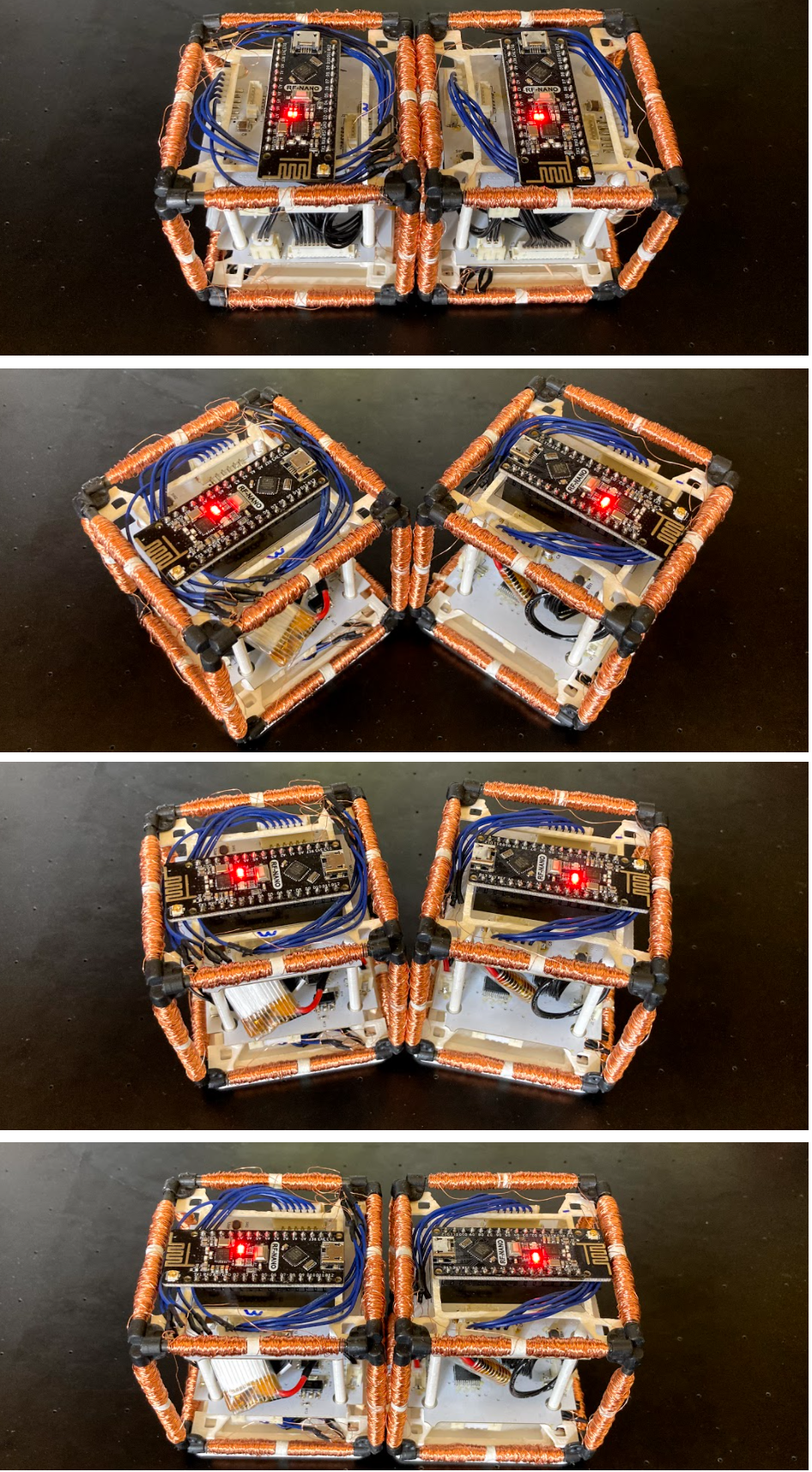}
    \caption{Air table}
    \label{fig:pivot_}
\end{subfigure}%
\begin{subfigure}{0.5\columnwidth}
\centering
    \includegraphics[width=1\columnwidth]{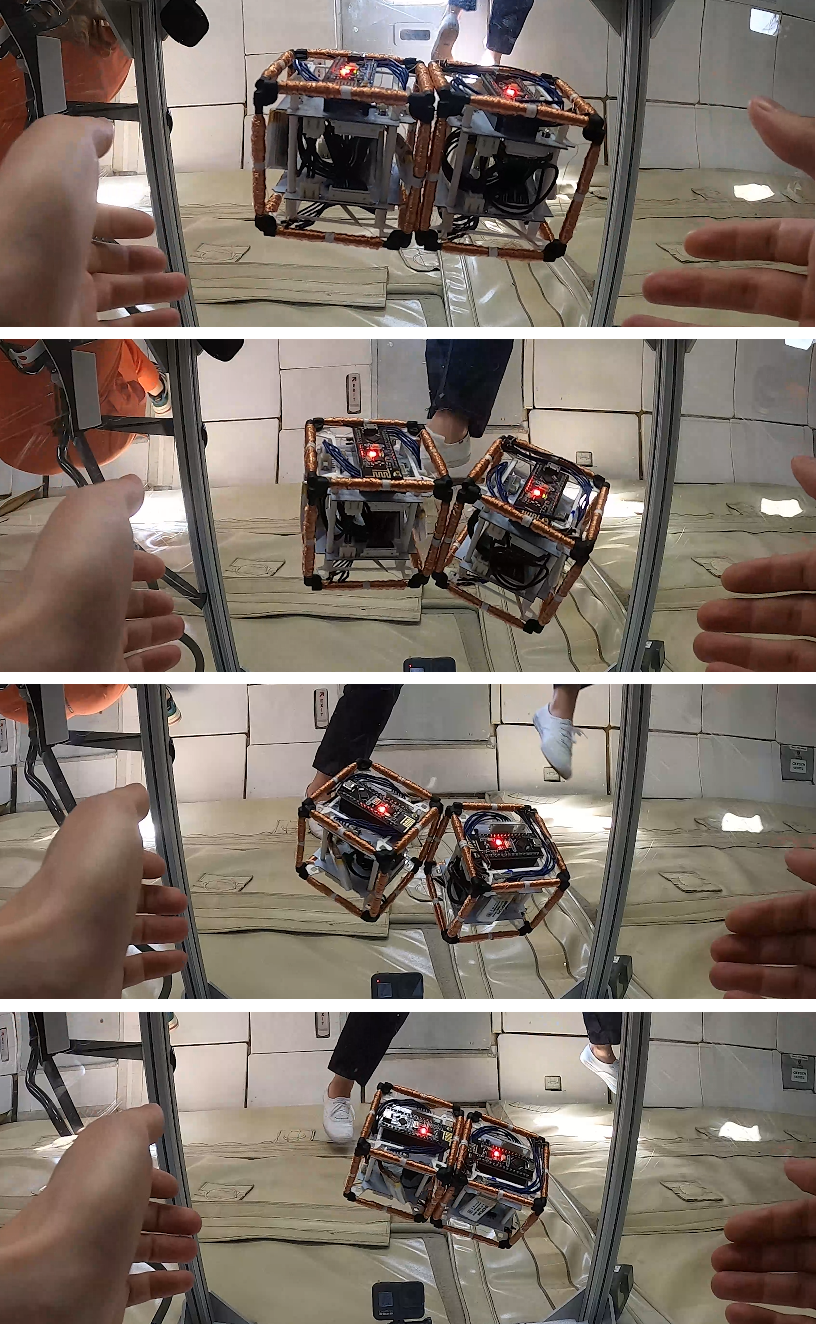}
    \caption{Parabolic flight}
    \label{fig:pivot_2}
\end{subfigure}
\caption{Two elements undergoing a pivoting maneuver (a) on an air table, a low friction surface used to simulate microgravity environments for 2D maneuvers, and (b) in microgravity aboard a parabolic flight for 3D maneuvers.  }
\label{Demos}
\end{figure}

The modules were deployed in microgravity on a parabolic flight (Figures \ref{fig:flight} and \ref{Demos}) to observe pivoting maneuvers unimpeded by kinematic constraints from the ground plane or sliding friction. The modules were deployed in a clear 460mm side length cubic polycarbonate box with two 120mm diameter arm holes in one side for the experimenter. Ten 15-second parabolas were flown. The first 7 parabolas were utilized for calibrating the steps and timing of the experimental protocol under microgravity conditions. Most significantly, the microgravity quality was found to vary significantly between and during individual parabolas, resulting in re-programming the modules for a shortened experiment window of approximately 4 seconds of stable microgravity available before free-falling modules would impact the polycarbonate enclosure. The last 3 parabolas were used to demonstrate the pivoting maneuver between two modules.
The procedure for each 15-second parabola involved the experimenter rising from a lying to a strapped-in kneeling position on the aircraft floor, inserting hands through arm holes, and positioning the cubes while waiting for stable microgravity to settle. A trigger button was pushed to wirelessly command execution of the pivoting maneuver, and a second button was pushed to power down all cubes. 

The pivoting maneuver was executed successfully each of the three times in 1.13 seconds, while electromagnets for launch, travel and catch phases performed correctly in all tests. However, small oscillations orthogonal to the axis of rotation were observed which led to the hinge disengaging early once completing the maneuver. This result was unique to microgravity experiments and unobserved on air tables as these kinematically constrain cubes in the ground plane, and arose due to imperfect alignments including protrusions of up to 0.6mm in our manually wound coils.

\subsection{Model Accuracy}

\begin{figure}[ht]
  \centering
  \includegraphics[width=0.85\columnwidth]{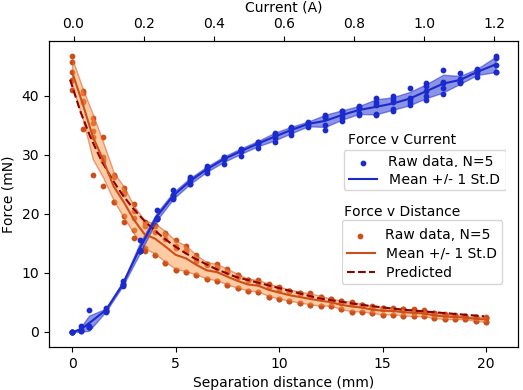}
  \caption{Force v current and electromagnet separation distance, showing raw data and mean $\pm$1 St.D for N=5. Predicted force-distance relationship from (\ref{eq_disc}) shown for comparison.}
  \label{fig:force_graph}
\end{figure}

We measured the force generated between two electromagnets using a Mettler Toledo Precision Balance ME203T/00 (10$\mu$N precision) while varying their separation distance and their currents, shown in Fig. \ref{fig:force_graph}. We held two electromagnets at a fixed separation distance of 0.5mm, varying the current applied to each electromagnet from 0A to 1.2A in 0.05A increments. We then held the electromagnets at a fixed current of 1.2A, varying the separation distance from 0.5mm to 20mm in 0.5mm increments. Each of these experiments were conducted 5 times. Measured data are plotted in raw form (scatter plots) and as mean$\pm$1 standard deviation (shaded plots) for both experiments. Using the force-current mean at I=1.2A to extract a characterised $\mu(1.2)$ of 874, we generated predicted values for force v distance (line graph) with our force model (\ref{eq_disc}), correlating well with measured data.

Finally, tentative results of our simple dynamical model including a simulation (supplemental video) agree qualitatively with the experiments, however as physical experiments only utilize a single force value (for $I$=$1.2A$), further work is required to capture more data to validate the model. %, including damping effects on the air table

\section{Discussion}

This section discusses the limitations of our current implementation and identifies avenues for future improvement. 

First, our cubes have to date been manually assembled and soldered, and each electromagnet hand-wound, resulting in imperfect alignments of electromagnets. While sufficient to successfully showcase the electromagnetic actuation framework on an airtable and in microgravity, Fig. \ref{fig:force_graph} reveals the degree to which these misalignments reduce attraction forces, at times disengaging the hinge. This resulted in a 6\% failure rate during air table traversals and minor instabilities in microgravity without the kinematic constraint of a ground plane; in the future, coils will be machine-wound and the assembly rigidized before deployment.

Our parameterized force model accurately predicts forces between electromagnets, supporting tailoring these actuators for different module designs. The concave downwards relationship in the force v current data (Fig. \ref{fig:force_graph}) indicates the onset of saturation of the ferrite cores in addition to potential heating effects at higher currents. Although current can be raised with higher voltage sources, this diminishes the returns on force that could potentially be achieved using greater currents; by exploiting larger cores with smaller currents, more favorable force-current relationships could be achieved with future prototypes. Nonetheless, the lack of opposing gravity moments in microgravity obviates the need for large forces, and our parabolic flight deployment required preparatory testing on an air table whose air pressure limited the total core mass.

Several avenues for future work present themselves. The success of our open-loop bang bang control suggests self-correction and robustness of the electromagnetic actuation method to small disturbances, however future work could support braking the traveling cube on impact via closed-loop control. Promising bases for control could include model-based strategies that leverage the pivoting cube model's tractable dynamics, combined with IMUs and electromagnet-based inductive sensing. To conserve power in a large scale system, future modules should embed passive permanent magnets in cube faces to form stable face-to-face bonds, or replace electromagnets with electropermanent \cite{an2008cube} or programmable \cite{hauser2020kubits} magnets. Further applications of our microgravity-adapted architecture, such as tangible or swarm user interfaces \cite{le2016zooids}, could be explored using air tables, caster wheels or low friction surfaces. Of equal interest would be to incorporate power electronics such as boost circuits for reconfiguring untethered cubes against gravity moments for terrestrial applications.

\section{Summary}

In this chapter, we introduced a modular assembly platform in the form of a cube-based reconfigurable robot. It utilizes an electromagnet-based actuation framework to reconfigure in three dimensions via pivoting. While a variety of actuation mechanisms for self-reconfigurable robots have been explored, they often suffer from cost, complexity, assembly and sizing requirements that prevent scaled production of such robots. To address this challenge, we developed an actuation mechanism based on electromagnets embedded into the edges of each cube  to interchangeably create identically or oppositely polarized electromagnet pairs, resulting in repulsive or attractive forces, respectively. By leveraging attraction for hinge formation, and repulsion to drive pivoting maneuvers, we showed how to reconfigure the robot by voxelizing it and actuating its constituent modules\textemdash termed \textit{Electrovoxels}\textemdash via \textit{electromagnetically actuated pivoting}. To demonstrate this, we developed fully untethered, three-dimensional self-reconfigurable robots and demonstrate 2D and 3D self-reconfiguration using pivot and traversal maneuvers on an air-table and in microgravity on a parabolic flight. We described the hardware design of our modules, its pivoting framework, our reconfiguration planning software, and an evaluation of the dynamical and electrical characteristics of our system to inform the design of modular assembly systems using scalable self-reconfigurable robots.

By developing a modular assembly platform that leveraged embedded actuation to assemble into target shapes, we demonstrated how to address the opportunity for in situ self-assembly in space environments. However, assembly using both machines and modular platforms require positioning parts with high resolution and constant power expenditure. For modular assembly in particular, like that demonstrated here and those developed in the modular robotics community more broadly, another key challenge arises. Assembling high resolution structures requires individual modules to be made increasingly \textit{numerous} and increasingly \textit{small}. Scaling these systems up in number and down in size scale creates a serious challenge for embedding electronics and actuators into individual parts. In the following chapter, we address this by introducing a method to design and build modularly assembled structures via folding, without the overhead of electronics or actuators.

%% file: Pullup.tex
\chapter{Modular Self-assembly via Folding}
\label{sec:Pullup}

In the preceding chapter, we introduced a method to self-assemble structures via reconfiguration using active control of individual modules. This architecture granted crucial control authority of the assembly procedure in the high-stakes microgravity environment, but to assemble high-resolution structures, the cost and algorithmic complexity of actively controlling large numbers of actuators becomes a serious challenge. In addition, while electromagnets are suitable for assembling structures in a microgravity environment, assembling a structure terrestrially in a gravity environment can place significant demands on actuator requirements. To address this, we introduce a folding-based modular assembly method that leverages two insights. First, by embedding folding instructions into the parts themselves, we can reduce the many folding steps into a single deployable trajectory actuated by 1 degree of freedom. Second, by reducing the actuated degrees of freedom to 1, we can outsource the actuation that folds the structure to lie offboard, and actuate this manually in the scope of this work. In this chapter, we introduce a method to rapidly create 3D geometries from 2D sheets using pull-up nets: a string routed through the planar faces which can be pulled by a user to fold the sheet into its target 3D structure. This provides a way to fold a sheet into its target shape using common string or nylon, using just \textit{a single actuated degree of freedom} controlled by a user.

\begin{figure}[H]
  \includegraphics[width=\textwidth]{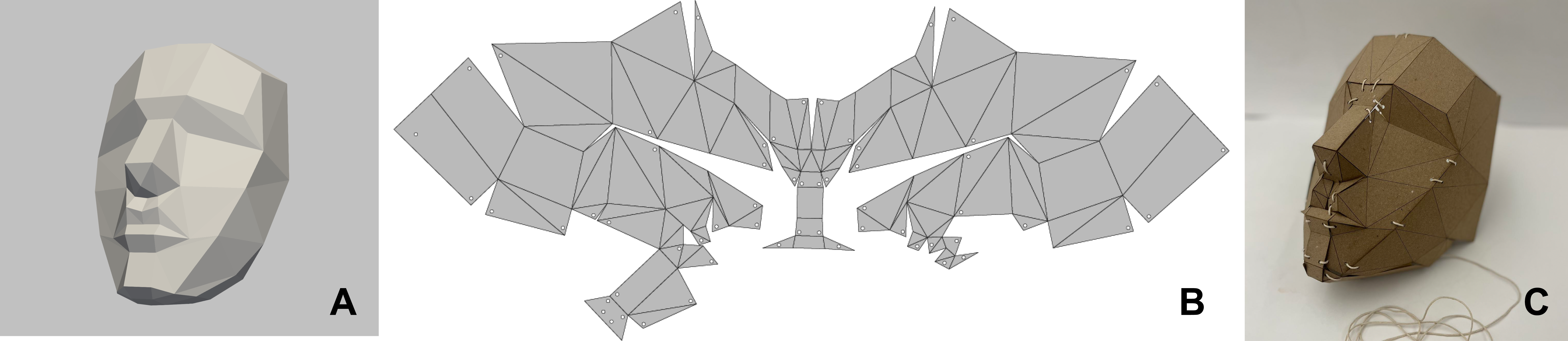}
  \centering
  \caption{Our digital fabrication pipeline enables users to rapidly prototype 3D objects. It features (A) a web-based tool for uploading and rendering 3D meshes. The tool executes our unpacking algorithm which unfolds the mesh into a (B) planar sheet partitioned into faces with through-holes in them. (C) We fabricate these unfolded geometries on a laser cutter, route thread through the holes, and pull the thread to fold the sheet into its target structure.}
  \label{fig: teaser}
\end{figure}

Methods for unfolding 3D polyhedral meshes into 2D sheets has been  studied across mathematics, physics, mechanical engineering and computer graphics \cite{takahashi2011optimized,demaine2005survey}. While prior work has explored pull-up nets to fold 3D geometries~\cite{meenan2008pull}, this was restricted to manual fabrication designs for the 5 platonic solids; in our work, we provide a tool to automatically compute these designs for all admissible geometries and introduce a digital fabrication pipeline to manufacture them. Given a 3D target structure, this process unfolds its 3D mesh into a planar 2D sheet populated with cutlines and throughholes. After laser-cutting the sheet and feeding thread through these throughholes to form a pull-up net, a user pulls on the thread to fold the sheet into the 3D structure. We introduce the fabrication process and build several prototypes demonstrating the method's ability to rapidly create a breadth of geometries suitable for low-fidelity prototyping across a wide range of applications. These include load-bearing stools, regular polyhedra and organic geometries such as the Stanford bunny, spanning a range of scales. We additionally develop a web-based simulation tool that translates 3D meshes into manufacturable 2D sheets, providing an existing set of 141 polyhedra for users to use as well as a feature allowing users to upload custom designs themselves.

This chapter is structured as follows. We introduce the algorithmic framework to our approach, then demonstrate their use in our user interface which permits uploading 3D meshes and simulating the unfolding and folding stages. We highlight the digital fabrication pipeline for fabricating structures based on the outputs of the user interface, and showcase a variety of fabricated geometries that span both aesthetic and functional applications in the space of low-fidelity rapid prototyping. Finally, we highlight our method's limitations and point to future avenues of research.

\begin{figure}[H]
	\centering
	\includegraphics[width=0.7\linewidth]{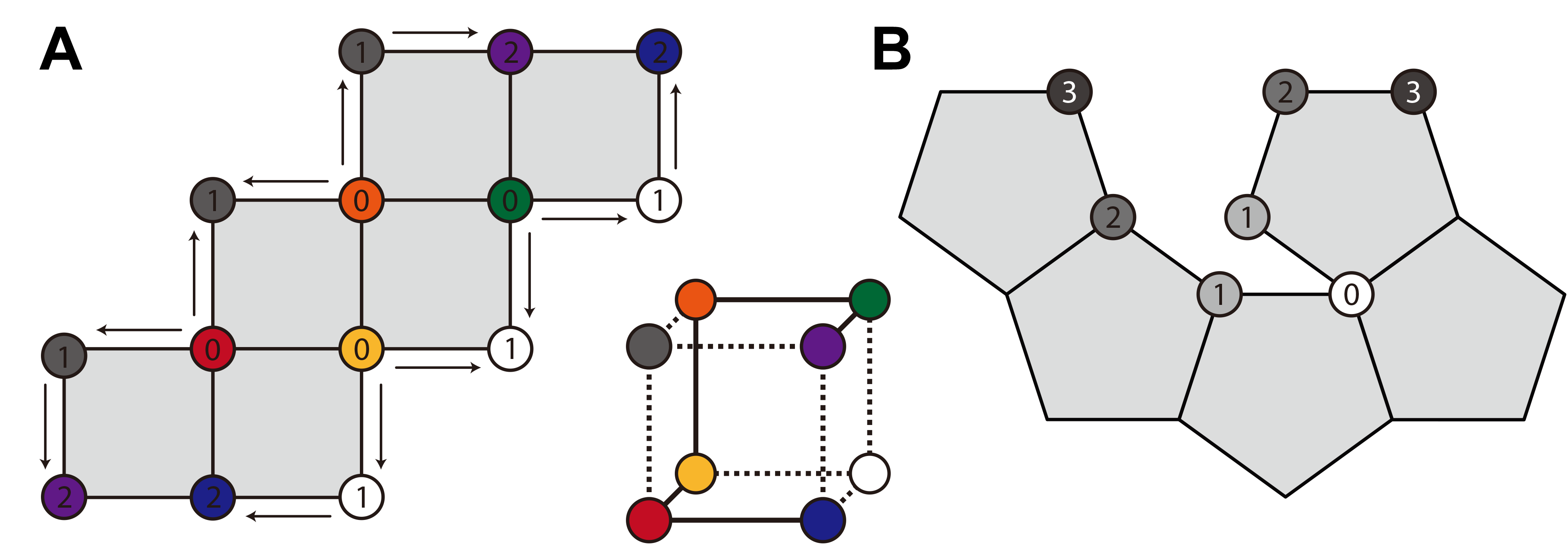}
	\caption{
        (A) Illustration of the algorithm to identify vertex sets that need to be joined by string. Color represents vertices of the net that are joined; numbers represent the rigidity depth of each vertex. Small arrows indicate the direction of the search from each vertex. Additionally, vertices of rigidity depth 1 do not need to be joined by string in order to assemble the final cube. In the 3D inset, cut edges are indicated by dotted lines.
        (B) The same concept illustrated with a portion of a dodecahedron net. The vertices of depth 1 and 2 do not necessarily have to join in order to determine a fully-joined geometric structure. Adding strings between all of these vertex pairs would require at least twice the amount of string, and would likely need to accommodate several sharp turns in its path, therefore increasing friction during assembly and complicating the assembly process. Depending on the assembled object's function and expected load, this may be a worthwhile tradeoff for extra rigidity in the final structure.
    }
	\label{fig:rigidity}
\end{figure}

\section{Algorithms}
This section introduces our algorithm for constructing an optimal net and string path for a given 3D mesh (Figure \ref{fig:rigidity}).

\textbf{Unfold the geometry} We begin by generating unfoldings of the mesh using a set of heuristic algorithms, including a steepest-edge cut tree, greatest-increase cut tree from Schlickenrieder~\cite{schlickenrieder1997nets} and a naive breadth-first unfolding from the largest face. If no non-overlapping nets are found, we randomly split the mesh in half (e.g., by cutting along edges that are nearest to a plane that passes through the center of mass) and repeat the search on each half.

\textbf{Join vertices} Next we use a breadth-first search to identify pairs of vertices that need to be joined (Figure \ref{fig:rigidity}): We identify vertices in the net that are already adjacent to $n$ connected faces, where the vertex is adjacent to $n$ faces in the 3D mesh. These vertices have a rigidity depth of 0. For each of these vertices: the vertex will be adjacent to two boundary edges in the net. The vertices at the other ends of these edges will need to be joined, and we sore these two vertices in a set and mark their rigidity depth as 1. Next, we join any sets with overlapping vertices. We finally iterate this joining procedure through the vertices of rigidity depth 1, until all vertices have been marked.

\textbf{Prune unwanted connections} We next remove unnecessarily joined vertices. The criteria for removing a vertex from a joined set are: (1) if removing the vertex from the set leaves only one vertex, then the other vertex must also be removed; (2) one of the vertices in the set must be connected by a single boundary edge to a joined vertex of greater rigidity depth; and (3) removing the vertex must leave each face with at least three joined vertices.

\textbf{Optimize string path} Using the final set of vertex sets to be joined, we lastly find the string path that passes through each pair of joined vertices which minimizes a combination of its turning angles and total length. If the mesh has been partitioned into multiple pieces, we optimize the string path on each piece separately.

\begin{figure} [H]
  \includegraphics[width= \linewidth]{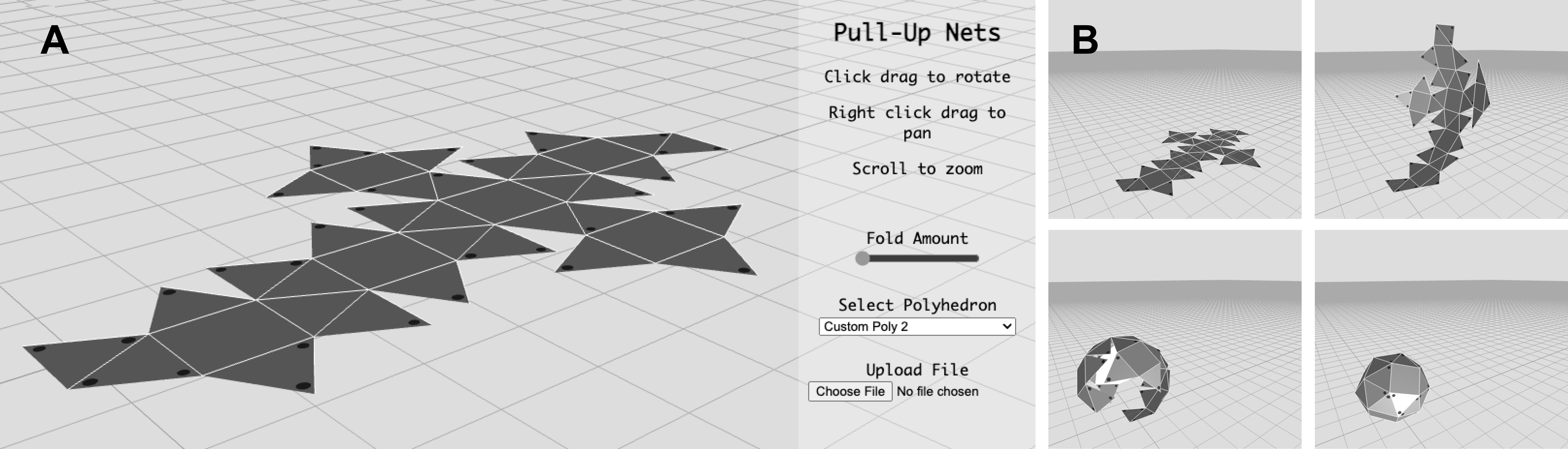}
  \caption{The web-based visualization tool.
  (A) The interface allows users to upload or select a geometry file, apply the folding algorithms to the geometry, and render the computed net with holes at each stage in its folding procedure.
  (B) Steps in the folding process can be visualized.}
  \label{fig:user_interface}
\end{figure}

\section{Web Tool}
We developed a web-based software tool based on our unfolding algorithm which allows users to upload their own custom 3D meshes from which to generate folded structures to fabricate. Users may upload either 3D mesh files in the OBJ file format, or existing nets in the SVG format; non-manifold meshes are rejected. In addition, an existing repertoire of 141 meshes that we successfully unfolded can be selected from a drop-down menu. The tool executes the algorithm on the chosen mesh and displays the mesh in the visualization window. Users can then render the folding procedure using a sliding toggle. 

The interface uses linear interpolation of edge folding angles to animate the assembly of the final shape. A ``base'' face is chosen arbitrarily to remain stationary. We use the Netlib polyhedra database~\cite{Netlib} to avoid calculating unfoldings for many well-known simple polyhedra, although the Netlib unfoldings may only be used for convex polyhedra that do not allow faces to intersect. Animation and controls are made using the \verb|THREE.js| Javascript library.

\section{Fabricated Structures and Applications}

\begin{figure}[H]
  \includegraphics[width=0.9\linewidth]{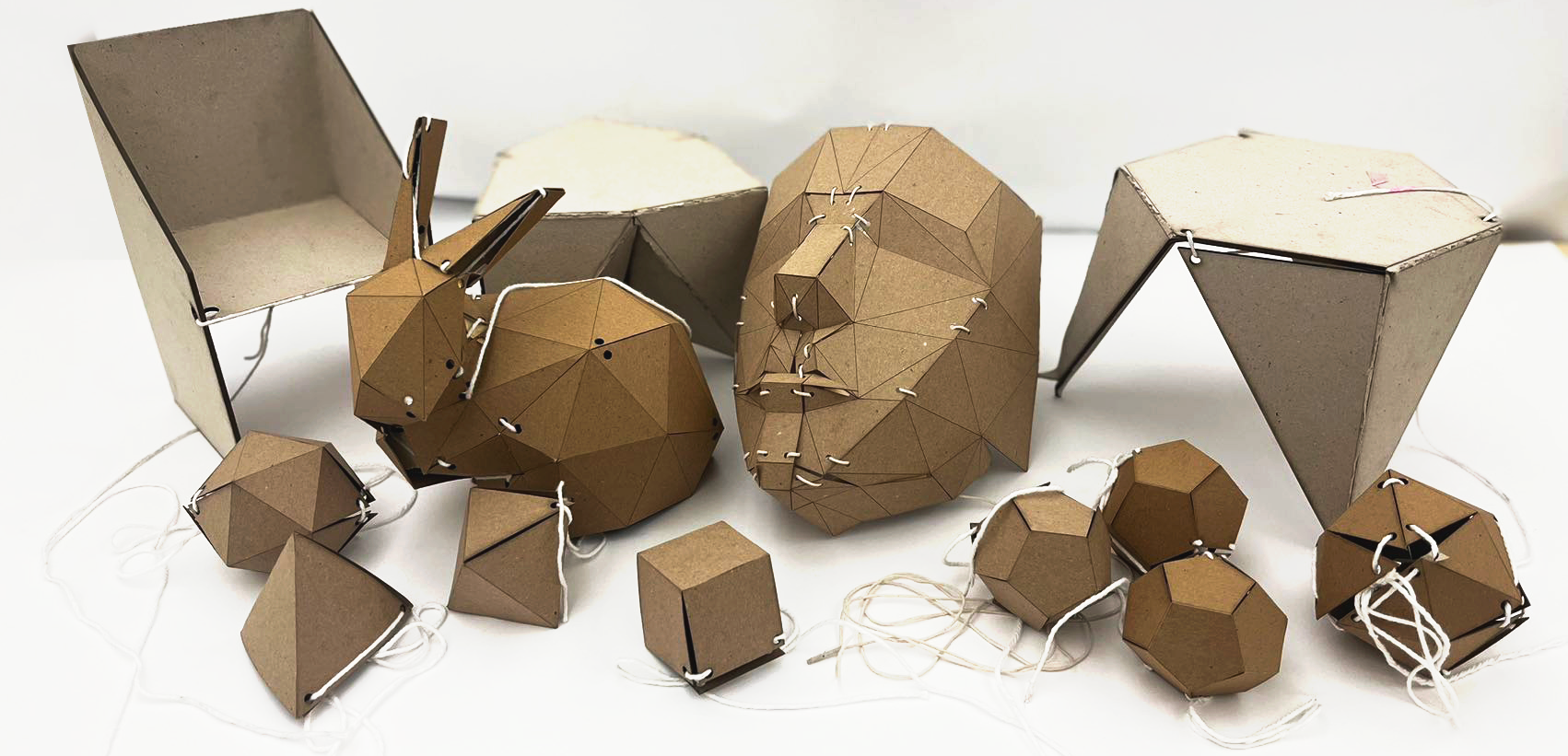}
  \caption{Structures fabricated using our pipeline range from regular polyhedra to organic shapes, and from load-bearing to aesthetic, across a range of scales, for applications that include VR, education and functional load-bearing contexts. }
  \label{fig: family}
\end{figure} 

%\begin{figure}[H]
  %\includegraphics[width=\linewidth]{stool.png}
  %\caption{Larger structures can be rapidly fabricated by cutting and folding, leveraging material stiffness to serve in load-bearing applications; here an assortment of stools and chairs.}
  %\label{fig:stools}
%\end{figure}

\begin{figure}[H]
  \includegraphics[width=0.99\linewidth]{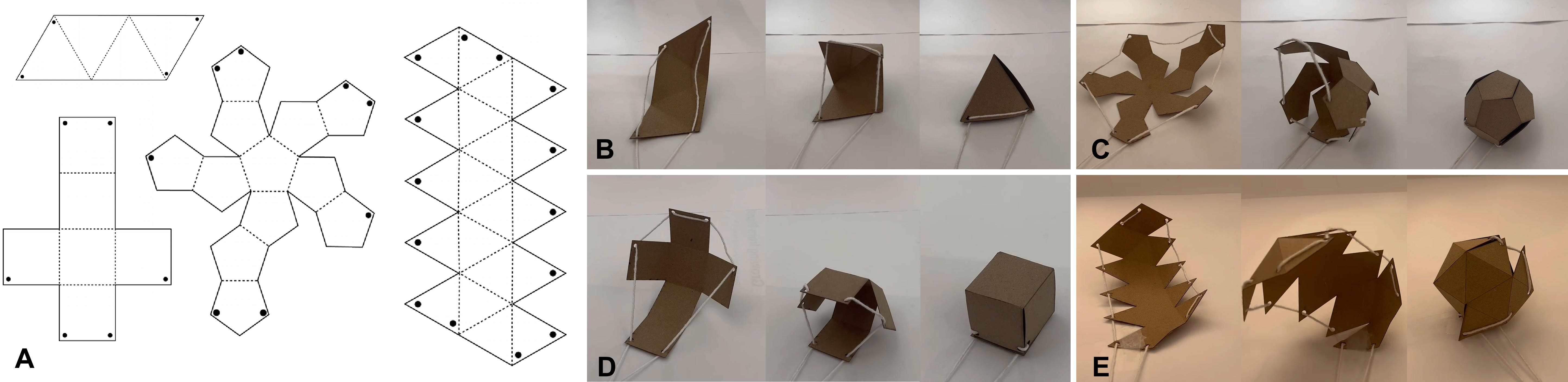}
  \caption{Fabricating polyhedra. (A) Unpacked layouts of four simple 3D solids. (B,C,D,E) Fabricated structures during folding.}
  \label{fig:fabricated-Polyhedra}
\end{figure}

\begin{figure}[H]
  \includegraphics[width=0.99\linewidth]{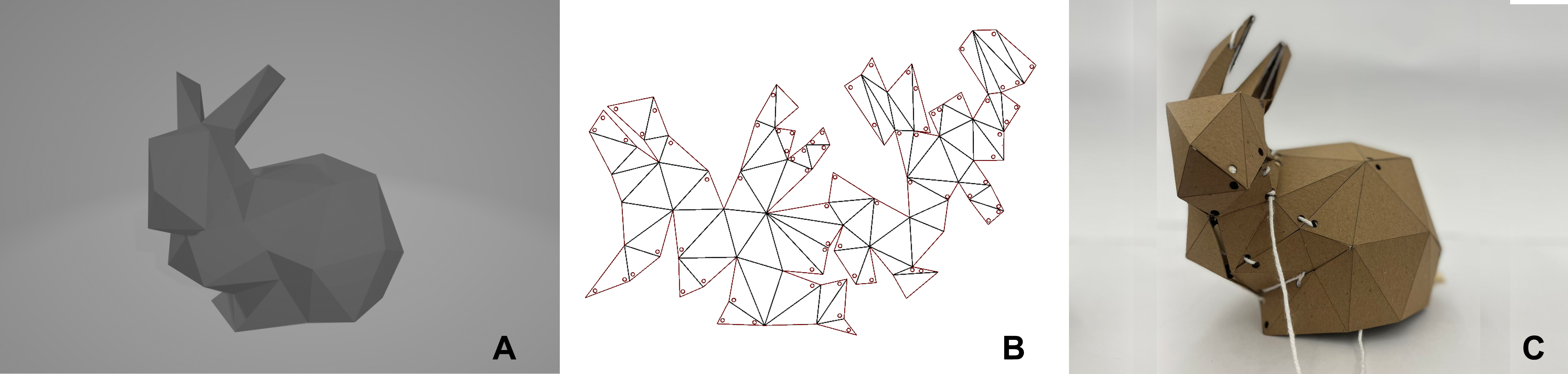}
  \caption{Fabricating organic shapes. (A) Rendering, (B) unpacking, and (C) fabrication of a Stanford bunny.}
  \label{fig:fabricated-Organic}
\end{figure}

\begin{figure}
  \includegraphics[width=\linewidth]{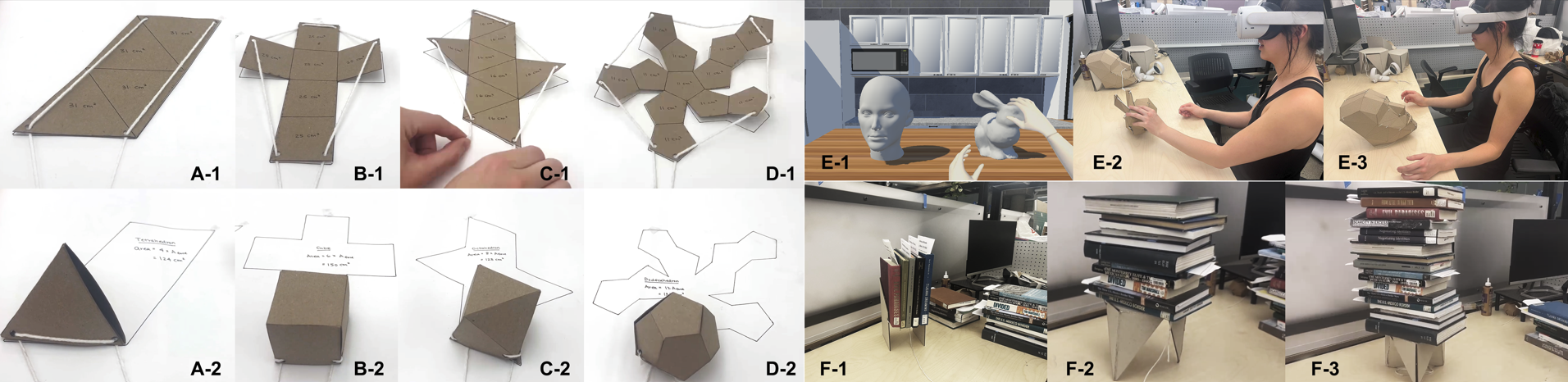}
  \caption{Applications include (A-D) Educational Toolkits to teach relationships between area and volume, (E) on-demand fabrication of low-resolution physical props for Virtual Reality, and (F) load-bearing applications including bookshelves or stools.}
  \label{fig:application}
\end{figure}

We fabricated a range of structures (Figure \ref{fig: family}) that are both aesthetic and load-bearing, including both traditional polyhedra (Figure \ref{fig:fabricated-Polyhedra}) and organic shapes (Figures \ref{fig:fabricated-Organic},\ref{fig: teaser}) across a range of scales to showcase the versatility of this technique in rapidly producing complex geometries. Possible applications include \textit{Educational Toolkits} (Figure \ref{fig:application}A-D), for example to teach children square-cube law relationships between surface area and volume, here with Platonic Solids. Visual dominance over some tactile cues~\cite{abtahi2018visuo} can be exploited to permit rapidly fabricating and folding \textit{low-resolution physical props for Virtual Reality} (VR) on-demand that are overlayed with high-resolution images (Figure \ref{fig:application}E). This manufacturing process is also suitable for \textit{load-bearing applications}, including bookshelves or small stools (Figure \ref{fig:application}F). These structures further benefit from their ability to be unfolded for compact storage or transport.

To build these structures, we used a laser cutter to cut the unfolded geometries produced by our web tool. We manually add cut holes where our algorithm selects them to appear, and then route string through these using our web tool rendering as a guide. We used 1/16'' chipboard and traditional knitting yarn to route through holes. Once routed, we pulled the yarn to fold the sheets into their target shapes.

% \begin{figure}[H]
%   \includegraphics[width=\linewidth]{process.png}
%   \caption{Laser cutting process and the 2D shapes that we get after fabrication(We also need to fold the mountains and valleys of the paper and insert the yarn into the paper's holes}
%   \label{fig: fabrication process}
% \end{figure}

\section{Discussion}

In this chapter, we introduced a method to rapidly create 3D geometries by folding 2D sheets via pull-up nets. Given a 3D structure, we unfold its mesh into a planar 2D sheet using heuristic algorithms and populate these with cutlines and throughholes. We develop a web-based simulation tool that translates users' 3D meshes into manufacturable 2D sheets. After laser-cutting the sheet and feeding thread through these throughholes to form a pull-up net, pulling the thread will fold the sheet into the 3D structure using a single degree of freedom. We introduce the fabrication process and build a variety of prototypes demonstrating the method's ability to rapidly create a breadth of geometries suitable for low-fidelity prototyping that are both load-bearing and aesthetic across a range of scales. Having presented our digital fabrication pipeline, we discuss its limitations and avenues for future work.

Each step in our algorithm may be computationally costly without optimization, and is ideally suited for smaller meshes with fewer than 100 elements. As noted, not all meshes may be unfolded into a single flat piece~\cite{bern1999ununfoldable,reitebuch2019discrete}, an issue exacerbated for meshes with excessively hyperbolic vertices (where all adjacent faces meet with angles summing to more than $2\pi$), common in geometries where convex polyhedral faces are extruded, or in periodic minimal surfaces. Although other assembly methods for such geometries exist \cite{overvelde2016three}, we do not yet consider them suitable for our method.

Friction limits the maximum length of string used, the severity of turning angles as it passes through holes, and the number of holes present. Like a shoelace woven through holes at tight angles, friction may prevent tension at one end of the string from diffusing equally to all other parts. During assembly of polyhedra with many faces, the string may therefore require tension at several different points along its length to decrease frictional resistance.

The interface generates vertex locations for routing string, but does not yet generate ready fabrication files with holes at vertex locations, or indicate how to route string through each vertex, though this has been straightforward to interpret for our geometries produced so far. Further work will seek to automate this for fabrication on a laser cutter, and to implement basic mesh editing features for coarsening, triangulation, and refinement. 

\begin{figure}[H]
  \includegraphics[width= 0.99\linewidth]{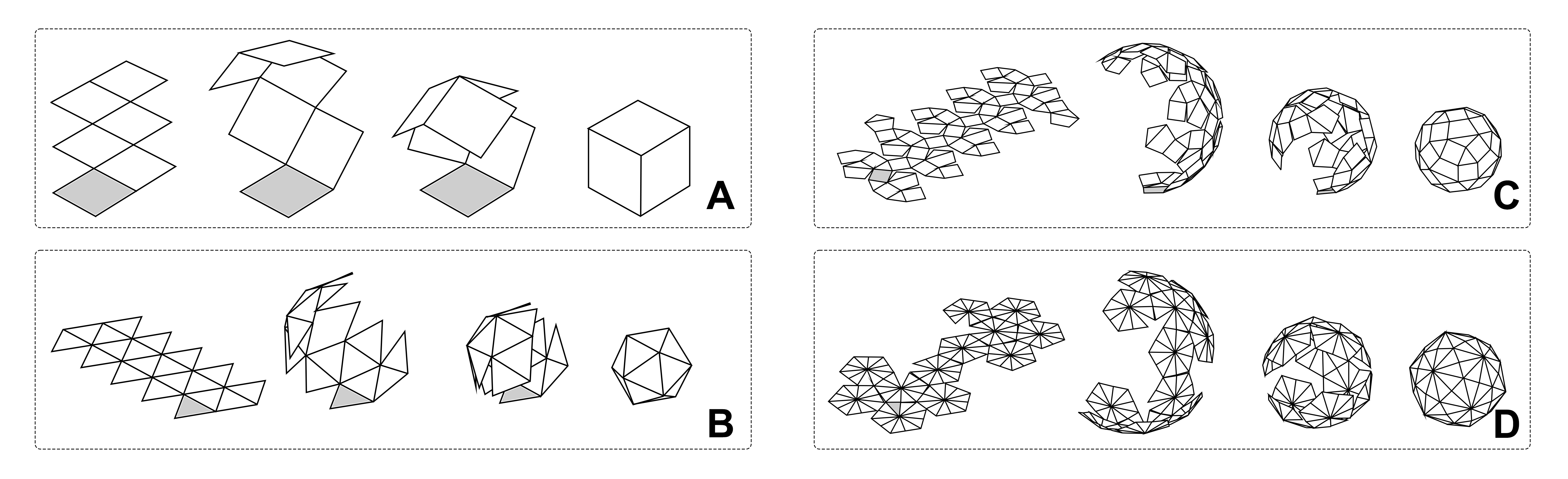}
  \caption{Examples of folding 3D polyhedra using pull-up nets: By convention, one surface (in grey) of the unfolding nets remains fixed during the folding procedure, helping stabilize the structure while users actuate the fold.}
  \label{fig: Examples}
\end{figure}

The constraint of small-complexity solids is enforced physically as structural rigidity decreases as more cuts are introduced. Due to the large number of cuts needed for a complete unfolding (for a closed mesh, the cuts must reach every non-flat vertex), the complexity of some meshes may compromise its structure's stability, further descriptions of which we leave to future works. Additionally, there is a trade-off between reducing the number of vertices explicitly joined by the string path and reducing the rigidity of the final structure (Figure~\ref{fig:rigidity}). Another trade-off lies in the material thickness: thicker material exacerbates folding difficulty while thinner material is less stiff and tolerant to load-bearing. In future work we will seek to physically evaluate the load-bearing capacities of our structures.

Finally, a single 3D mesh can often be unfolded into different 2D surfaces (Figure \ref{fig: Examples}C). While we optimize string path by length and turning angles, we found that our algorithm's selected design is not always the easiest to pull up; more work is required to understand how to optimize for ease of fabrication.

% \begin{figure}[H]
%   \includegraphics[width=0.7\linewidth]{extruded shapes.png}
%   \caption{Examples of certain meshes that cannot be unfolded into a non-overlapping net.}
%   \label{fig:limitations}
% \end{figure}

\section{Summary}

In this chapter, we introduced a method to rapidly create 3D geometries from 2D sheets using pull-up nets. We developed an algorithm run by our web-based software tool that allows users to upload 3D meshes and generate their unfolded geometries. We show how to fabricate these on a commercial laser cutter and route string through the faces which are pulled by a user to fold the sheet into its target 3D structure. This lets users rapidly create large 3D geometries using 2D fabrication machines using just a single actuated degree of freedom. We fabricated a variety of polyhedra and organic structures, highlighted applications, and laid out the work's current limitations and avenues for future work. We introduced a low-fidelity rapid prototyping option for 3D structures that is simple, inexpensive and rapid, obviating the need for significant design or fabrication expertise, and serves as a compelling alternative to more complicated material actuator-based folding paradigms. However, a limitation with this approach is that the target structure must be known ahead of time, and a particular sheet can be designed to fold into only one specific target shape. In the following chapter, we address this by introducing a method to program magnetic materials themselves to both align and assemble parts, and that allows these materials to be reprogrammed for new target shapes.

%% file: SelectiveMixels.tex
\chapter{Programmable Materials}
\label{sec:SelectiveMixels}

In the previous chapters, we introduced a method to automate assembly of structures at the part (or "module") level. We identified the challenge of constructing actively-controlled reconfigurable systems at small scales and in high numbers, due to the difficulty in miniaturizing and embedding large numbers of discrete electronic and electromechanical components. We also identified the challenge facing new self-folded architectures in their inability to be re-programmed for new target shapes. In this chapter, we address this by formulating a method to program magnetic materials selectively in order to achieve passive assembly under stochastic agitation, without the need for electronics or actuators, in a way that is re-programmable.

Stochastic self-assembly~\cite{tibbits2012self} requires no local control and is governed instead only by global parameters such as excitation magnitude. Stochastic assembly sacrifices efficiency and predictability for advantages in cost, complexity and scale; by enabling the environment to actuate reconfiguration, it trades off deterministic assembly times of individual modules for statistical assembly rates of the collective. To assemble stochastically, modules require pre-programming to enforce correct mating during random collisions with their intended mate. This programmed specificity between pairs of mating faces is typically achieved via minimization of free surface energy via topology\cite{hacohen2015meshing}, wettability \cite{bowden1997self}, magnetic forces \cite{lu2021enumeration} or electrostatic \cite{grzybowski2003electrostatic} interaction. While the wider scientific community has often been interested in constraining the self-assembly problem to 2D, for instance by using a shaker table \cite{jilek2020centimeter}, roboticists have leveraged liquid tanks to study assembly in 3D. Fluidic assembly at the mesoscale has become a particularly widely studied problem in robotics \cite{tolley2008dynamically,tolley2010fluidic,tolley2011programmable,krishnan2008increased,kalontarov2010hydrodynamically, zykov2007experiment}. Existing stochastically self-assembling modules typically include two features to enable assembly: first, embedded magnets that generate near-field forces to bring modules close, and second, selective geometry on module faces that encodes the specificity to only permit bonds between mating pairs \cite{jilek2021towards,hacohen2015meshing,jilek2020centimeter, tsutsumi2007multistate}. However, three key challenges remain for the development of stochastic self-assembling systems: (1) \textit{scalability} that shows how modules can be made both numerous and small; (2) \textit{selectivity guarantees} that help bound module misassembly; and (3) \textit{reconfigurability} that let modules acquire different target shapes. 

\begin{figure}[ht]
  \centering
  \includegraphics[width=1\columnwidth]{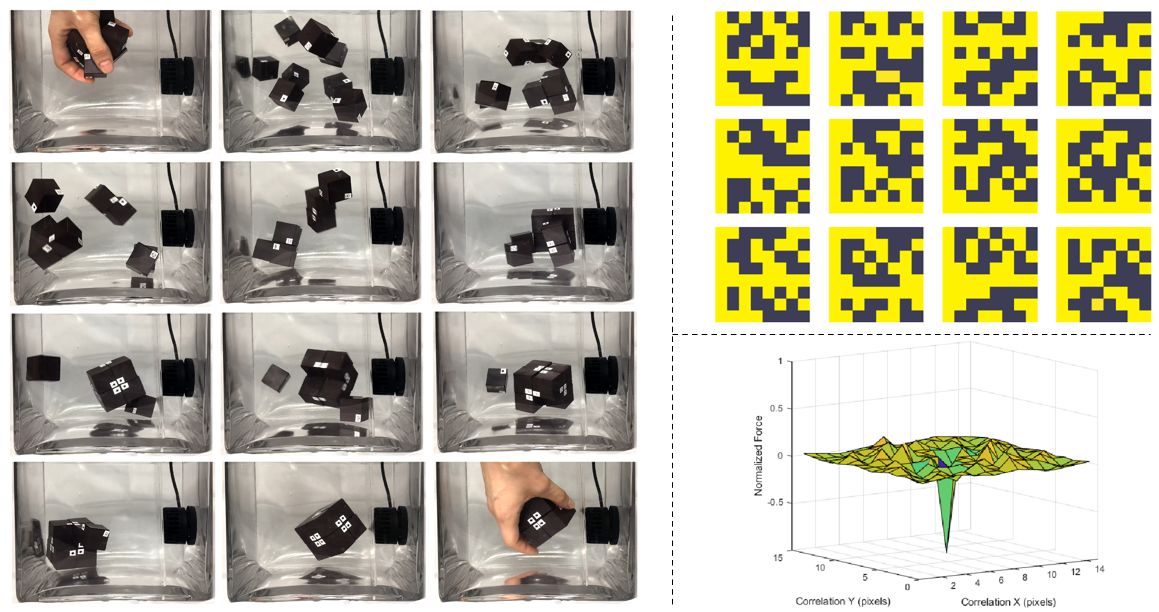}
\caption{(Left) We enable stochastic self assembly using inexpensive (\$0.23) cubic modules. (Above, right) We accomplish this by magnetically programming module faces with uniquely mating pairs of encodings based on Hadamard matrices, and show bounds on their performance. (Below, right) Key to the modules' success is their ability to attract strongly to their mates, while remaining agnostic in all other translations, rotations, and to non-mating modules.}
  \label{fig:teaser}
\end{figure}

\textit{(1) Scalability:} To assemble arbitrarily complex geometries, encodings for 3D modules must support selectivity great enough to permit uniquely mating pairs of modules in the hundreds or even thousands. In addition, modules must be inexpensive and simple enough to be fabricated in these quantities. Due to this dual problem, a significant corpus of previous research demonstrates the stochastic self-assembly for tiled 2D arrays, such as chessboards\cite{grzybowski2003electrostatic,jilek2021towards,miyashita2009influence}, with only two module types where each module is selective to entirely half of all modules in the set. On the other hand, the individual fabrication of heterogeneous module topologies with manually embedded permanent magnets poses a significant challenge to scalable fabrication. 

\textit{(2) Selectivity guarantees:} Because magnet arrangements typically used to generate near-field forces are poorly discriminating to each other, this framework often leads to misassemblies, because near field forces between both mating and non-mating face magnets are equally strong. In addition, protruding geometrical features used for selectivity can lead to "jamming" by obstructing assembly paths \cite{jilek2021towards}, and bounds on the expected misassembly rate between geometrically dissimilar modules may be difficult to compute.

\textit{(3) Reconfigurability:} To date, structures self-assembled at the mesoscale are not reconfigurable. Because module selectivity is achieved by fabricating individualized geometries, any set of fabricated modules encode only a single target shape (or single set, for non-deterministic encodings). Such modules are therefore unable to be "re-programmed" to self-assemble new target shapes: new shapes require a unique batch of modules to be fabricated from scratch, inhibiting their utility and increasing their potential unit cost.

\begin{figure*}[ht]
  \centering
  \includegraphics[width=1\textwidth]{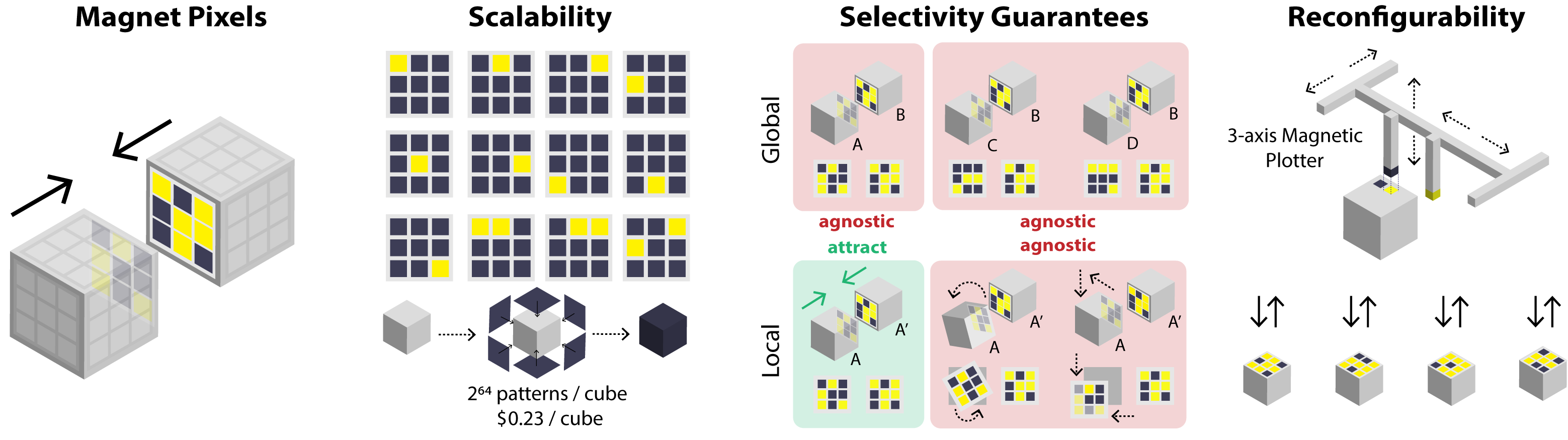}
\caption{Overview of our stochastic self assembly approach. (Left) \textit{Magnetic Pixels:} Cubic modules are programmed with matrices of magnetic pixels. These permit module faces to mate selectively to assemble target geometries. 3x3 Matrices shown for clarity; our modules utilize 8x8. (Center left) \textit{Scalability:} Our binary-valued, 8x8 matrices can encode 2$^{64}$ module faces with unique permutations, and modules are inexpensive (\$0.23). (Center right) \textit{Selectivity Guarantees:} We leverage Hadamard matrices to encode magnetic pixels on faces with 2 criteria. "Locally", mating faces attract in only one configuration; "globally", non-mating faces attract in no configuration. (Right) \textit{Reconfigurability:} Encodings are "programmed" as magnetic pixels using magnets installed on a 3-axis CNC. Modules can be re-programmed to self assemble into new target geometries.}
  \label{fig:overview}
\end{figure*}

In this chapter, we introduce a method to design and "program" selective encodings~\cite{nisser2022stochastic} onto cubic modules in a way that addresses all three challenges above (Fig. \ref{fig:overview}). We program module faces with patterns of magnetic pixels which can attract or repulse pixels of another face (Fig. \ref{fig:overview} left), and if the number of pixels in attraction match those in repulsion, the faces are agnostic to each other. Using this observation, we show how to program modules with encodings that allow them to selectively mate with other cubes to self assemble in a unique target structure. Formulated as matrices, we demonstrate the number of unique encodings that can be programmed given criteria on attraction and agnosticism. 

Our modules consist of PLA cubes, with squares of soft magnetic material (inexpensive COTS fridge magnets) bonded to their sides (Fig. \ref{fig:overview} center left). The encodings on these faces generate both near-field forces \textit{and} selectivity. This selectivity lets us use homogeneous cubic modules with planar faces, making modules both non-jamming and inexpensive to mass fabricate across scales. Key to our approach is the generation of selective encodings, for which we leverage Hadamard matrices (Fig. \ref{fig:overview} center right), and similar procedures may have been used to create industrial-grade magnets with tailored selectivity properties in industry~\cite{CorrMag}. The two polarities of the magnetic pixels we encode onto faces correspond to elements of these binary-valued matrices. Our matrices enforce two key criteria. A local criterion dictates behavior between faces intended to mate. If every pixel on one face, face A, is magnetically opposite to pixels on another, then we call that face its mate, A'. These faces thus form a maximally attractive mating pair. Our matrix pairs are designed to exhibit maximal attraction in this one configuration, while placing guarantees on agnosticism in all other translations and rotations. A global criterion dictates behavior between two faces not intended to mate. For this, we ensure that any given matrix pair, A and A', cannot mate with any other faces B, B', C, C' etc., in any configuration, and we place guarantees on this agnosticism. We further demonstrate how to program module faces using a magnetic plotter consisting of an electromagnet affixed to a 3-axis CNC. Crucially, these soft magnetic faces are re-programmable, and thus modules can be repeatedly re-programmed with new encodings in order to self-assemble into new target geometries in 3 dimensions. Finally, we design and fabricate a set of 8 modules and demonstrate their stochastic self-assembly in a water tank. 

The chapter is structured as follows. We begin by introducing a procedure to generate selective magnetic encodings, and derive bounds on the number of modules that can be utilized given a threshold on agnosticism between programmed encodings. We describe the physical modules themselves, and the magnetic plotting technique used to program them. We demonstrate our ability to make predictions with regard to the attraction and agnosticism between various magnetic encodings, and verify these experimentally. Finally, we demonstrate stochastic self-assembly of our system using 8 modules in a water tank.

\section{Encoding generation}

This section describes how we generate encodings that satisfy the global and local criteria given above. Our encodings are based on Hadamard matrices, whose unique properties have lent their use to applications including Code Division Multiple Access and error correcting code. The Hadamard matrix is a square matrix whose rows are all mutually orthogonal and whose elements are either $1$ or $-1$ (here representing magnetic pixel polarization). As a consequence of its row orthogonality, it follows that its columns are mutually orthogonal too. As a result, the dot product of any pair of rows, or any pair of columns, is equal to 0. In addition, if a single row or column is multiplied by -1 before taking the dot product, the product remains 0. 

Defining the mate of matrix A as A'$=$-A, the Hadamard product (\ref{eq_hada_prod}) between a matrix A of order N and its mate, normalized by the number of elements N$^2$, is -1. This implies maximum magnetic attraction. Conversely, the normalized Hadamard product between A and itself is +1, connoting maximal repulsion. Let $S_G$ be a global score indicating the maximum attraction between non-mating pairs A and B in any configuration, and let $S_L$ be the local score indicating maximum attraction between mating pairs A and A' \textit{in all wrong configurations}. To permit self assembly, satisfying the global and local agnosticism criteria requires the force enacted by our fluid $F_f$ to satisfy $-1 < F_f < min(S_L,S_G)$ in order to both break apart unintended misassemblies and allow correct assemblies to survive. Because controlling the fluidic force is challenging, the goal is then to maximize $min(S_L,S_G)$, making them maximally agnostic, in order to place $F_f$ between these values. 

Now, the row and column orthogonality described above yields that taking the Hadamard product between A and A' becomes 0 if one matrix is translated in only X or only Y, yielding maximal agnosticism. However, agnosticism is not guaranteed for matrices translated in both X \textit{and} Y, or if they are rotated. To find matrices that maximize agnosticism in these configurations, we perform two searches. The first search shows that Hadamard matrices perform optimally among the set of all square matrices. The second search identifies the number of matrices that can be generated for a given bound on agnosticism performance.

\begin{equation}
   (A \odot B)_{ij} = (A)_{ij} (B)_{ij}
\label{eq_hada_prod}
\end{equation}

\subsection{Matrix search}

We generate every binary-valued (values 1 and -1) matrix A of order N=4 (yielding $2^{16}$ matrices), and compute the cross-correlation between A and A', which is equivalent to their Hadamard product in every translation ($2N-1$ values) and also compute their Hadamard product in rotations of 0$^{\circ}$, 90$^{\circ}$, 180$^{\circ}$, and 270$^{\circ}$ (4 values, upper-bounded by computational resources). These values are used to assess the local criterion. We then check which of these are Hadamard matrices by evaluating which matrices A satisfy $AA^T = N(I_N)$, finding 768 matrices (1.2\%) out of the total 65,536 ($2^{16}$). We then rank every matrix by its most attractive (most negative) value for all translations and rotations (excluding its one mating configuration of -1) to give its local criterion score $S_L$. An optimal $S_L$ score is 0, implying perfect agnosticism in all configurations besides its mating configuration.

Assessing a set of matrices' global criterion score, $S_G$, requires finding the worst (most negative) Hadamard product in translation and rotation of every combination of matrices in the set. This is the power set of matrices, $2^{65,536}$ combinations for the matrices above, a computationally intractable search; searching for all possible cliques (maximal complete subgraphs) in a graph is an NP-complete problem. However, any guarantee of a matrix' performance in agnosticism is given by $min(S_L,S_G)$. Therefore, to narrow our search, we build sets by searching those with best $S_L$ first, stopping our search once $S_G=S_L$. We consider a graph $G$ where vertices represent these matrices, and where two vertices ($V_1, V_2$) are adjacent if their Hadamard product (in translation and rotation) is above some $S_G$. Hence we reduce the problem of searching for mutually compatible sets in the power set to a search for cliques in G. We determine $S_G$ for progressively larger clique sizes of the 65,536 matrices using a variation of the algorithm by Bron and Kerbosch~\cite{bron1973algorithm,hagberg2008exploring}, and find that given a $S_G$ and $S_L$, the set of Hadamard matrices produce a clique size equivalent to the set of all possible matrices. Thus, the maximal clique search, which exhibits exponential complexity, is reduced to the set of Hadamard matrices (a small fraction, 1.2\%, of the total matrices possible) and still finds maximally sized cliques given some $S_G$ and $S_L$. 

Initial experiments revealed that matrices of order N=4 were unable to produce adequately agnostic values for $S_G$ and $S_L$ for clique sizes greater than 5 such that our stochastic fluid could discriminate between correct and incorrect mates. However, with a goal to assemble an 8-module system into a "meta cube" (Fig. \ref{fig:assembly}), we require a minimum clique size of 12. We therefore use our result regarding maximal cliques to search across Hadamard matrices of order N=8.

\subsection{Generating Hadamard Matrices}

We generate a normalized Hadamard matrix of order N=8 using the recursive procedure described below: let $H$ be a Hadamard matrix of order N. We can use this to create the partitioned matrix of order $2N$ shown in (\ref{eq1}).
% As a consequence, the matrix can be arranged such that the first row and column are populated entirely with +1; in this form, it is said to be normalized. 

\begin{figure}[ht]
  \centering
  \includegraphics[width=0.68\columnwidth]{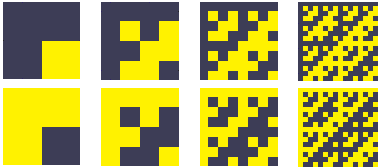}
\caption{(Above) Normalized, naturally ordered Hadamard matrices generated in a procession of orders N of 2, 4, 8 and 16. Binary values of 1 and -1 are represented as dark and light pixels, respectively. (Below) Their mates.}
  \label{fig:hadamard}
\end{figure}

\begin{figure}[ht]
  \centering
  \includegraphics[width=0.78\columnwidth]{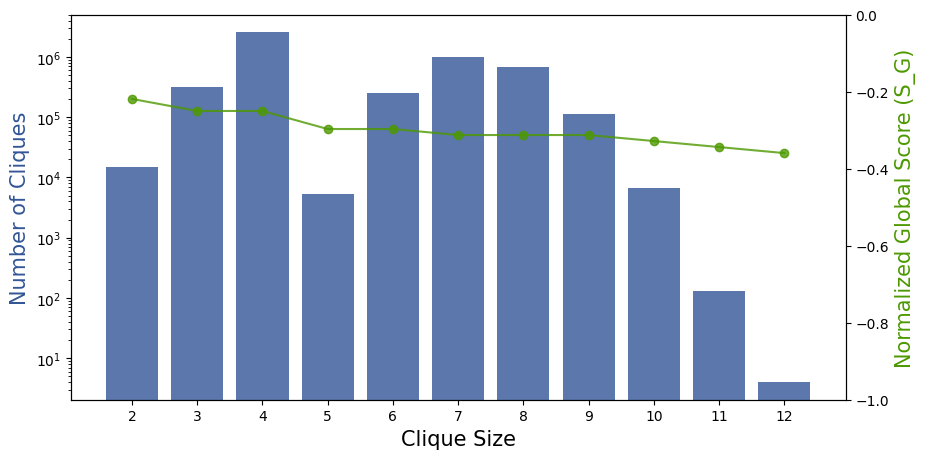}
  \caption{The size of a clique (the number of mutually compatible encodings) related to both 1) the number of such cliques, and 2) its global agnosticism score $S_G$}
  \label{fig:clique_size_score}
\end{figure}

\begin{equation}
\begin{bmatrix}
H & H \\ H & -H
\end{bmatrix}
\label{eq1}
\end{equation}
We can apply this rule generally for higher orders using:
 \begin{equation}
H_{2^k} = 
\begin{bmatrix}
H_{2^{k-1}}  &  H_{2^{k-1}} \\ 
H_{2^{k-1}}  & -H_{2^{k-1}}
\end{bmatrix}
\label{eq2}
\end{equation}
Initializing $H_1$ as 1, we generate $H_8$ and see that the Hadamards produced in this construction are fractal (Fig. \ref{fig:hadamard}). We then generate all $8!$ matrices that are permutations of its rows. As the rows of $H_8$ are orthogonal, the rows of all its permutations remain orthogonal and Hadamard.

To meet our goal of finding a clique of size 12, we seed a threshhold for both $S_L$ and $S_G$ of -0.2, and iteratively search for maximal cliques. We continue to lower the threshold of $S_G$ by 0.02 until we grow the maximal clique size to 12. Fig. \ref{fig:clique_size_score} illustrates the relationship between maximum clique size, its mutual $S_G$ score, and the number of cliques of that size. We select the 12 matrices from one of the 4 size-12 cliques to program and assemble a meta cube, exhibiting a combined agnosticism score of $min(S_G,S_L)=S_G=-0.36$.

\section{Magnetic Plotter}

\begin{figure}[ht]
  \centering
  \includegraphics[width=0.99\columnwidth]{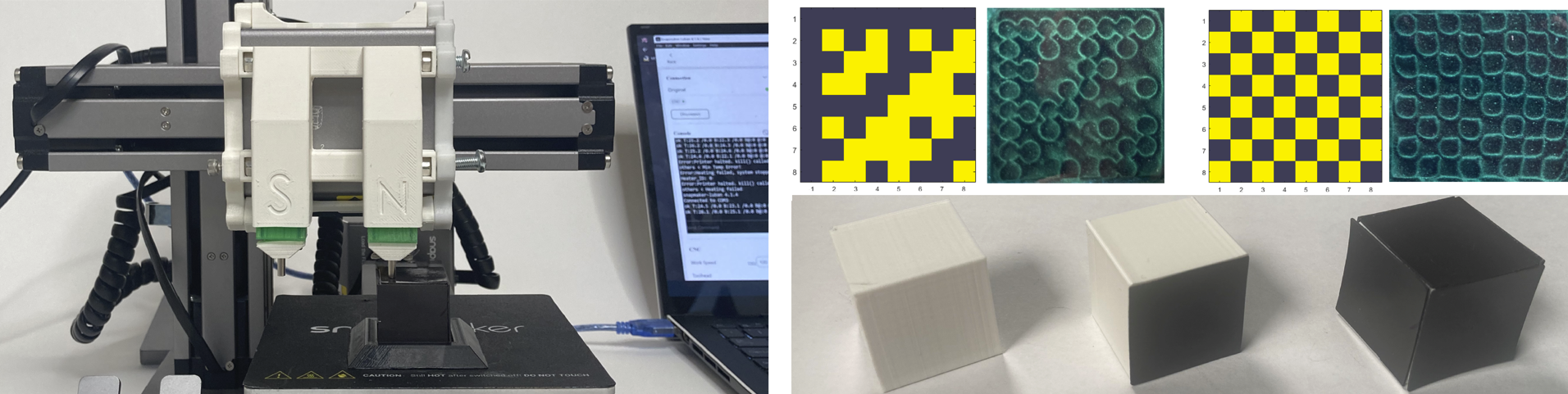}
  \caption{(Above) First prototype of the magnetic plotter programming a module. (Middle) Simulated and plotted matrices viewed through magnetic viewing film; (Left) A normalized order-8 Hadamard; (Right) A checkerboard. Dark and light pixels in simulation represent opposite magnetic polarities. The same material was reprogrammed to produce these patterns in turn. (Below) Module fabrication. (Left) a white PLA cube is covered with (middle) squares of black soft magnetic material; (Right) once all six squares are bonded, it is ready for programming.}
  \label{fig:plotter}
\end{figure}

\begin{figure}[ht]
  \centering
  \includegraphics[width=0.97\columnwidth]{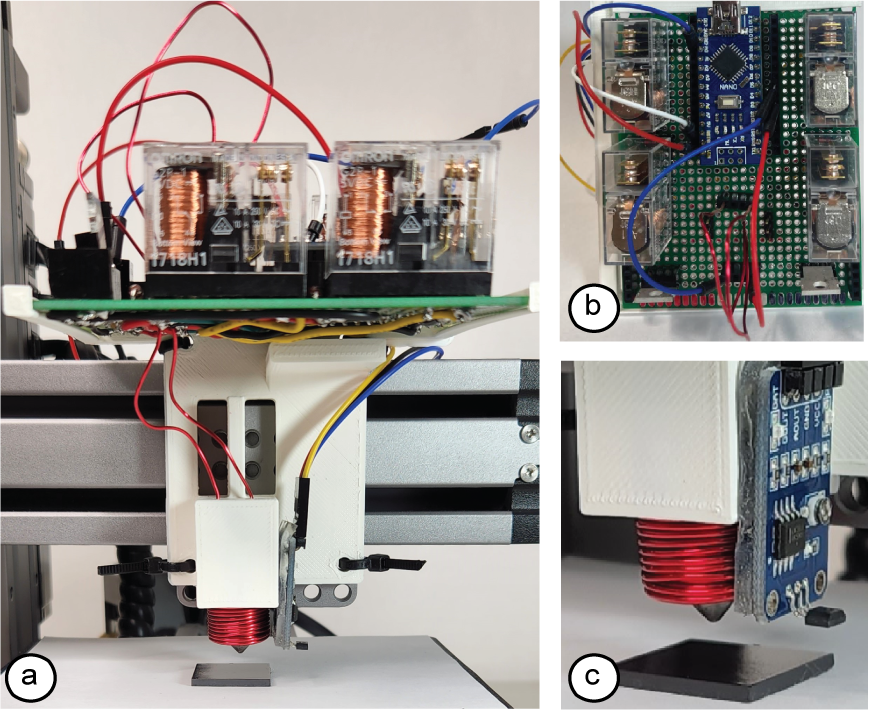}
  \caption{Magnetic plotter: (a)~add-on mounted onto a 3-axis CNC. (b)~add-on electronics viewed from above. (c)~Close-up of the plotting end effector, consisting of an cone-tipped electromagnet for writing and a hall effect sensor for reading magnetic programs.}
  \label{fig:Mixelplotter}
\end{figure}

We built a magnetic plotter to stamp our cubes' soft magnetic faces with magnetic encodings. A first prototype of the plotter is shown in Fig. \ref{fig:plotter} (left) together with programmed cubes and two encodings. This plotter consisted of a pair of oppositely polarized permanent magnets (3mm diameter, 6mm length) installed in a housing mounted onto a 3-axis CNC (SnapMaker 3-in-1). Each magnet thereby plots opposite pixel values, where binary-valued pixels correspond to oppositely polarized regions\textemdash magnetic pixels\textemdash of soft magnetic material. As a result, this plotter could strictly program faces with binary-valued encodings of either North- or South-aligned magnetic pixels. To accelerate programming times and support investigations into continuously programmed pixel values, we upgraded the prototype add-on to use an electromagnet (Fig. \ref{fig:Mixelplotter}). Our final add-on system consists of an Arduino Nano microcontroller, an electromagnet, an H-bridge, and a hall effect sensor encased in a 3D-printed housing, costing only \$62 in parts. We use our add-on to program commercially available and inexpensive soft magnetic sheet (X-bet, 26 mil thickness) that costs \$0.008/cm$^2$. A script translates our generated matrices into G-code, allowing the plotter to program faces without manual intervention. Once plotted, encodings can be viewed using magnetic viewing film (Fig. \ref{fig:plotter}, top right).

\vspace{5pt}
\noindent\textbf{Cubes and Magnetic Sheet Material:} We use a soft magnetic material to bond to our cubes and program. Soft magnetic material is material that can be easily coerced to become magnetized when placed in an external magnetic field. When removed from the field, they nonetheless retain a significant fraction of their magnetization, letting them function as magnets. This process is repeatable, allowing modules with soft magnetic faces to be re-programmed with new encodings that encode different target structures. We 3D print 25mm cubes from PLA and bond square faces of 26-mil thick soft magnetic material to its 6 faces (Fig. \ref{fig:plotter}, bottom right). The cubes are printed with internal cubic cavities of side length 18mm to neutrally buoy them in tap water. Cubes are placed in the CNC platform and programmed with encodings that produce the desired target configuration once mated. Each module costs \$0.23 in materials (\$0.19 PLA, \$0.04 for 6 soft magnetic squares), and requires 12 minutes to program all 6 faces with our un-optimized G-code. Taken together, cubes programmed with these encodings are therefore inexpensive, easy to manufacture, physically re-programmable, attractive over distances in contrast to contact adhesives (such as glue), and generate forces selectively without consuming power during operation.

\vspace{5pt}
\noindent\textbf{Electromagnet:} We chose an electromagnet over a permanent magnet because we can change the polarity of an electromagnet digitally by changing the direction of current applied to it. In addition, electromagnets allow us to continuously vary the magnetic strength of each pixel by regulating the magnitude of current through them~\cite{yasu2020magnelayer,yasu2017magnetic}. However, cylindrical electromagnets of the same diameter as permanent magnets exhibit less magnetic strength and thus, the resulting magnetic pixels are weaker. To create magnetic pixels of the same strength without trading-off resolution, we therefore shaped the electromagnet into a cone that narrows where it touches the magnetic sheet material. Since the narrow tip concentrates the magnetic flux, our shaped electromagnet creates a stronger magnetic pixel than a permanent magnet for the same pixel size.

\vspace{5pt}
\noindent\textbf{Writing Magnetic Pixels:} To create magnetic pixels, we use an electromagnet that is comprised of a cylindrical permalloy core (10mm diameter, 20mm length) wrapped with 250 turns of 20 AWG wire, with the last 5mm of one end filed to a cone whose tip writes 3mm wide pixels. The core has a relative permeability of $\mu_r=90000$, a factor of 40 greater than most brittle ferrite cores typically used for electromagnets. The electromagnet is coupled to the hardware add-on via a spring-loaded pogo pin, giving it compliance as the plotter touches it to the sheet surface. To drive the electromagnet bidirectionally, we connect it to a full H bridge we built using four relays (Omron 1718H1) driven by 2 MOSFETs (IRFZ44N) and shunted with flyback diodes (1N4752A). Given these features, our method allows us to program magnetic pixels in both polarizations using a single magnet without user intervention. Each pixel requires 0.7 seconds to program, drawing 130W from an offboard power supply to energize the electromagnet during this period and 200mW otherwise. We recorded no excessive heating of the electromagnet, and no observable wear on the CNC even after prolonged use (i.e., we programmed 1500 pixels consecutively to test the add-on's durability in operation). 

\vspace{5pt}
\noindent\textbf{Reading Magnetic Pixels:} To read magnetic pixels programmed onto the sheet, we use a hall effect sensor (MUZHI 49E) conditioned by a voltage comparator (LM393) on a breakout board. This allows us to read both the direction and magnitude of the magnetic polarities of individual pixels and to store these values digitally, which can be later used with our user interface to copy, edit and "paste" (program) pixels, even if the pixel value was previously unknown to the user. In contrast, passive magnetic viewing film can only detect the magnetic strength but cannot distinguish "North" from "South". In addition, since it is an analog method the results seen under viewing film cannot be easily transferred to digital tools.

\subsection{Control Software}

Our control software sends (1) movement commands to the CNC machine to move the hardware add-on over a specific magnetic pixel, and (2) commands to the electromagnet and hall effect sensor for writing and reading magnetic pixel values at the specific location. The CNC and the hardware add-on's microcontroller (Arduino) are both connected to a laptop via USB cables. We run a local server on the laptop to communicate with the CNC machine and the add-on. A python script on the server accesses the CNC's and hardware add-on's serial channels using the Pyserial library to both send commands and retrieve data via the serial port. 

\vspace{5pt}
\noindent{\textbf{Writing Magnetic Pixels}}: To plot a pattern, users run a python script with the 'plot' parameter and input a previously designed magnetic pixel matrix (.pkl file format). The .pkl file contains the magnetic pixel matrix as a 2D array with elements stored as 1 for 'North' pixels and -1 for 'South' pixels. To plot the matrix, the CNC moves to the first pixel and after the pixel has been programmed by the electromagnet moves translationally (in X or Y) one pixel width at a time. Between each pixel, the CNC rises and descends by 3 mm (in Z) to clear the sheet surface. Unchanged pixels are defined as 0 in the .pkl file and the CNC skips these to save plotting time. After the CNC moved to a specific pixel location, the python script sends a command to the microcontroller connected to the electromagnet to polarize the electromagnet in the correct direction (i.e., either 'North' or 'South') before turning it off. 

\vspace{5pt}
\noindent{\textbf{Reading Magnetic Pixels}}: To read a pattern, users run the python script with the 'scan' parameter and the size of the matrix as input. To read the matrix, the CNC moves to the first pixel and after the pixel has been read by the hall effect sensor moves translationally (in X or Y) one pixel width at a time. Similar to plotting, between each pixel, the CNC rises and descends by 3 mm (in Z) to clear the sheet surface. After the CNC moved to a specific pixel location, the python script sends a command to the microcontroller connected to the hall effect sensor to read the magnetic pixel value. The magnetic pixel readings are then saved as a .pkl file. The .pkl file can then be uploaded to the user interface, which then shows the magnetic polarity at each pixel.

\section{Results}

In this section, we evaluate how reliably the electromagnet can program the magnetic sheet, if the electromagnet can fully saturate the sheet to endow it with its greatest possible strength, how long the magnetic sheet remains magnetized when removed from the electromagnet, how accurately we can create pixels with continuous magnetic strength, and how accurately we can read magnetic pixel values. We also measure the magnetic force of the programmed encodings in attraction and repulsion, and compare this to predictions made by taking Hadamard products of the associated matrices. We finally evaluate the performance of the matrices in terms of the global and local agnosticism criteria.

\subsection{Reliability of Magnetic Programming}

We first evaluated how reliably the magnetic strength of individual pixels can be programmed, which is necessary to ensure consistent behavior.

\vspace{5pt}
\noindent{\textbf{Procedure:}} We collected data by energizing the electromagnet between 0A and 10A, in increments of 1A, in both North and South directions. For each applied current, we recorded the magnetic field at the conic tip of the electromagnet using a Gaussmeter (AlphaLab GM-2). We repeated the measurements 4 times, and computed the mean and standard deviation at each increment.

\begin{figure}[ht]
  \centering
  \includegraphics[width=0.77\columnwidth]{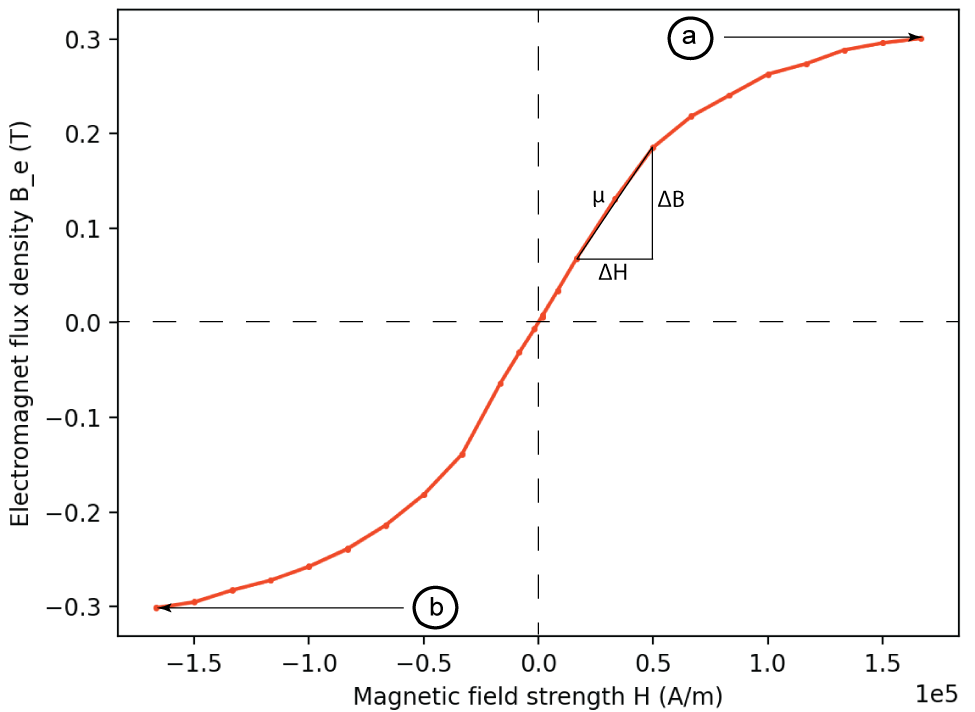}
  \caption{Electromagnet B-H curve. The electromagnet saturates at 0.34T at both (a) positive ('North') and (b) negative ('South') ends.}
  \label{fig:BH-emag}
\end{figure}

\vspace{5pt}
\noindent{\textbf{Results:}} Figure~\ref{fig:BH-emag} shows the resulting magnetization curve, also known as the B-H curve, for the electromagnet. The electromagnet saturates at 0.302T and the measurements exhibit an average standard deviation of 1.01mT, yielding a highly reliable field that can be generated within 0.3\%. The curve is symmetric and exhibits no hysteresis, allowing programming both North and South polarities reliably. The electromagnet can be turned completely off by removing power from the coil; this is illustrated by the curve intersecting the origin, showing that the induced B field collapses when the H field is set to 0, signifying very low coercivity and remanence.

\subsection{Maximum Magnetic Field Strength}

Next, we evaluated if the electromagnet is strong enough to fully saturate the magnetic sheet, which allows creating applications that leverage the material's maximum magnetic strength (flux density).

\begin{figure}[ht]
  \centering
  \includegraphics[width=0.77\columnwidth]{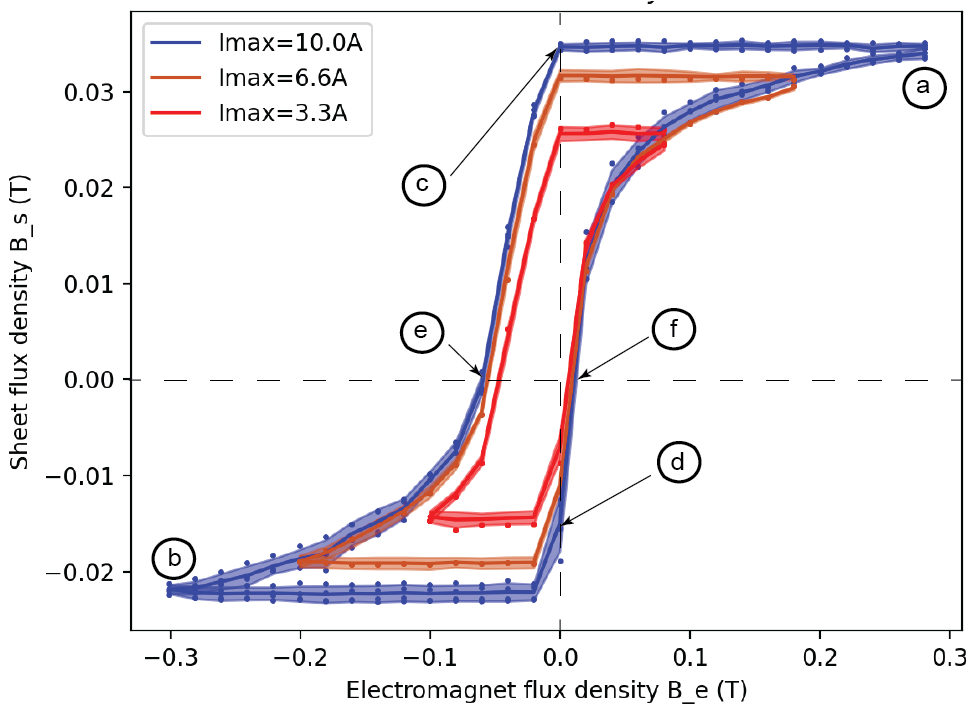}
  \caption{Sheet hysteresis curves when the sheet is fully saturated (major loop, blue) and when it is less than fully saturated (minor loops, red). Major loop is labeled to indicate (a) positive and (b) negative saturation; (c) positive and (d) negative remanence; (e) positive and (f) negative coercivity.}
  \label{fig:BH-sheet}
\end{figure}

\vspace{5pt}
\noindent{\textbf{Procedure:}} Data was generated by touching the conic end of the electromagnet to the sheet material, energizing the electromagnet, then de-energizing the electromagnet and measuring the field strength of the material sheet where it was programmed. We increased the strength of the electromagnet's magnetic field by increasing the current applied to it from 0A (Figure \ref{fig:BH-sheet}f) upwards in 0.6A increments, measuring the field strength of the sheet where it was programmed with each increment. We continued this procedure until an increase the electromagnet's field produced no additional magnetization of the sheet (Figure \ref{fig:BH-sheet}a).

\vspace{5pt}
\noindent{\textbf{Results:}} Figure \ref{fig:BH-sheet} shows the plotted data starting from \ref{fig:BH-sheet}(f) and ending at \ref{fig:BH-sheet}(a). The procedure showed that an external field generated using $I_{max} = 10A$ was the minimum current required to saturate the sheet. The curve shows that our electromagnet design is strong enough to saturate the sheet (at 0.0344T), generating as much force as possible for applications. In contrast, an equivalently sized pixel programmed by a cylindrical permanent magnet (neodymium, 3mm diameter, 6mm length) was 0.032T.

\subsection{Permanence of Magnetic Field}

We also evaluated the ability of pixels programmed in the magnetic sheet to stay magnetized after programming; their remanence, and the ease with which they can be de-magnetized; their coercivity. 

\vspace{5pt}
\noindent{\textbf{Procedure:}} We begun by following the procedure used to evaluate saturation, detailed above, using $I_{max} = 10A$ to saturate the sheet (Figure \ref{fig:BH-sheet}a). We then de-energized the electromagnet, reducing the current in increments of 0.6A, until the applied current was 0A, and measured the field strength of the material. The magnetic field strength of the sheet with the electromagnet turned off is shown in Figure \ref{fig:BH-sheet}(c), and is the material's remanence. We then continued by applying current to the electromagnet in the opposite direction; a negative current in Figure \ref{fig:BH-sheet}, increasing the magnitude of this negative current in increments of 0.6A until the field strength of the sheet was 0T. This is indicated in Figure \ref{fig:BH-sheet}(e), and is the material's coercivity. The same procedure used to generate the curve \ref{fig:BH-sheet}(f,a,c,e), is used to plot the remaining curve \ref{fig:BH-sheet}(e,b,d,f) by continuing to polarize the electromagnet in the opposite direction starting from \ref{fig:BH-sheet}(e). We repeated this 4 times and computed the mean and standard deviation at each increment.

\vspace{5pt}
\noindent{\textbf{Results:}} Unlike for the electromagnet, the magnetization curve for the magnetic sheet generated in Figure \ref{fig:BH-sheet} reveals a phenomenon known as magnetic hysteresis, which is the dependence of the current magnetization of the sheet on its magnetic history. The hysteresis generated by saturating the material, shown in blue, is known as the major loop. The asymmetry in this hysteresis is what enables the sheet to stay magnetized after programming. The hysteresis curve indicates a low coercivity, indicating that the material is easily re-programmed with a weak external magnetic field, and exhibits high remanence, illustrated by the negligible attenuation of the material's magnetic field strength after the electromagnet is turned off following saturation. The average standard deviation for the major loop was 6.81mT, and the standard deviation at each point is illustrated in the figure to indicate its repeatability.

\subsection{Continuous Magnetic Strength}

We evaluated if we can program pixels with continuous magnetic strengths, by reliably programming the sheet in a way that does not fully saturate it. This results is so-called 'minor' hysteresis loops, as shown by red and orange curves in Figure \ref{fig:BH-sheet}.

\vspace{5pt}
\noindent{\textbf{Procedure:}}  Data used to evaluate the minor curves were gathered using the same strict order as for the major loop, following the loop anti-clockwise. As mentioned previously, an external field generated using $I_{max} = 10A$ fully saturates the sheet. To investigate the effect of not fully saturating the sheet, we therefore chose fields created with lower current using $I_{max} = 3.3A,6.6A$. We then repeated the procedure outlined in section 7.3, beginning at Figure \ref{fig:BH-sheet}(f) and following each loop anti-clockwise back to its starting point. We repeated this 4 times for each $I_{max}$, and computed the mean and standard deviation at each increment. 

\vspace{5pt}
\noindent{\textbf{Results:}} Figure \ref{fig:BH-sheet} shows the two magnetization curves when the field strengths were generated with currents through the electromagnet of 3.3A and 6.6A. The average standard deviation for 3.3A and 6.6A curves are 5.85mT and 6.69mT respectively, and the standard deviation at each point is illustrated in the figure. This shows that the hysteresis loops can be utilized for programming a range of magnetization strengths.

\begin{figure}[ht]
  \centering
  \includegraphics[width=0.82\columnwidth]{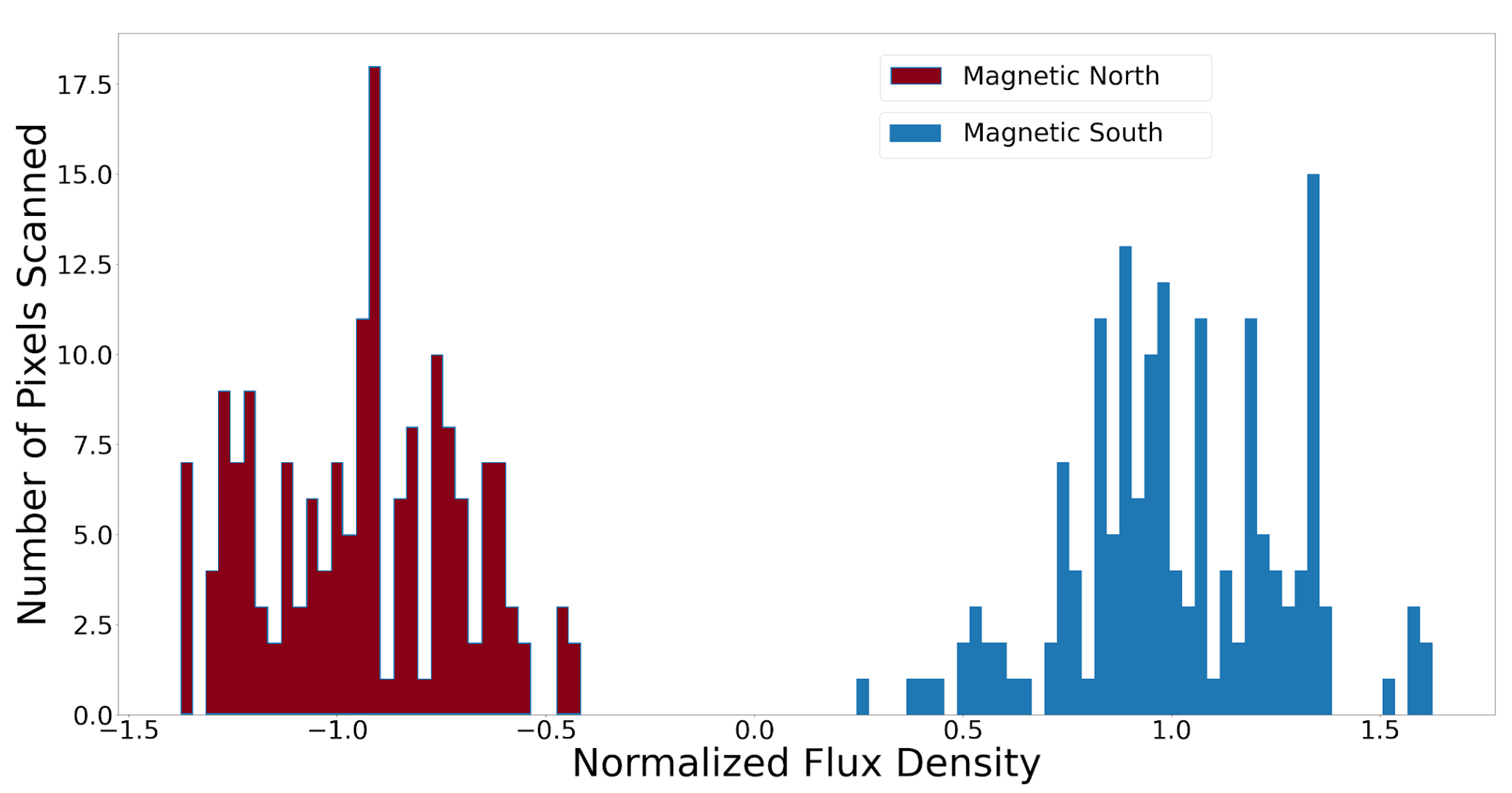}
  \caption{Histogram of magnetic pixel values scanned by hall effect sensor. North- and South-oriented pixels are clearly differentiated.}
  \label{fig:histo}
\end{figure}

\subsection{Accuracy of Reading Magnetic Pixel Values}

We evaluated how accurately we can read magnetic pixel values with our hall effect sensor.

\vspace{5pt}
\noindent{\textbf{Procedure:}} We programmed 150 North-oriented (with normalized flux of -1) and 150 South-oriented pixels (with normalized flux +1), and recorded the magnetic strength both with the hall effect sensor and with the Gaussmeter to obtain a ground-truth estimate.

\vspace{5pt}
\noindent{\textbf{Results:}} The recorded magnetic pixel strengths as measured by the hall effect sensor are shown in normalized form as a histogram in Figure~\ref{fig:histo}. North-oriented pixels are shown in red and South-oriented pixels are shown in blue. As can be seen in the figure, the read values exhibit significant noise with recorded values distributed widely around their ground truths of -1 and +1. Ground-truth measurements taken with a Gaussmeter showed that the programmed values were accurate, thus the noise is introduced by the inaccuracy of the hall effect sensor. However, since both the south and north regions have non-overlapping distributions, we can still differentiate between magnetically North- and South-oriented pixels. A higher quality hall effect sensor would be required to accurately measure pixels that were programmed with continuously variable magnetic strengths.

\subsection{Predicted vs Measured Interaction}

Finally, we evaluated how accurately we can predict magnetic interactions in terms of attraction, repulsion and agnosticism.

\begin{figure}[ht]
  \centering
  \includegraphics[width=0.99\columnwidth]{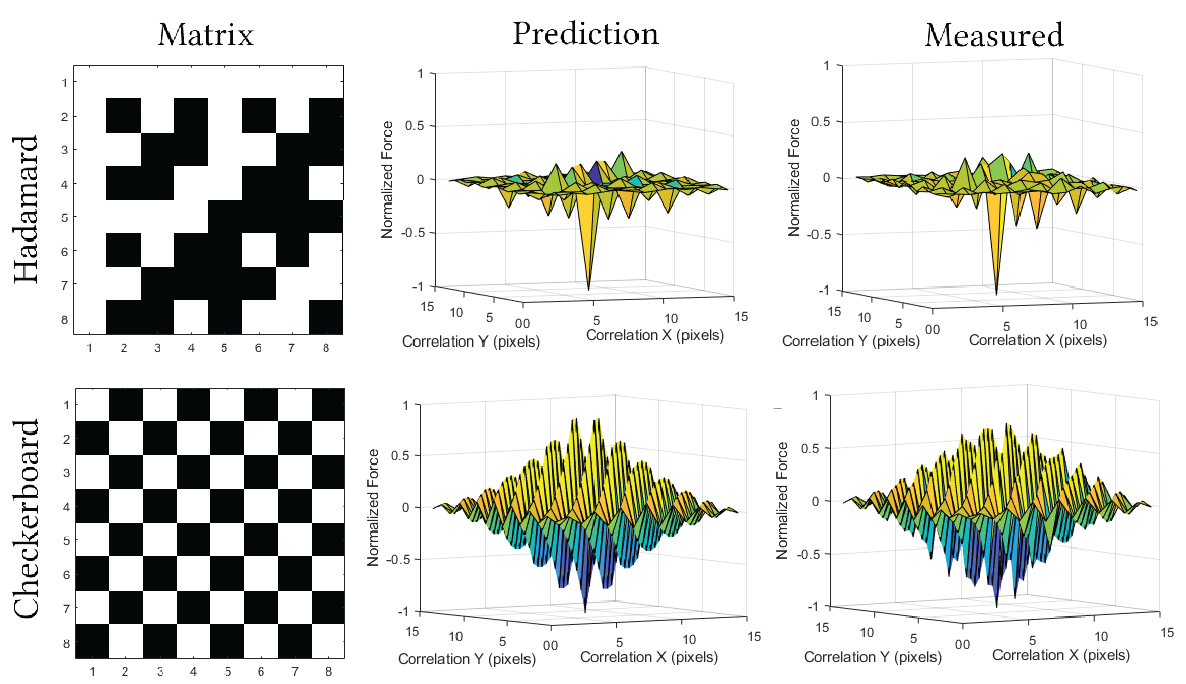}
  \caption{Predicted and measured magnetic interactions when two patterns, a checkerboard and a Hadamard matrix, are cross-correlated with their negatives.}
  \label{fig:theory-predict}
\end{figure}

\vspace{5pt}
\noindent{\textbf{Procedure:}} We created two magnetic pixel designs in our user interface (Figure~\ref{fig:theory-predict}a): (1)~an 8x8 checkerboard pattern of 'North' and 'South' programmed pixels, and (2)~an order-8 Hadamard matrix. We also created their complement matrices: the corresponding magnetic pixel pattern produced by multiplying each matrix by -1. We then translated each matrix pixel-by-pixel, in both X and Y directions, across its complement and evaluated the resulting force at each increment. 

\vspace{5pt}
\noindent{\textbf{Simulation Results:}} Figure~\ref{fig:theory-predict}(b) shows the predicted interaction between each matrix. We generated the predicted values by computing the normalized cross-correlation between each matrix and its complement. This effectively implies taking the sum of all attractive (-1) and repulsive (+1) pixel interactions, and dividing by the number of pixels. As a result, cross-correlation values of -1 designate perfect attraction, +1 is perfect repulsion and 0 is agnosticism. When the checkerboard matrix and Hadamard matrix are centered on their respective complements, they are by definition attractive at every pixel, yielding a value of -1. Elsewhere, the checkerboard produces oscillating attractive and repulsive interactions with every pixel-wise translation, whereas the Hadamard remains perfectly agnostic in pure X-translation and Y-translation, and maximally agnostic for mixed translation. 

\begin{figure}[ht]
  \centering
  \includegraphics[width=0.72\columnwidth]{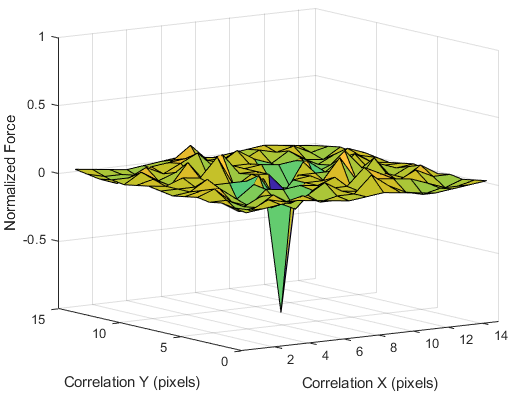}
  \caption{Translational agnosticism of a matrix A with its mate A'. They exhibit maximal attraction (-1) when centered, remaining largely agnostic (0) elsewhere.}
  \label{fig:correlation-local-trans}
\end{figure}

\vspace{5pt}
\noindent{\textbf{Physical Results:}} Figure~\ref{fig:theory-predict}(c) shows measured data. We measured this data by first programming the patterns on magnetic sheets (25mm side square). We then affixed one magnetic sheet onto a scale (KUBEI pocket, 0.1mN accuracy) placed on the CNC baseplate. The other sheet was mounted onto the CNC arm, which translated the patterns pixel-wise one pixel at a time (while keeping the magnetic sheet in a planar orientation and 0.5mm apart from the mounted sheet). We then recorded the force at each location using the scale. As can be seen in Figure~\ref{fig:theory-predict}, the measured data corresponds well visually with the simulated cross correlation, showing we can predict magnetic interactions between arbitrarily programmed magnetic sheets accurately before physically programming them. However, the measured repulsive values are weaker than those predicted by correlating the matrix values. We calibrated the scale to rule out ascribing this result to anisotropic measurement sensitivity. Rather, this is likely an effect of coercivity of the pixels on each other; if two attractive magnetic dipoles are brought into contact, they reinforce their attractive alignments. However, repulsive dipoles will realign to an attractive equilibrium if free to rotate; a condition which the low coercivity of the soft magnetic faces may support. Accounting for this, we implement a scaling factor of 0.09 to the repulsive forces predicted by correlation that normalizes the magnitude of the repulsive pixels to those in attraction. We compute a normalized sum of squared differences of 0.014 between the measured and predicted results using this scaling factor, supporting our model as an accurate predictor of force between magnetically programmed faces. The two square faces of 25mm side length were measured to exhibit an attractive force of 1.09N, corresponding to 1.74 kPa. In shear, the faces could withstand 1.31N, a high value likely caused by the exceptionally high friction coefficient (1.15) of rubber.

\begin{figure}[ht]
  \centering
  \includegraphics[width=0.95\columnwidth]{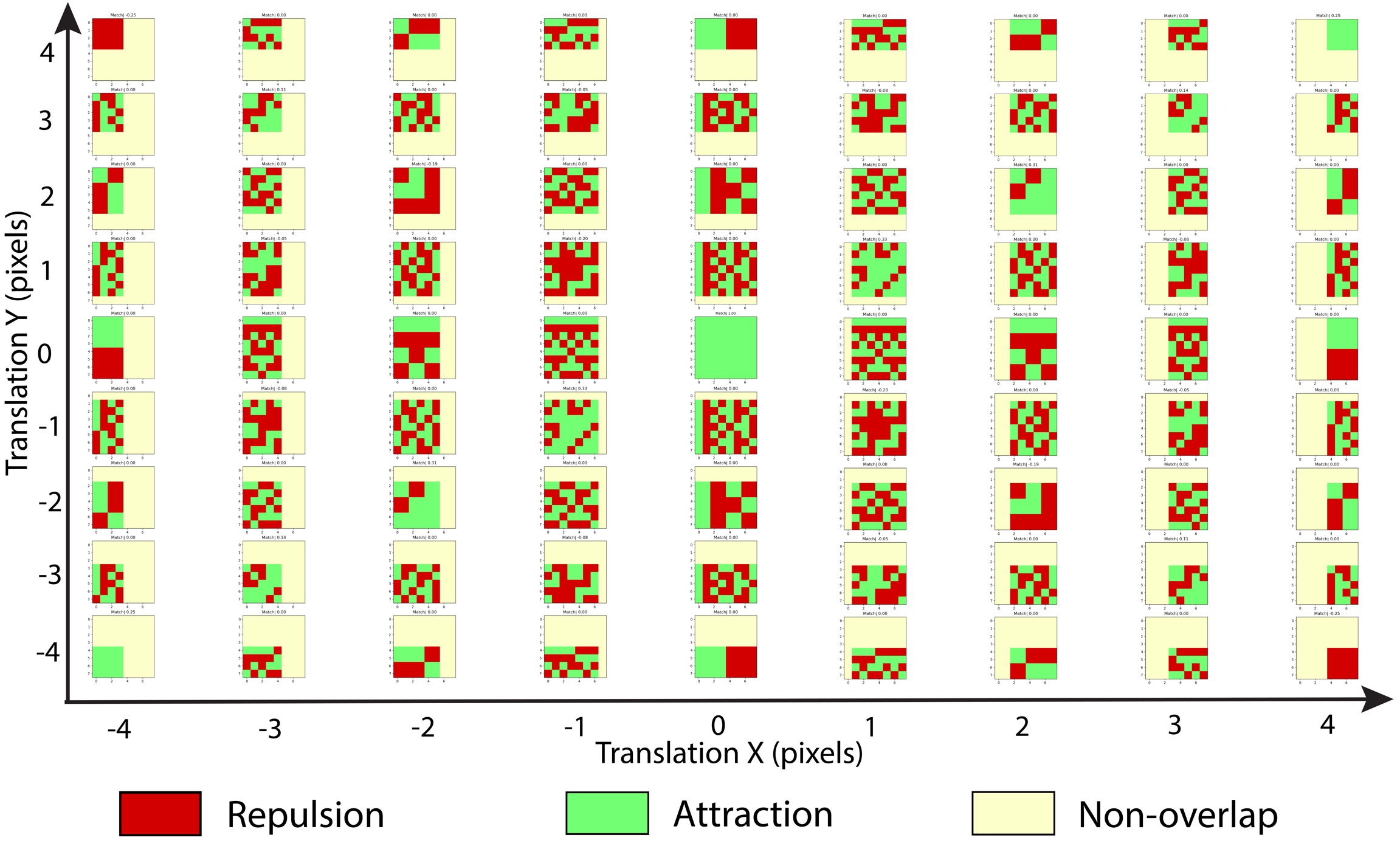}
   \caption{Pixel-wise attraction and repulsion of a normalized order-8 Hadamard matrix during translation with its mate. These are maximally attractive with the matrix pair translationally centered at (0,0), during full overlap. Elsewhere, attractive and repulsive forces largely cancel. 
   }
   \label{fig:translation-viz}
\end{figure}

\subsection{Local agnosticism criterion}
In this section, we use our model to evaluate the success of the local agnosticism criterion; the forces between our generated matrices and their mates. The figures are emblematic of the performance of all matrices in our clique.

To evaluate selectivity in translation, we correlate the matrices in X and Y, taking the Hadamard product at each pixel increment (Fig. \ref{fig:correlation-local-trans}). A peak normalized attractive force of -1 is produced with the matrices translationally centered at (0,0); this corresponds to 256 Pascals, or 160 mN between square faces of side 25mm. Elsewhere, the correlation is dominantly agnostic (centered about 0) or repulsive (positive), bounded in attraction by -0.25. 

To illustrate how the Hadamard product produces a dominantly agnostic interaction between the matrices besides their mating configuration, Fig. \ref{fig:translation-viz} visualizes the pixel-wise attraction and repulsion during the correlation of a normalized Hadamard with its inverse. Here, red pixels indicate repulsion (+1), green attraction (-1), and yellow non-overlap (0). Summing the pixels over a square produces the Hadamard product, or the net force, that is plotted in each data point in Fig. \ref{fig:correlation-local-trans}. The translationally centered position (0,0) in Fig. \ref{fig:translation-viz} produces uniform attraction, whereas other positions produce an exactly or largely agnostic interaction due to equal numbers of attractive or repulsive pixels cancelling out.

% \subsubsection{Agnosticism via Translation}
\begin{equation}
\begin{bmatrix}
x'\\y'
\end{bmatrix} = 
\begin{bmatrix}
cos(\theta) & -sin(\theta)\\ sin(\theta) & cos(\theta)
\end{bmatrix}
\begin{bmatrix}
x\\y
\end{bmatrix}
%  \leq f^2 + 6\sum_{n=2}^{N-1} nf - 2(n-1)
\label{eq_rot}
\end{equation}

\begin{figure}[ht]
  \centering
  \includegraphics[width=0.77\columnwidth]{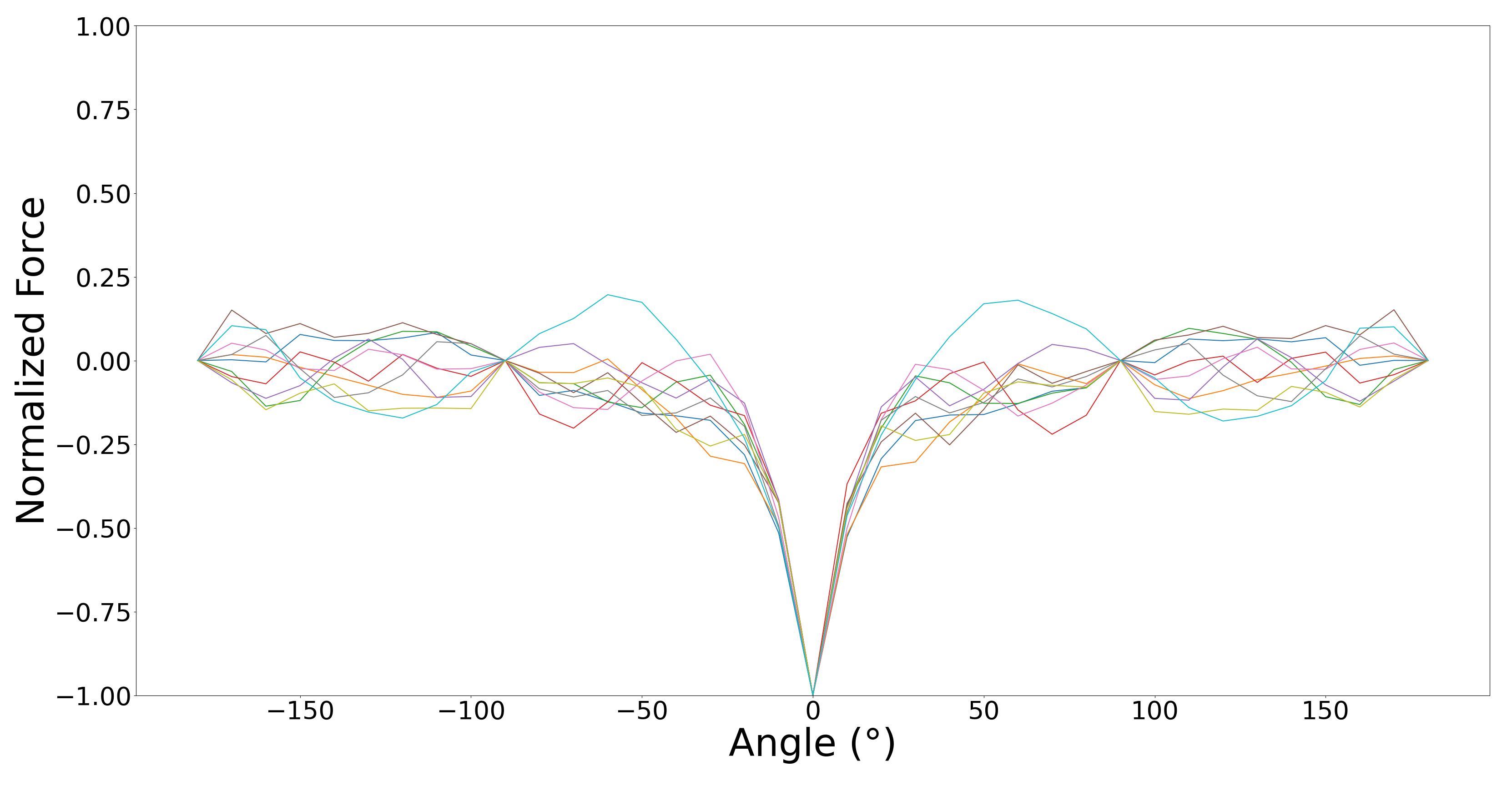}
  \caption{Rotational agnosticism of a matrix A with its mate A'. They exhibit maximal attraction (-1) when centered, remaining largely agnostic (0) elsewhere.}
  \label{fig:rotation-local}
\end{figure}

\begin{figure}[t]
  \centering
  \includegraphics[width=0.65\columnwidth]{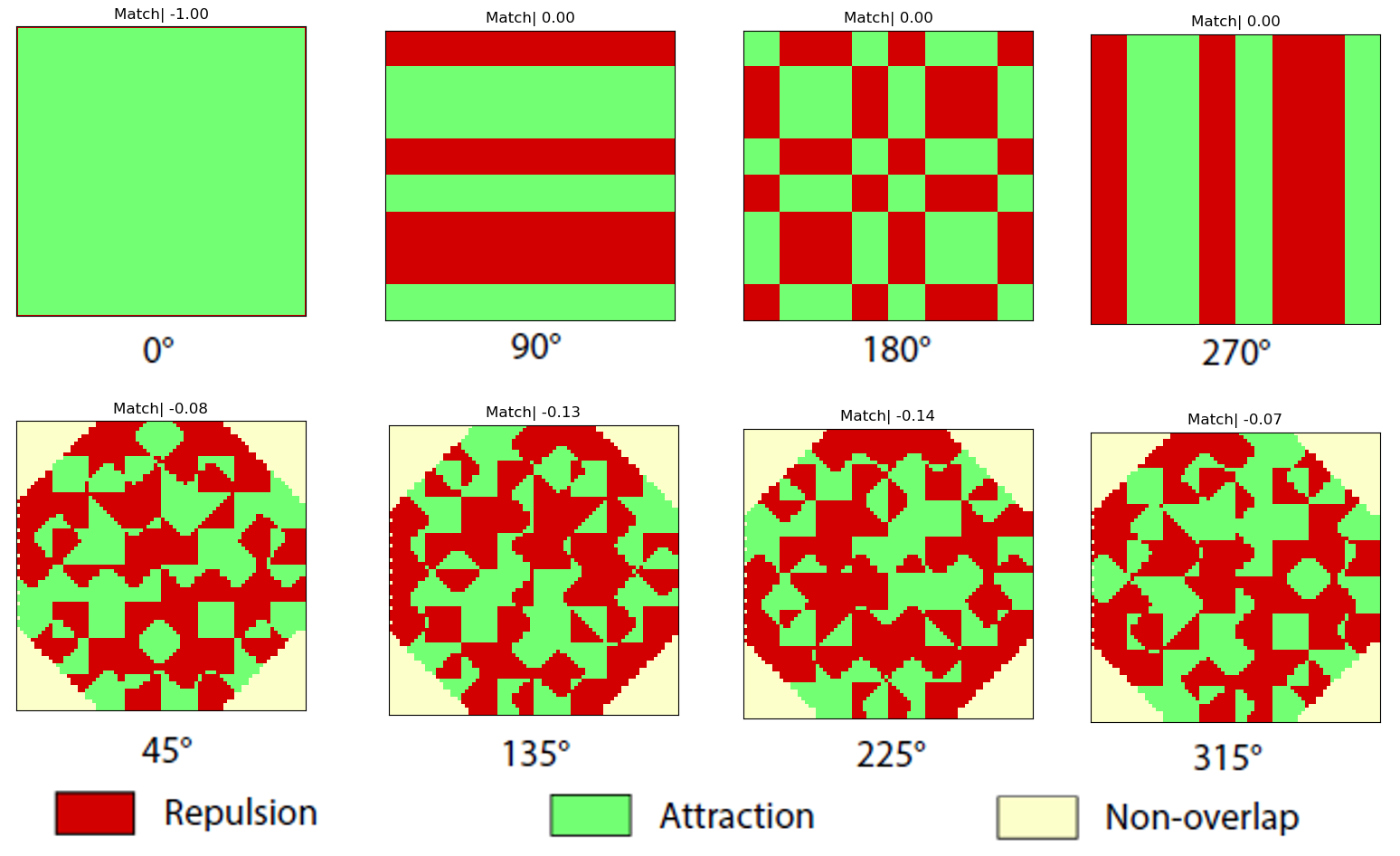}
   \caption{Pixel-wise attraction and repulsion of a normalized order-8 Hadamard matrix during rotation with its mate. For rotations of 90n$^{\circ}$, attraction and repulsion cancel exactly; elsewhere, approximately. }
   \label{fig:rotation-viz}
\end{figure}

To evaluate selectivity in rotation, Hadamard products can be evaluated between the matrices for orientations of 90$^{\circ}n$, where n is an integer. Elsewhere, the matrices do not superimpose. Therefore to evaluate these products at arbitrary angles, we first discretize the order-8 matrices by a factor of 10 to produce 80x80 grids. Each matrix element is then said to be at location (x,y) with respect to an origin placed at the center of the matrix, and we use the rotation matrix (\ref{eq_rot}) to compute its new location (x',y') after an arbitrary rotation of $\theta$. We smooth the result using a 3x3 averaging kernel to remove artefacts from the discretization, then evaluate the Hadamard product to evaluate the net force, assigning a value of 0 (non-overlap, or agnosticism) where pixels do not overlap. Fig. \ref{fig:rotation-local} plots this force for 10 matrices from -180$^{\circ}$ to 180$^{\circ}$ in 10$^{\circ}$ increments. While our search procedure optimized only for rotations of multiples $90^{\circ}$, this procedure verifies performance for arbitrary rotations. The attractive force is bounded by -0.25 in all orientations besides the intended mating configuration (0$^{\circ}$). Pure agnosticism is further enforced at 90$^{\circ}n$, where the square matrices line up.

Fig. \ref{fig:rotation-viz} visualizes the pixel-wise attraction and repulsion during the rotation of a normalized Hadamard with its inverse. At 0$^{\circ}$, 90$^{\circ}$, 180$^{\circ}$ and 270$^{\circ}$, locally attractive and repulsive pixels sum to 0, producing agnosticism. Elsewhere, attractive and repulsive pixel interactions cancel to within an attractive bound of -0.2.

\subsection{Global agnosticism criterion}

\begin{figure}[ht]
  \centering
  \includegraphics[width=0.72\columnwidth]{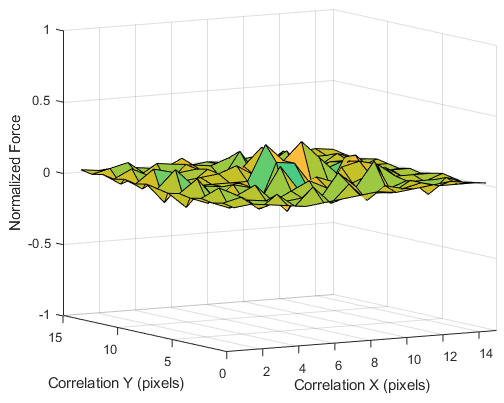}
  \caption{Translational agnosticism between non-mating matrices; agnosticism dominates for all configurations.}
%   \Description{matrix-inverse-correlation}
  \label{fig:correlation-global}
\end{figure}

\begin{figure}[ht]
  \centering
  \includegraphics[width=0.77\columnwidth]{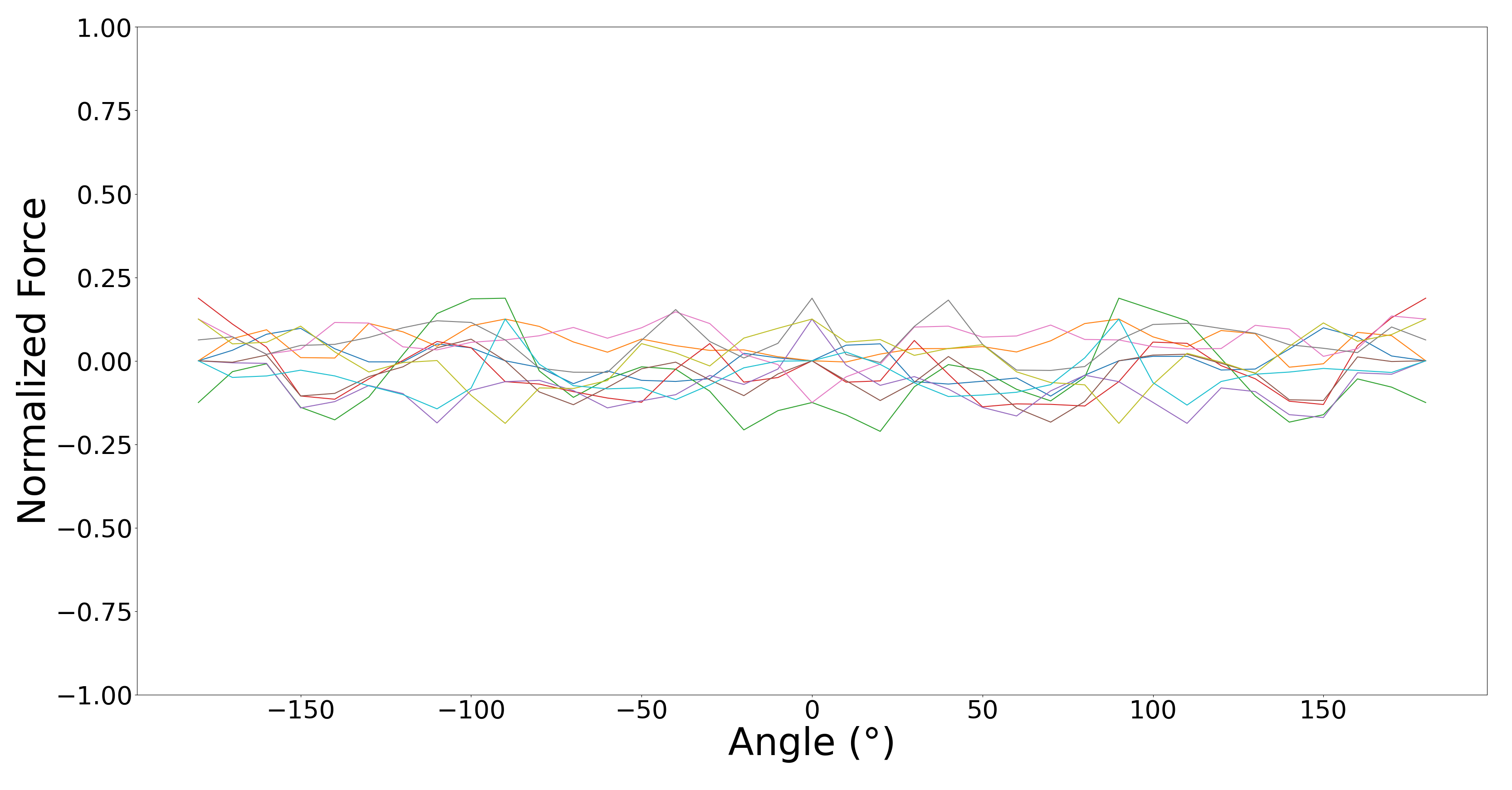}
  \caption{Rotational agnosticism between non-mating matrices; agnosticism dominates for all configurations.}
  \label{fig:rotation-global}
\end{figure}

In this section, we evaluate the global agnosticism criterion between two emblematic matrices from our clique. Using the same tools used in the local case above, Fig. \ref{fig:correlation-global} shows the correlation between two matrices, illustrating their agnosticism over all translations. Equivalently, Fig. \ref{fig:rotation-global} illustrates their agnosticism in rotation, with a negative bound of -0.36 that indicates that attraction between non-mating faces is never greater than 36\% of the attractive force between mating faces in alignment, as derived in our search.      

\subsection{Stochastic self-assembly}

\begin{figure}[ht]
  \centering
  \includegraphics[width=0.95\columnwidth]{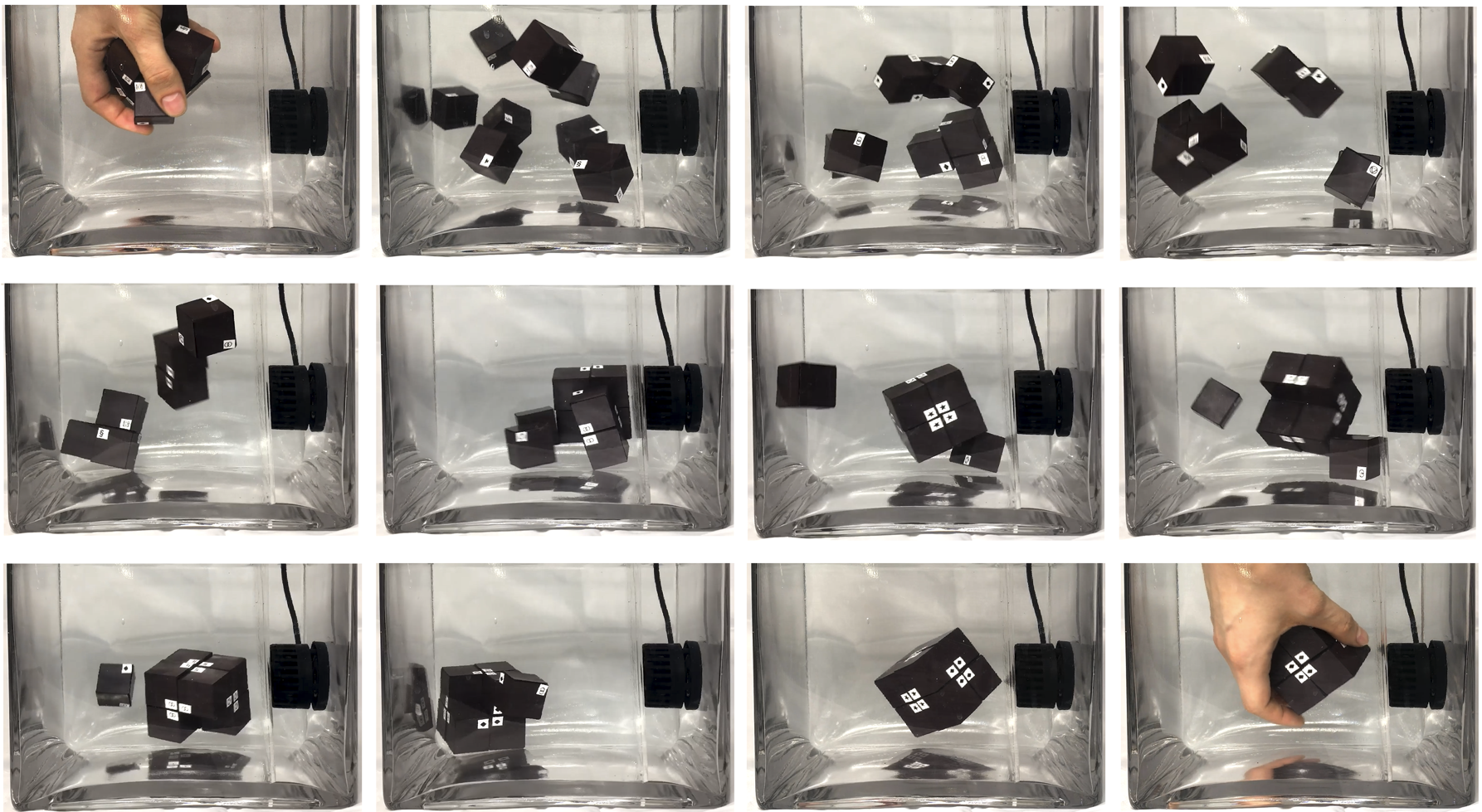}
  \caption{Stochastic self-assembly from (top left) a random arrangement of 8 cubes to a (bottom right) meta cube.}
  \label{fig:assembly}
\end{figure}

We built and programmed 8 cubes to stochastically self-assemble into a "meta cube" (Fig. \ref{fig:assembly}), to suggest how cubes could be programmed to assemble structures into recursively larger cubes. In this configuration, each cube occupies a vertex in an octree and connects to 3 other cubes, yielding an assembly that requires 12 pairs of mating encodings. To do this, we selected one of the 4 maximal cliques (size 12) of mutually compatible encodings (Fig. \ref{fig:teaser} above, right). We wrote a script to translate these encodings into G-code and deployed this on our magnetic plotter, programming each module face in 2 minutes per face. We released all cubes into a glass container (cubic, 200mm side length) filled with tap water, that was stochastically perturbed by a hydraulic pump (Hygger WaveMaker 1600gph). The pump was programmed to produce stochastic flows of random magnitude and frequency to stimulate brownian motion of the cubes. We inserted a laser-cut mesh between the cubes and the pump to promote turbulent flow and to prevent cubes from becoming drawn into the pump inlet. We experimentally calibrated the force of our stochastic disturbance to exceed the attractive force (-0.36) of misassemblies until no permanent misassemblies were observed. Following this procedure, the cubes acquired their correct positions to self-assemble the structure in 32 hours (see supplemental video). After assembly, we re-programmed all faces with new encodings to acquire different final target shapes and measured individual mate forces, observing no difference in the strength of individual mates after reprogramming.

\section{Programming material across media and scales: DNA}

Through code-theoretic programming, we have seen that it is possible to exploit magnetic materials themselves to align and assemble parts without active control. However, there exist many materials that exhibit the philic and phobic properties that can be exploited for self-assembly. These can help make tractable the challenge of fabrication, not just for machines and modular platforms, but to enable assembly at the micro and nano scales where top-down manipulation is most challenging. One such material is DNA, and its application area known as DNA self-assembly or DNA origami. In this section, we introduce initial results showing how our code-theoretic programming approach for self-assembling magnetic materials at the cm-scale can be applied to assemble biological materials in the form of DNA at the nano-scale.  

\subsection{Motivation}
Biologists have worked on programming DNA to fold into structures to create biological storage devices and molecular machines, such as vaccines. However, these methods have typically suffered from relatively small yield rates due to the non-selectivity in their programmed codes, and the non-existence of any proxy technologies at the cm-scale that can be used to test self-assembly easily.

Self-assembled DNA nanostructures enable nanometer-precise patterning that can be used to create programmable molecular machines and arrays of functional materials. DNA origami is particularly versatile in this context because each DNA strand in the origami nanostructure occupies a unique position and can serve as a uniquely addressable pixel. However, the scale of such structures has been limited to approximately 0.05 square micrometers, hindering applications that demand a larger layout and integration with more conventional patterning methods. Wireframe DNA origami has emerged as a powerful and versatile approach for fabricating arbitrary 2D and 3D nanoscale structures using top-down computer-aided design. However, limitations in overall object size imposed by DNA origami scaffold length (typically 7 kb from the M13 phagemid) requires multi-origami objects to be self-assembled into finite and extended assemblies and arrays to realize larger-scale material fabrication. Extending the design space of DNA origami via self-assembly into programmable superstructures is a major goal for constructing well-defined 2D and 3D nanomaterials with structural control spanning from the nanometer to micrometer and larger scales. 

\begin{figure}[ht]
  \centering
  \includegraphics[width=0.95\columnwidth]{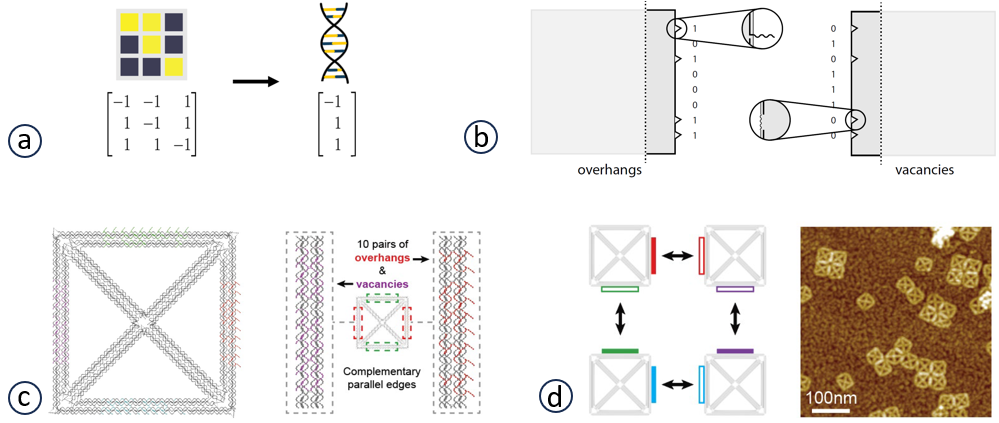}
  \caption{Overview of translating our code-theoretic arrays for use in DNA assembly. (a) we translate our 2D arrays, instantiated as binary-valued magnetic poles, into 1D strings, using a quaternary alphabet of A,T,G and C. (b) We design square DNA tiles with 10 features along their edges, where each feature is either a DNA overhang or a vacancy. (c) We synthesize real DNA and and instantiate our 1D codes as a series of vacancies and overhangs that are designed to make a particular overhang sequence bind to its complementary vacancy. (d) We designed our square tiles to assemble into hierarchically larger squares (left), and Atomic Force Microscopy illustrates the formation of assemblies (right).}
  \label{fig:dna}
\end{figure}

\subsection{Approach}
Given the interest in developing programmable superstructures using DNA origami techniques, we investigated porting our code-theoretic array design from magnetic materials into the DNA domain. A key insight here is that DNA exhibits a behavior that is some ways analogous to how magnets interact. In double-stranded DNA, the molecular double-helix is comprised of two backbones that are held together by four types of nucleotides, or bases. These bases are adenine (A), thymine (T), cytosine (C), and Guanine (G). Importantly, the bases connecting the backbones are formed in pairs, with adenine forming a base pair with thymine, and cytosine forming a base pair with guanine; other base pair permutations do not readily arise. This echos the behavior of how magnetic poles interact, where North-aligned magnetic poles will bond to South-aligned magnetic poles, but not with other North-aligned poles. 

Identifying that orthogonal codes can be used across both of these regimes, we used our 2D arrays, instantiated as binary-valued magnetic poles, and translated these into 1D strings, using a quaternary alphabet of A,T,G and C. We synthesized real DNA, using DNA origami techniques to fold DNA into square tiles of 50nm side length, and programmed our 1D strings along the edges of these tiles. We programmed the codes in pairs such that 4 tiles would be elicited to bond into a square superstructure, and imaged the results using Atomic Force Microscopy.

\subsection{Tile design}
DNA nanotechnology theoretically allows the synthesis of large assemblies with nanoscale pattern control. Researchers typically use polygonal DNA "tiles" called DNA origami, and the edges of two tiles may bind using DNA overhangs that extend outwards and bind to dedicated sites on another tile edge. The sequence and position of these overhangs is varied to prevent interactions between incorrect pairs of edges. We designed square DNA tiles, and each edge presented either up to 10 overhangs (short DNA strands extending out of an edge) or up to 10 vacancies (short single-stranded DNA regions within an edge). Each overhang sequence was designed to bind into a particular vacancy on another edge. Given a binary code of 1s and 0s (our 1D string), an edge with overhangs presents an overhang for each 1, while an edge with vacancies presents a vacancy for each 0.

\subsection{Tile fabrication}
Each DNA tile was formed from a long DNA strand (the scaffold) and several short DNA strands (the staples) that bind to the scaffold to fold it into the desired shape. We used tiles with one edge presenting 10 vacancies and another edge presenting the 10 corresponding overhangs. Scaffold and staples were mixed in a buffer containing 12.5mM MgCl2 with the staples at 15-fold excess concentration compared to the scaffold, and then cooled (annealed) in a PCR thermocycler from 95C to 25C over 22 hours. The annealed solution was washed with a 1 x TAE 12.5 mM Mg(OAc)2 buffer and the extra staple strands removed with MWCO = 100 kDa spin filter columns. The purified DNA origami solution was adjusted to desired concentrations for superstructure self-assembly. 

To initiate finite origami superstructure self-assembly, origamis were mixed with a 1:1 stoichiometry at a concentration of 15-50 nM each. The mixed samples were incubated at 37C for a day, and then thermally annealed from 37C to 20C at a rate of 3C per hour. For agarose gel electrophoresis, the self-assembled origami superstructures were validated by 1\% Agarose gel in 1x TAE buffer with 12.5 mM Mg(OAc)2 and 1x SybrSafe. Gels were run at 60 V and subsequently imaged using Typhoon imager (FLA 7000). The final step was the Atomic Force Microscope (AFM) sample preparation and imaging, for which the assembled superstructures were deposited on to mica substrates for AFM imaging.

\subsection{Results for programmed DNA self-assembly}

Our initial AFM images reveal the formation of planned tile assemblies at the nanoscale using our codes. This suggests that the application of code-theoretic programming for self-assembly may span both magnetic and biological material domains, across 6 order of magnitude. However, the scope of this initial work was limited, and further work is required beyond these preliminary results to evaluate its efficacy. In the course of these initial experiments, assembly yields were not calculated, and misassemblies were present among correct assemblies; a result that may result from limited fine-tuning of our experimental process as much as from our method itself. Moving forward, immediate next steps are to evaluate these questions before attempting the creation of large, finite assemblies using the orthogonal binary codes to ensure that only correct pairs of edges bind to each other. In principle, this should allow researchers to design reliable systems with a large number of distinct edges, allowing for larger and/or more complex structures.

\section{Discussion}

In this chapter, we used programmable materials to introduce a method to build and program modules capable of selective self-assembly. These modules are scalable; they are passive, inexpensive (\$0.23/module) and consist of COTS bulk material. We have introduced a method of generating encodings that are provably selective. We demonstrate a method of generating highly selective cliques of mutually compatible face encodings for modules, and place guarantees on agnosticism for non-mating configurations in translation and rotation, verifying these results experimentally by instantiating encodings as magnetic polarities. We have shown that these modules can be programmed with encodings that result in specified self-assembled geometries using a simple magnetic plotter, and demonstrated that modules can be re-programmed for new target shapes. We demonstrated self-assembly using 8 modules in tap water. Finally, we highlighted preliminary results showing how code-theoretic assembly can be applied DNA, suggesting its generalizability across materials and scales. While our technique was successful at self-assembling conservatively sized systems, a number of limitations and avenues for future work present themselves.

The high selectivity of these encodings\textemdash agnosticism in translation and rotation\textemdash result in significant assembly times, as it constrains the influence of its near field force to areas directly above module faces. This diminishes the ability of programmed faces to easily attract their mates over distances, requiring them to enter each others' narrow volumes of magnetic influence before mutual attraction results in a bond. First, the water chamber size and flow rate could be better calibrated. Future work could investigate adapting the matrices to exhibit larger volumes of influence in the form of magnetic potential wells, with gradients of attraction that help direct mates to their correct orientation. In addition, these modules may be a promising candidate to be used in conjunction with semi-directed stochastic assembly methods \cite{tolley2008dynamically,tolley2011programmable,tolley2010fluidic}, which manipulate the fluid to guide mating modules into their basins of attraction. This may also help address the current system's inability to control assembly order, which would be crucial for complex target geometries. Using COTS fridge magnet, the attractive pressure between our mating faces is relatively conservative (250 Pa). Future work could investigate the use of more strongly magnetizable materials to increase mating strength. These encodings could equivalently be used to build strong interfaces by replacing programmed pixels with arrangements of permanent magnets, and for active self-assembly using electromagnets~\cite{nisser2021programmable} or electropermanent magnets whose encodings could be changed online. In addition, the programming step may be rapidly accelerated using parallel arrays of electromagnets to program every pixel simultaneously. Our search procedure successfully founded matrices that permitted self-assembly for matrices of order N=8, however future work could improve our search to extend to larger cliques and matrix orders in shorter time. Finally, our initial experiments applying our code-theoretic programming technique to self-assemble DNA suggest the technique's promise for generalizing across materials and scales. While we have demonstrated the implementation of selectively mating encodings magnetically, and shown initial results applying these in DNA, future work could also investigate the encoding of these matrices in other binary media, such as in the patterning of electrostatic charge, wettability or chemical bonding, that may also permit self-assembly at smaller scales. 

\section{Summary}

In previous chapters, we introduced how to automate fabrication and assembly at the machine level and part level. In this chapter, we developed a method that complements these at the third level, by illustrating how to automate assembly using materials themselves. Together, these provide three hierarchically complementary ways in which to automate manufacturing and assembly in situ using machines, modules, and materials. In the following chapter, we will summarize the contributions of these works and introduce challenges and opportunities for future work. 

%% file: Learnings.tex
\chapter{Lessons Learned}
\label{sec:Learnings}

The goal of this chapter is to summarize some of the lessons learned during the development of this thesis. The intention is to share some of the pitfalls and challenges encountered during the research process that aren't typically covered in published works. These anecdotes are included with the hope that they may be useful for other researchers pursuing work on similar topics.  

\subsection{Start simple and iterate}

At the start of a research project, there is typically an idea of what the final artefact, software pipeline or hardware will look like, and how it improves on the state of the art. It is also tempting to start building an artefact with all the complexity that the final demonstration is thought to require. However, one of the most useful pieces of advice I have received is to start with a simple model, or a simple prototype, and then iterate from there. When building a robotic device, this typically means starting with a physical prototype with fewer degrees of freedom, where you have the opportunity to better understand the basic dynamics before increasing the scope of the project. Other times, the simple prototype might show you that the complexity you initially envisioned being required to solve your problem isn't required at all. In \textit{ElectroVoxel}, one of the initial ideas for the project was to build an inexpensive, modular robotic system that we could build in high numbers in order to develop and test a series of complex, model-based controllers for reconfiguring cube-based pivoting systems. However, two things happened that meant that this research direction was on the one hand not needed, and on the second hand not practical. To begin with, the very first controller we built, a bang-bang controller to drive the pivoting maneuver, worked remarkably well. The development of a more complex controller would have been difficult to justify for a modular system consisting of just a few modules. This justification ties into the second reason; a key motivation for this project was to develop a reconfigurable system that could function in microgravity, and during the development of the project, we received confirmation that our prototype was to be manifested on a parabolic flight in 3 months time. The duration of stable microgravity during a single flown parabola is on the order of a few seconds, and in addition to the space and logistical constraints of the flight, deploying more than 3 modules would have been inappropriate for maximizing the chance of a successful microgravity demonstration. The advice was to start simple and iterate, but the lesson here is twofold: first, sometimes simple is good enough, and even if it isn't, it will put you on track to understanding what is; second, research is often about adaptation. A project rarely leads exactly where you think it will, either because you discover something new along the way, or because of new challenges and opportunities arising along the way. Being able to respond quickly is always in your favor, and the new research direction is often more interesting simply because you didn't anticipate it\textemdash which means others weren't likely to think of it either. 

\subsection{Co-developing robot software and hardware}

Co-developing software and hardware like that developed throughout this work introduces a number of challenges, not least of which is that bugs are hard to weed out. One example of this is in \textit{ElectroVoxel}, during the initial testing of the untethered modules' ability to command its neighbours to pivot in a given direction. We designed a custom wireless communications protocol on the software side, and a new electronics architecture on the hardware side, and when a command failed, debugging could turn into a tedious, iterative search to identify the source of the problem. One lesson through this process was to spin out printed circuit boards as quickly as possible. During the prototyping phase, it is tempting to use breadboards and protoboards to facilitate rapid hardware iterations. However the multitude of problems that can arise, including both short circuits and open circuits due to connection problems, often slowed progress due to time required to identify faults.

Co-developing software and hardware can also introduce opportunities however, as one can be used to make up for shortfalls in the other. Above, we detailed how a simple bang bang controller worked where a more complex controller might have been expected to be required. The reason for this is in part because the hardware was developed first in a way to ease programming. For example, the electromagnetic actuators were sized large enough to avoid saturation effects, and effective thermal dissipation was used via heat sinks which meant individual electromagnets could be activated for longer time periods, eliminating the need to draw on auxiliary electromagnets to complete pivoting maneuvers as part of the control strategy.

\subsection{A Pareto Principle for systems research}

Over a century ago, an economist called Vilfredo Pareto showed how around 80\% of the land in Italy was owned by 20\% of the population. The Pareto Principle, or the 80/20 rule, is just one example of a power law that has wide-ranging applications as models that capture asymmetric cause and effect. It's also perhaps useful for capturing research in general. For systems research in particular, one of the lessons we learned is that oftentimes 20\% of the work gets you 80\% of the results. This may at first glance seem like a good thing, and maybe it is. But if you need 100\% of the results, the corollary is that the remainder of the research project will entail working four times as hard for a quarter of the payoff. The issue is that if you generate 80\% of your required results in 2 months, it can be difficult to anticipate the fact that it will take an additional 8 months to finish.

The \textit{LaserFactory} project was a significant case in point. The goal was to create a hardware system capable of creating an artefact's geometry, depositing its circuit traces, and assembling its components, all the while facilitating custom designs for these artefacts in a software tool. Building up the hardware add-on, the associated software tool, and the communication interface had proceeded with rapid progress. The final piece, adding pick-and-place functionality, required significantly more time. First, we developed the physical pick-and-place mechanism in addition to the associated software. This included a drag-and-drop interface, a component library, and auxiliary tasks that included calibrating center of masses and pin locations for each component, as well as automating collision-free trajectories for part placement. While the isolated development of this system proceeded relatively smoothly, it was primarily how this part interfaced with the whole that caused challenges. Embedding the mechanism into the add-on's chassis resulted in significantly more load to the laser cutter gantry, prompting a redesign of all hardware in order to redistribute the weight. As we explored the design space and used the mechanism to collect increasingly heavy components, the greater add-on mass combined with the payload masses meant further design iterations were required to bolster the structure to prevent sagging. Taller components yielded yet another reason for redesign, both in order to provide enough for the hardware add-on to travel over these components, but also because components could drag across the acrylic substrate, smearing the uncured silver traces that had been deposited. While these are just a few examples of how one new feature can cause design problems through its interaction with all existing features in a system, there is also another kind of problem. With the added complexity of placing dozens of components, there were now vastly more steps in the fabrication process that all had to work perfectly, in one go. While this kind of robustness is expected in commercial products, this can require considerable resources, and the decision to build this into a research prototype must be weighed against its academic value and to what extent the artefact will be used once the development phase culminates.

\subsection{Parabolic flight deployments}

In the course of this work, we deployed 4 separate research projects in microgravity across 4 separate parabolic flight campaigns. Only one of these is included in this thesis, but much of the work surrounding deployment has been circumvented due to the thesis' focus on research outcomes. To address this, this subsection summarizes some of the work required to deploy a research payload on a parabolic flight.

The logistical work behind a flight is one aspect to consider. Preparing for and approving a payload for flight requires building a compartment with sufficient volume that can be bolted to the aircraft floor, and an electrical specification that is compatible with the onboard power sources. A safety review is required of all electronic, mechanical, free-floating, sharp or hazardous artefacts, as well as a mechanical evaluation of what loading conditions are expected across each bolt used to secure the payload; both nominally as well as if one bolt fails. A concept of operations must be evaluated, specifying every action taken during the course of the flight, during and between every parabola, which typically number between 20 and 30. Besides these items, there is another vital consideration. A key way in which a microgravity demonstration differs from a lab test is that the microgravity demonstration must be all but guaranteed to work on the first try. It is common in research prototypes that an electrical connection becomes loose, friction causes a mechanism to stall, or a nut loosens and causes an assembly to become misaligned. While issues like these are commonplace and can often be resolved in a few minutes in a lab, there is typically no way to do so on the aircraft. Trying to swap out a faulty battery the first time one is subject to microgravity or a prolonged 1.8g loading condition is a good recipe for disaster. When an experiment is secured in a compartment with minimal access, there is no soldering iron, and the experimenter is in free fall, the success of an experiment relies on both primary and contingency plans to be executed automatically. In our case, one of our experiments was expected to deplete one of its battery sources before the end of the flight. As a result, we built custom quick-swap battery systems in addition to pre-building replacement experiments that could be activated in the event that the primarily payload misfired or the battery swap failed; both contingencies were ultimately required. One incredibly useful piece of advice we received was to build a dongle with a series of buttons that could be pressed to command our experiment with pre-programmed functions and contingency protocols. This allowed us to avoid re-programming our artefacts during the flight or replacing components, instead executing functions with a simple button push. The key learning from these experiences was that every failure case must be anticipated ahead of time, and solutions to these must be prepared before the flight. 

%% file: Conclusion.tex
\chapter{Discussion and Future Work}
\label{sec:Discussion}

The ability to manufacture and assemble customized artefacts at points of need can significantly enhance our ability to meet the technical and personal needs of diverse populations and use cases, respond rapidly to unforeseen events, and address critical gaps in supply chain security. Towards that end, this thesis has introduced a set of consolidated manufacturing techniques capable of assembling physical artefacts in situ. These techniques were partitioned into three classes of strategy: (1) multi-process manufacturing machines, (2) modular assembly platforms, and (3) programmable materials. Together, these illustrated how automating fabrication in situ can be accomplished at three complementary levels: at the machine level, the part level, and at the material level. In addition to showcasing hardware platforms for each class, I introduced custom software tools that showed how non-expert end users can design and define assembly protocols for each system. We used these platforms to demonstrate fabrication of advanced, functional artefacts across both scales and use cases, including robots and space structures. Below, I summarize the thesis and point toward future work.

\subsection{Summary of Thesis and Contributions}

In Chapter \ref{sec:Laserfactory}, we introduced a multi-process manufacturing machine called LaserFactory, an integrated fabrication platform that can rapidly create the geometry of a device, create its circuit traces and assemble components without manual intervention. We demonstrated how we can augment an existing laser cutter with a hardware add-on without interfacing with its underlying firmware by using a motion-based signaling technique that can inform the add-on when to start and stop its operation. We illustrated the two main features of our hardware add-on, a silver dispenser used for circuit trace creation and a pick-and-place mechanism used for assembling electronic components, and showed that the add-on can create high-resolution traces of high conductivity and assemble a range of different electronic components. We then showcased laser sintering, a technique that uses a $\mathrm{CO_2}$ laser to cure dispensed silver paste and discussed which laser cutter settings are most suitable to cure the traces. Finally, we showed our end-to-end design and fabrication pipeline consisting of a design tool, a visualization tool, and a post-processing script that transforms the design file into machine instructions for LaserFactory. We also showed example applications that included a quadcopter with actuators, a sensor-enhanced wristband, and a PCB assembled from basic transistors and resistors. Researchers have in recent decades made significant progress toward the long-term vision of being able to download a device file and have it fabricated at the push of a button. While laypeople can today do so for passive, primarily decorative objects via commercially available laser cutters and 3D printers, the fabrication of fully functional, electromechanical devices demonstrated in this chapter took a step toward that shared vision.

By developing multi-process manufacturing machines that consolidate different manufacturing processes into a single platform, we can build mesoscale artefacts like robots in a single machine. This is useful for a variety of applications, but there are many instances that require the ability to re-configure our existing infrastructure for new needs, instead of fabricating them from the ground up. One such application is space structures, like optical telescopes. We need to leverage the most advanced of our manufacturing technologies to make these on earth, but the size of all current space-borne telescopes are limited as a result. One way to overcome this is to still manufacture them on earth, but to do so in modular parts, and enable them to assemble in orbit. To address this, in Chapter \ref{sec:Electrovoxel} we introduced a modular assembly framework to allow structures to be partitioned into modules, transported compactly, and self-assemble in orbit using embedded electromagnets via reconfiguration. It utilizes an electromagnet-based actuation framework to reconfigure in three dimensions via pivoting. While a variety of actuation mechanisms for self-reconfigurable robots have been explored, they often suffer from cost, complexity, assembly and sizing requirements that prevent scaled production of such robots. To address this challenge, we developed an actuation mechanism based on electromagnets embedded into the edges of each cube  to interchangeably create identically or oppositely polarized electromagnet pairs, resulting in repulsive or attractive forces, respectively. By leveraging attraction for hinge formation, and repulsion to drive pivoting maneuvers, we showed how to reconfigure the robot by voxelizing it and actuating its constituent modules\textemdash termed \textit{Electrovoxels}\textemdash via \textit{electromagnetically actuated pivoting}. To demonstrate this, we developed fully untethered, three-dimensional self-reconfigurable robots and demonstrate 2D and 3D self-reconfiguration using pivot and traversal maneuvers on an air-table and in microgravity on a parabolic flight. We described the hardware design of our modules, its pivoting framework, our reconfiguration planning software, and an evaluation of the dynamical and electrical characteristics of our system to inform the design of modular assembly systems using scalable self-reconfigurable structures.

However, a limitation we found for assembly using both machines and self-actuated modular platforms was their requirement to position parts with high resolution and constant power expenditure. For modular assembly in particular, like that demonstrated in Chapter \ref{sec:Electrovoxel} and those developed in the modular robotics community more broadly, another key challenge arises. Assembling high resolution structures requires individual modules to be made increasingly \textit{numerous} and increasingly \textit{small}. Scaling these systems up in number and down in size scale creates a serious challenge for embedding electronics and actuators into individual parts. While our electromagnet-based actuation architecture granted crucial control authority of the assembly procedure in the high-stakes microgravity environment, for assembling high-resolution structures, the cost and algorithmic complexity of actively controlling large numbers of actuators becomes a serious challenge. In addition, while electromagnets are suitable for assembling structures in a microgravity environment, assembling a structure terrestrially in a gravity environment can place significant demands on actuator requirements. To address this, in Chapter \ref{sec:Pullup} we introduce a modular folding-based method that accomplishes assembly without the overhead of electronics or actuators by leveraging two key insights. First, by embedding folding instructions into the parts themselves, we can reduce the many folding steps into a single deployable trajectory actuated by 1 degree of freedom. Second, by reducing the actuated degrees of freedom to 1, we can outsource the actuation that folds the structure to lie offboard, and actuate this manually in the scope of this work. Our solution was to introduce a method to rapidly create 3D geometries from 2D sheets using pull-up nets: a string routed through the planar faces which can be pulled by a user to fold the sheet into its target 3D structure. This provides a way to fold a sheet into its target shape using common string or nylon, using just \textit{a single actuated degree of freedom} controlled by a user. We developed an algorithm run by our web-based software tool that allows users to upload 3D meshes and generate their unfolded geometries. We show how to fabricate these on a commercial laser cutter and route string through the faces which are pulled by a user to fold the sheet into its target 3D structure. We fabricated a variety of polyhedra and organic structures and highlighted a variety of applications. 

Our folding-based modular assembly pipeline provided a low-fidelity rapid prototyping option for 3D structures that is simple, inexpensive and rapid, obviating the need for significant design or fabrication expertise, and served as a compelling alternative to more complicated, actively controlled self-assembly paradigms. However, a limitation with this approach is that the target structure must be known ahead of time, and a particular sheet can be designed to fold into only one specific target shape. To address these limitations, in Chapter \ref{sec:SelectiveMixels} we introduced a method to program magnetic materials themselves to both align and assemble parts, and that allows these materials to be reprogrammed for new target shapes. We formulated a method to program magnetic materials selectively in order to achieve passive assembly under stochastic agitation, without the need for electronics or actuators, in a way that is re-programmable. We accomplished this by developing a method to generate highly selective encodings that can be magnetically "programmed" onto physical modules to enable them to self-assemble in chosen configurations. We generate these encodings based on Hadamard matrices, and show how to design the faces of modules to be maximally attractive to their intended mate, while remaining maximally agnostic to other faces. We derive guarantees on these bounds, and verify their attraction and agnosticism experimentally. Using cubic modules whose faces have been covered in soft magnetic material, we show how inexpensive, passive modules with planar faces can be used to selectively self-assemble into target shapes without geometric guides. We show that these modules can be easily re-programmed for new target shapes using a CNC-based magnetic plotter, and demonstrate self-assembly of 8 cubes in a water tank. In the preceding chapters, we had introduced how to automate fabrication and assembly at the machine level and part level. In this chapter, we developed a method that complements these at the third level, by illustrating how to automate assembly using materials themselves. Together, these provide three hierarchically complementary ways in which to automate manufacturing and assembly in situ using machines, modules, and materials.

\subsection{Future Work}

Avenues for future work have been discussed by chapter, inline with the specific limitations highlighted therein, in order to preserve clarity. In those sections, we discussed possible improvements and research directions that built on the specific implementations we introduced for automating fabrication in situ using hybrid manufacturing machines, modular assembly platforms, and programmable materials. In this section, we will highlight future research avenues for each class on a longer horizon, and comment on prospects for in situ manufacturing as a whole.

Parallel to the ability to fabricate custom hardware automatically and across scales, the ability to design them automatically is emerging in tandem. While generative models in AI have been widely popularized by the appearance of Deepfakes in video creation, related technologies are already employed for hardware creation. AutoDesk, one of the worlds leading CAD developers, has employed generative design in its flagship product since 2017 to suggest and complete users' CAD designs based on functional needs like load requirements. That same year, Autodesk published a tool equipping users with similar abilities for design and assembly of electronic circuits~\cite{anderson2017trigger}. Tools like Midjourney can today generate images, such as "a blueprint for a quadrotor" from natural language alone, and ChatGPT can write basic Gcode today. As Generative tools progress from using textual input for generating aesthetic 2D images, to using textual prompts for generating manufacturable 3D artefacts, I believe that new multi-process fabrication machines will be a key enabler to translating user-defined designs into manufacturable products. An important avenue for future research is to investigate new ways with which to hybridize manufacturing techniques to explore end-to-end production capabilities. 

Developing end-to-end manufacturing machines also opens up new opportunities for sustainability. If we are able to parameterize and automate the full manufacturing process, this means we have a blueprint not just of how to assemble our artefacts, but also, by running those processes in reverse, how to disassemble them. Another key research avenue is to develop machines capable of recycling existing artefacts, harvesting their components, and re-assembling them for new tasks.

Another promising area for future work lies in reconfigurable space systems. Automated assembly methods are key to enabling sustained presence in space by addressing constraints on launch volume, and the James Webb Space Teleschope is just one example of how deployability is key to this. However, current technologies typically only support a single deployable target shape. By developing new hardware architectures, future work can exploit computational techniques capable of reversibly folding and reconfiguring target geometries to create adaptable, large-scale assemblies in space. 

We have shown how to instantiate code-theoretic, selectively paired interfaces in magnetic materials. However, there exist many materials that exhibit the philic and phobic properties that can be exploited for self-assembly, including biological materials, hydrophilic materials and electrostatics. Future research should explore how to program code-theoretic signatures into these materials to investigate this method's ability to allow self-assembly across new materials and scales, particularly for architected materials and biological applications. Beyond self-assembly, another compelling avenue for research is to investigate how to program materials for microrobotics applications, for example by programming materials with local rules that encode behavior. We have demonstrated how programmable materials can be used to assemble target structures in both fluidic environments, but the difficulty of actuating programmed materials against gravity moments remain. Building on this idea, another avenue for future work is to leverage microgravity as a catalyst to develop modular self-assembly technologies here on Earth. Self-assembly is a promising manufacturing paradigm at the small scale for applications ranging from architected materials to microrobotics. While these still require tremendous resources to develop on Earth, space provides a means to accelerate research in self-assembly, by obviating the need to reconfigure against gravity moments; a possibility that is now being catalysed by miniaturization and falling launch costs. By also developing accessible, cm-scale proxies that can be tested in microgravity conditions, interdisciplinary researchers have a new opportunity to develop and test new assembly techniques. 

Finally, I have presented these three classes of manufacturing methods as separate ways to achieve assembly at the machine, part ("module"), and material levels. However, because they operate at different hierarchies, these are ultimately \textit{complementary} ways that can be integrated to support assembly at points of need\textemdash at the macro, meso, and micro scales, for both space and terrestrial applications. Top-down manufacturing and bottom-up assembly methods each create unique opportunities and challenges, and a key line of future work is to explore ways these can be integrated that leverage the best of both worlds.

%% file: references.bib
@article{tibbits2012self,
  title={The self-assembly line},
  author={Tibbits, Skylar},
  year={2012},
  publisher={CUMINCAD}
}

@inproceedings{yasu2017magnetic,
  title={Magnetic plotter: a macrotexture design method using magnetic rubber sheets},
  author={Yasu, Kentaro},
  booktitle={Proceedings of the 2017 CHI Conference on Human Factors in Computing Systems},
  pages={4983--4993},
  year={2017}
}

@inproceedings{tolley2011programmable,
  title={Programmable 3d stochastic fluidic assembly of cm-scale modules},
  author={Tolley, Michael and Lipson, Hod},
  booktitle={2011 IEEE/RSJ International Conference on Intelligent Robots and Systems},
  pages={4366--4371},
  year={2011},
  organization={IEEE}
}

@article{hacohen2015meshing,
  title={Meshing complex macro-scale objects into self-assembling bricks},
  author={Hacohen, Adar and Hanniel, Iddo and Nikulshin, Yasha and Wolfus, Shuki and Abu-Horowitz, Almogit and Bachelet, Ido},
  journal={Scientific reports},
  volume={5},
  number={1},
  pages={1--8},
  year={2015},
  publisher={Nature Publishing Group}
}

@inproceedings{yasu2020magnelayer,
  title={MagneLayer: Force Field Fabrication by Layered Magnetic Sheets},
  author={Yasu, Kentaro},
  booktitle={Proceedings of the 2020 CHI Conference on Human Factors in Computing Systems},
  pages={1--9},
  year={2020}
}

@article{tolley2008dynamically,
  title={Dynamically programmable fluidic assembly},
  author={Tolley, Michael T and Krishnan, Mekala and Erickson, David and Lipson, Hod},
  journal={Applied Physics Letters},
  volume={93},
  number={25},
  pages={254105},
  year={2008},
  publisher={American Institute of Physics}
}

@inproceedings{lo2014shrinkycircuits,
  title={ShrinkyCircuits: sketching, shrinking, and formgiving for electronic circuits},
  author={Lo, Joanne and Paulos, Eric},
  booktitle={Proceedings of the 27th annual ACM symposium on User interface software and technology},
  pages={291--299},
  year={2014}
}

@inproceedings{romanishin20153d,
  title={3D M-Blocks: Self-Reconfiguring robots capable of locomotion via pivoting in three dimensions},
  author={Romanishin, John W and Gilpin, Kyle and Claici, Sebastian and Rus, Daniela},
  booktitle={Robotics and Automation (ICRA), 2015 IEEE International Conference on},
  pages={1925--1932},
  year={2015},
  organization={IEEE}
}

@inproceedings{peng20163d,
  title={A 3d printer for interactive electromagnetic devices},
  author={Peng, Huaishu and Guimbreti{\`e}re, Fran{\c{c}}ois and McCann, James and Hudson, Scott},
  booktitle={Proceedings of the 29th Annual Symposium on User Interface Software and Technology},
  pages={553--562},
  year={2016}
}

@inproceedings{anderson2017trigger,
  title={Trigger-action-circuits: Leveraging generative design to enable novices to design and build circuitry},
  author={Anderson, Fraser and Grossman, Tovi and Fitzmaurice, George},
  booktitle={Proceedings of the 30th Annual ACM Symposium on User Interface Software and Technology},
  pages={331--342},
  year={2017}
}

@article{baca2014modred,
  title={Modred: Hardware design and reconfiguration planning for a high dexterity modular self-reconfigurable robot for extra-terrestrial exploration},
  author={Baca, Jos{\'e} and Hossain, SGM and Dasgupta, Prithviraj and Nelson, Carl A and Dutta, Ayan},
  journal={Robotics and Autonomous Systems},
  volume={62},
  number={7},
  pages={1002--1015},
  year={2014},
  publisher={Elsevier}
}

@article{gilpin2008miche,
  title={Miche: Modular shape formation by self-disassembly},
  author={Gilpin, Kyle and Kotay, Keith and Rus, Daniela and Vasilescu, Iuliu},
  journal={The International Journal of Robotics Research},
  volume={27},
  number={3-4},
  pages={345--372},
  year={2008},
  publisher={Sage Publications Sage UK: London, England}
}

@inproceedings{maccurdy2016printable,
  title={Printable hydraulics: A method for fabricating robots by 3D co-printing solids and liquids},
  author={MacCurdy, Robert and Katzschmann, Robert and Kim, Youbin and Rus, Daniela},
  booktitle={2016 IEEE International Conference on Robotics and Automation (ICRA)},
  pages={3878--3885},
  year={2016},
  organization={IEEE}
}

@inproceedings{roudaut2016cubimorph,
  title={Cubimorph: Designing modular interactive devices},
  author={Roudaut, Anne and Krusteva, Diana and McCoy, Mike and Karnik, Abhijit and Ramani, Karthik and Subramanian, Sriram},
  booktitle={2016 IEEE International Conference on Robotics and Automation (ICRA)},
  pages={3339--3345},
  year={2016},
  organization={IEEE}
}

@article{daudelin2018integrated,
  title={An integrated system for perception-driven autonomy with modular robots},
  author={Daudelin, Jonathan and Jing, Gangyuan and Tosun, Tarik and Yim, Mark and Kress-Gazit, Hadas and Campbell, Mark},
  journal={Science Robotics},
  volume={3},
  number={23},
  year={2018},
  publisher={Science Robotics}
}

@article{neubert2016soldercubes,
  title={Soldercubes: a self-soldering self-reconfiguring modular robot system},
  author={Neubert, Jonas and Lipson, Hod},
  journal={Autonomous Robots},
  volume={40},
  number={1},
  pages={139--158},
  year={2016},
  publisher={Springer}
}

@inproceedings{schmitz2015capricate,
  title={Capricate: A fabrication pipeline to design and 3D print capacitive touch sensors for interactive objects},
  author={Schmitz, Martin and Khalilbeigi, Mohammadreza and Balwierz, Matthias and Lissermann, Roman and M{\"u}hlh{\"a}user, Max and Steimle, J{\"u}rgen},
  booktitle={Proceedings of the 28th Annual ACM Symposium on User Interface Software \& Technology},
  pages={253--258},
  year={2015}
}

@inproceedings{an2008cube,
  title={Em-cube: cube-shaped, self-reconfigurable robots sliding on structure surfaces},
  author={An, Byoung Kwon},
  booktitle={Robotics and Automation, 2008. ICRA 2008. IEEE International Conference on},
  pages={3149--3155},
  year={2008},
  organization={IEEE}
}

@inproceedings{gilpin2011making,
  title={Making self-disassembling objects with multiple components in the robot pebbles system},
  author={Gilpin, Kyle and Koyanagi, Kent and Rus, Daniela},
  booktitle={Robotics and Automation (ICRA), 2011 IEEE International Conference on},
  pages={3614--3621},
  year={2011},
  organization={IEEE}
}

@inproceedings{yim2000polybot,
  title={PolyBot: a modular reconfigurable robot},
  author={Yim, Mark and Duff, David G and Roufas, Kimon D},
  booktitle={Proceedings 2000 ICRA. Millennium Conference. IEEE International Conference on Robotics and Automation. Symposia Proceedings (Cat. No. 00CH37065)},
  volume={1},
  pages={514--520},
  year={2000},
  organization={IEEE}
}

@article{sprowitz2014roombots,
  title={Roombots: A hardware perspective on 3D self-reconfiguration and locomotion with a homogeneous modular robot},
  author={Spr{\"o}witz, Alexander and Moeckel, Rico and Vespignani, Massimo and Bonardi, St{\'e}phane and Ijspeert, Auke Jan},
  journal={Robotics and Autonomous Systems},
  volume={62},
  number={7},
  pages={1016--1033},
  year={2014},
  publisher={Elsevier}
}

@article{yim2003modular,
  title={Modular reconfigurable robots in space applications},
  author={Yim, Mark and Roufas, Kimon and Duff, David and Zhang, Ying and Eldershaw, Craig and Homans, Sam},
  journal={Autonomous Robots},
  volume={14},
  number={2},
  pages={225--237},
  year={2003},
  publisher={Springer}
}

@article{felton2014method,
  title={A method for building self-folding machines},
  author={Felton, Samuel and Tolley, Michael and Demaine, Erik and Rus, Daniela and Wood, Robert},
  journal={Science},
  volume={345},
  number={6197},
  pages={644--646},
  year={2014},
  publisher={American Association for the Advancement of Science}
}

@inproceedings{nisser2016feedback,
  title={Feedback-controlled self-folding of autonomous robot collectives},
  author={Nisser, Martin EW and Felton, Samuel M and Tolley, Michael T and Rubenstein, Michael and Wood, Robert J},
  booktitle={2016 IEEE/RSJ International Conference on Intelligent Robots and Systems (IROS)},
  pages={1254--1261},
  year={2016},
  organization={IEEE}
}

@inproceedings{nisser2021laserfactory,
  title={LaserFactory: A Laser Cutter-based Electromechanical Assembly and Fabrication Platform to Make Functional Devices \& Robots},
  author={Nisser, Martin and Liao, Christina Chen and Chai, Yuchen and Adhikari, Aradhana and Hodges, Steve and Mueller, Stefanie},
  booktitle={Proceedings of the 2021 CHI Conference on Human Factors in Computing Systems},
  pages={1--15},
  year={2021}
}

@inproceedings{mueller2013laserorigami,
  title={LaserOrigami: laser-cutting 3D objects},
  author={Mueller, Stefanie and Kruck, Bastian and Baudisch, Patrick},
  booktitle={Proceedings of the SIGCHI Conference on Human Factors in Computing Systems},
  pages={2585--2592},
  year={2013}
}

@inproceedings{abtahi2018visuo,
  title={Visuo-Haptic Illusions for Improving the Perceived Performance of Shape Displays},
  author={Abtahi, P. and Follmer, S.},
  booktitle={Proc. of CHI '18},
  pages={150},
  year={2018},
  organization={ACM}
}

@inproceedings{romanishin2013m,
  title={M-blocks: Momentum-driven, magnetic modular robots},
  author={Romanishin, John W and Gilpin, Kyle and Rus, Daniela},
  booktitle={2013 IEEE/RSJ International Conference on Intelligent Robots and Systems},
  pages={4288--4295},
  year={2013},
  organization={IEEE}
}

@article{rus2001crystalline,
  title={Crystalline robots: Self-reconfiguration with compressible unit modules},
  author={Rus, Daniela and Vona, Marsette},
  journal={Autonomous Robots},
  volume={10},
  number={1},
  pages={107--124},
  year={2001},
  publisher={Springer}
}

@inproceedings{sung2015reconfiguration,
  title={Reconfiguration planning for pivoting cube modular robots},
  author={Sung, Cynthia and Bern, James and Romanishin, John and Rus, Daniela},
  booktitle={2015 IEEE international conference on robotics and automation (ICRA)},
  pages={1933--1940},
  year={2015},
  organization={IEEE}
}

@inproceedings{gilpin2010robot,
  title={Robot pebbles: One centimeter modules for programmable matter through self-disassembly},
  author={Gilpin, Kyle and Knaian, Ara and Rus, Daniela},
  booktitle={2010 IEEE International Conference on Robotics and Automation},
  pages={2485--2492},
  year={2010},
  organization={IEEE}
}

@article{stoy2010self,
  title={Self-reconfigurable robots: an introduction},
  author={Stoy, Kasper and Brandt, David and Christensen, David J and Brandt, David},
  year={2010},
  publisher={Mit Press Cambridge}
}

@inproceedings{le2016zooids,
  title={Zooids: Building blocks for swarm user interfaces},
  author={Le Goc, Mathieu and Kim, Lawrence H and Parsaei, Ali and Fekete, Jean-Daniel and Dragicevic, Pierre and Follmer, Sean},
  booktitle={Proceedings of the 29th Annual Symposium on User Interface Software and Technology},
  pages={97--109},
  year={2016}
}

@inproceedings{umapathi2015laserstacker,
  title={LaserStacker: Fabricating 3D objects by laser cutting and welding},
  author={Umapathi, Udayan and Chen, Hsiang-Ting and Mueller, Stefanie and Wall, Ludwig and Seufert, Anna and Baudisch, Patrick},
  booktitle={Proceedings of the 28th Annual ACM Symposium on User Interface Software \& Technology},
  pages={575--582},
  year={2015}
}

@article{nisser2017electromagnetically,
  title={An electromagnetically actuated, self-reconfigurable space structure},
  author={Nisser, Martin and Izzo, Dario and Borggraefe, Andreas},
  year={2017}
}

@article{hauser2020kubits,
  title={Kubits: solid-state self-reconfiguration with programmable magnets},
  author={Hauser, Simon and Mutlu, Mehmet and Ijspeert, Auke J},
  journal={IEEE Robotics and Automation Letters},
  volume={5},
  number={4},
  pages={6443--6450},
  year={2020},
  publisher={IEEE}
}

@article{yim2007modular,
  title={Modular self-reconfigurable robot systems [grand challenges of robotics]},
  author={Yim, Mark and Shen, Wei-Min and Salemi, Behnam and Rus, Daniela and Moll, Mark and Lipson, Hod and Klavins, Eric and Chirikjian, Gregory S},
  journal={IEEE Robotics \& Automation Magazine},
  volume={14},
  number={1},
  pages={43--52},
  year={2007},
  publisher={IEEE}
}

@inproceedings{lu2021enumeration,
  title={Enumeration of polyominoes \& polycubes composed of magnetic cubes},
  author={Lu, Yitong and Bhattacharjee, Anuruddha and Biediger, Daniel and Kim, MinJun and Becker, Aaron T},
  booktitle={2021 IEEE/RSJ International Conference on Intelligent Robots and Systems (IROS)},
  pages={6977--6982},
  year={2021},
  organization={IEEE}
}

@article{bowden1997self,
  title={Self-assembly of mesoscale objects into ordered two-dimensional arrays},
  author={Bowden, Ned and Terfort, Andreas and Carbeck, Jeff and Whitesides, George M},
  journal={Science},
  volume={276},
  number={5310},
  pages={233--235},
  year={1997},
  publisher={American Association for the Advancement of Science}
}

@article{grzybowski2003electrostatic,
  title={Electrostatic self-assembly of macroscopic crystals using contact electrification},
  author={Grzybowski, Bartosz A and Winkleman, Adam and Wiles, Jason A and Brumer, Yisroel and Whitesides, George M},
  journal={Nature materials},
  volume={2},
  number={4},
  pages={241--245},
  year={2003},
  publisher={Nature Publishing Group}
}

@inproceedings{tolley2010fluidic,
  title={Fluidic manipulation for scalable stochastic 3D assembly of modular robots},
  author={Tolley, Michael T and Lipson, Hod},
  booktitle={2010 IEEE international conference on robotics and automation},
  pages={2473--2478},
  year={2010},
  organization={IEEE}
}

@article{papadopoulou2017self,
  title={From Self-Assembly to Evolutionary Structures},
  author={Papadopoulou, Athina and Laucks, Jared and Tibbits, Skylar},
  journal={Architectural Design},
  volume={87},
  number={4},
  pages={28--37},
  year={2017},
  publisher={Wiley Online Library}
}

@article{krishnan2008increased,
  title={Increased robustness for fluidic self-assembly},
  author={Krishnan, Mekala and Tolley, Michael T and Lipson, Hod and Erickson, David},
  journal={Physics of Fluids},
  volume={20},
  number={7},
  pages={073304},
  year={2008},
  publisher={American Institute of Physics}
}

@article{kalontarov2010hydrodynamically,
  title={Hydrodynamically driven docking of blocks for 3D fluidic assembly},
  author={Kalontarov, Michael and Tolley, Michael T and Lipson, Hod and Erickson, David},
  journal={Microfluidics and Nanofluidics},
  volume={9},
  number={2},
  pages={551--558},
  year={2010},
  publisher={Springer}
}

@article{jilek2021towards,
  title={Towards a passive self-assembling macroscale multi-robot system},
  author={J{\'\i}lek, Martin and Somr, Michael and Kulich, Miroslav and Zeman, Jan and P{\v{r}}eu{\v{c}}il, Libor},
  journal={IEEE Robotics and Automation Letters},
  volume={6},
  number={4},
  pages={7293--7300},
  year={2021},
  publisher={IEEE}
}

@inproceedings{jilek2020centimeter,
  title={Centimeter-scaled self-assembly: A preliminary study},
  author={J{\i}lek, Martin and Kulich, Miroslav and Preucil, Libor},
  booktitle={Proceedings of the 17th International Conference on Informatics in Control, Automation and Robotics (ICINCO). ScitePress},
  pages={438--445},
  year={2020}
}

@inproceedings{zykov2007experiment,
  title={Experiment design for stochastic three-dimensional reconfiguration of modular robots},
  author={Zykov, Victor and Lipson, Hod and others},
  booktitle={IEEE Int. Conf. Intell. Robots Syst., Self-Reconfigurable Robot. Workshop, San Diego, CA},
  year={2007},
  organization={Citeseer}
}

@inproceedings{tsutsumi2007multistate,
  title={Multistate part for mesoscale self-assembly},
  author={Tsutsumi, Daiko and Murata, Satoshi},
  booktitle={SICE Annual Conference 2007},
  pages={890--895},
  year={2007},
  organization={IEEE}
}

@article{miyashita2009influence,
  title={The influence of shape on parallel self-assembly},
  author={Miyashita, Shuhei and Nagy, Zolt{\'a}n and Nelson, Bradley J and Pfeifer, Rolf},
  journal={Entropy},
  volume={11},
  number={4},
  pages={643--666},
  year={2009},
  publisher={Molecular Diversity Preservation International}
}

@article{whitesides1991molecular,
  title={Molecular self-assembly and nanochemistry: a chemical strategy for the synthesis of nanostructures},
  author={Whitesides, George M and Mathias, John P and Seto, Christopher T},
  journal={Science},
  volume={254},
  number={5036},
  pages={1312--1319},
  year={1991},
  publisher={American Association for the Advancement of Science}
}

@article{beaman1991selective,
  title={Selective laser sintering with assisted powder handling. 1990},
  author={Beaman, JJ and Deckard, CR},
  journal={Google Patents},
  year={1991}
}

@article{meiners1998shaped,
  title={Shaped body especially prototype or replacement part production},
  author={Meiners, W and Wissenbach, K and Gasser, A},
  journal={DE Patent},
  volume={19},
  year={1998}
}

@misc{crump1992apparatus,
  title={Apparatus and method for creating three-dimensional objects},
  author={Crump, S Scott},
  year={1992},
  publisher={Google Patents},
  note={US Patent 5,121,329}
}

@inproceedings{zhu2020curveboards,
  title={CurveBoards: Integrating breadboards into physical objects to prototype function in the context of form},
  author={Zhu, Junyi and Blumberg, Lotta-Gili and Zhu, Yunyi and Nisser, Martin and Carlson, Ethan Levi and Wen, Xin and Shum, Kevin and Quaye, Jessica Ayeley and Mueller, Stefanie},
  booktitle={Proceedings of the 2020 CHI Conference on Human Factors in Computing Systems},
  pages={1--13},
  year={2020}
}

@inproceedings{nisser2022mixels,
  title={Mixels: Fabricating Interfaces using Programmable Magnetic Pixels},
  author={Nisser, Martin and Makaram, Yashaswini and Covarrubias, Lucian and Bah, Amadou Yaye and Faruqi, Faraz and Suzuki, Ryo and Mueller, Stefanie},
  booktitle={Proceedings of the 35th Annual ACM Symposium on User Interface Software and Technology},
  pages={1--12},
  year={2022}
}

@article{kianian2017wohlers,
  title={Wohlers report 2017: 3d printing and additive manufacturing state of the industry, annual worldwide progress report: Chapters titles: The middle east, and other countries},
  author={Kianian, Babak},
  year={2017},
  publisher={Wohlers Associates, Inc.}
}

@misc{Forbes3dp,
  title = {Forbes, Printing Industry},
  howpublished = {\url{https://www.forbes.com/sites/michaelmolitch-hou/2022/04/25/three-areas-holding-back-the-106b-3d-printing-industry/?sh=65132bf84935/}},
  note = {Accessed: 2023-02-25}
}

@misc{CorrMag,
  title = {Correlated Magnetics},
  howpublished = {\url{http://www.polymagnet.com/}},
  note = {Accessed: 2021-11-25}
}

@techreport{hagberg2008exploring,
  title={Exploring network structure, dynamics, and function using NetworkX},
  author={Hagberg, Aric and Swart, Pieter and S Chult, Daniel},
  year={2008},
  institution={Los Alamos National Lab.(LANL), Los Alamos, NM (United States)}
}

@article{bron1973algorithm,
  title={Algorithm 457: finding all cliques of an undirected graph},
  author={Bron, Coen and Kerbosch, Joep},
  journal={Communications of the ACM},
  volume={16},
  number={9},
  pages={575--577},
  year={1973},
  publisher={ACM New York, NY, USA}
}

@inproceedings{nisser2022electrovoxel,
  title={ElectroVoxel: Electromagnetically actuated pivoting for scalable modular self-reconfigurable robots},
  author={Nisser, Martin and Cheng, Leon and Makaram, Yashaswini and Suzuki, Ryo and Mueller, Stefanie},
  booktitle={2022 International Conference on Robotics and Automation (ICRA)},
  pages={4254--4260},
  year={2022},
  organization={IEEE}
}

@article{nisser2022stochastic,
  title={Stochastic Self-Assembly with Magnetically Re-programmable Voxels},
  author={Nisser, Martin and Makaram, Yashaswini and Mueller, Stefanie},
  year={2022}
  }

@inproceedings{nisser2021programmable,
  title={Programmable Polarities: Actuating Interactive Prototypes with Programmable Electromagnets},
  author={Nisser, Martin and Cheng, Leon and Makaram, Yashaswini and Suzuki, Ryo and Mueller, Stefanie},
  booktitle={The Adjunct Publication of the 34th Annual ACM Symposium on User Interface Software and Technology},
  pages={121--123},
  year={2021}
}

@misc{quinlan2017industrial,
  title={Industrial and consumer uses of additive manufacturing: A discussion of capabilities, trajectories, and challenges},
  author={Quinlan, Haden Edward and Hasan, Talha and Jaddou, John and Hart, A John},
  journal={Journal of industrial ecology},
  volume={21},
  number={S1},
  pages={S15--S20},
  year={2017},
  publisher={Wiley Online Library}
}

@online{Voxel819,
  year =         "2019",
  title =        "Voxel8",
  author="Voxel8",
  organization="Voxel8", 
  url =          "https://www.voxel8.com/",
  month =        nov,
  lastaccessed = "November 21, 2019",
}

@article{valentine2017hybrid,
  title={Hybrid 3D printing of soft electronics},
  author={Valentine, Alexander D and Busbee, Travis A and Boley, John William and Raney, Jordan R and Chortos, Alex and Kotikian, Arda and Berrigan, John Daniel and Durstock, Michael F and Lewis, Jennifer A},
  journal={advanced Materials},
  volume={29},
  number={40},
  pages={1703817},
  year={2017},
  publisher={Wiley Online Library}
}

@inproceedings{rubenstein2012kilobot,
  title={Kilobot: A low cost scalable robot system for collective behaviors},
  author={Rubenstein, Michael and Ahler, Christian and Nagpal, Radhika},
  booktitle={2012 IEEE international conference on robotics and automation},
  pages={3293--3298},
  year={2012},
  organization={IEEE}
}

@article{kosmal2022hybrid,
  title={Hybrid additive robotic workcell for autonomous fabrication of mechatronic systems-A case study of drone fabrication},
  author={Kosmal, Tadeusz and Beaumont, Kieran and Link, Eric and Phillips, Dalton and Pulling, Conner and Wotton, Heather and Kudrna, Camille and Kubalak, Joseph and Williams, Christopher},
  journal={Additive Manufacturing Letters},
  volume={3},
  pages={100100},
  year={2022},
  publisher={Elsevier}
}

@article{genina2012tailoring,
  title={Tailoring controlled-release oral dosage forms by combining inkjet and flexographic printing techniques},
  author={Genina, Natalja and Fors, Daniela and Vakili, Hossein and Ihalainen, Petri and Pohjala, Leena and Ehlers, Henrik and Kassamakov, Ivan and Haeggstr{\"o}m, Edward and Vuorela, Pia and Peltonen, Jouko and others},
  journal={European journal of pharmaceutical sciences},
  volume={47},
  number={3},
  pages={615--623},
  year={2012},
  publisher={Elsevier}
}

@inproceedings{nisser2022selective,
  title={Selective Self-Assembly using Re-Programmable Magnetic Pixels},
  author={Nisser, Martin and Makaram, Yashaswini and Faruqi, Faraz and Suzuki, Ryo and Mueller, Stefanie},
  booktitle={2022 IEEE/RSJ International Conference on Intelligent Robots and Systems (IROS)},
  pages={12659--12666},
  year={2022},
  organization={IEEE}
}

@inproceedings{mellis2013microcontrollers,
  title={Microcontrollers as material: crafting circuits with paper, conductive ink, electronic components, and an" untoolkit"},
  author={Mellis, David A and Jacoby, Sam and Buechley, Leah and Perner-Wilson, Hannah and Qi, Jie},
  booktitle={Proceedings of the 7th International Conference on Tangible, Embedded and Embodied Interaction},
  pages={83--90},
  year={2013}
}

@inproceedings{savage2012midas,
  title={Midas: fabricating custom capacitive touch sensors to prototype interactive objects},
  author={Savage, Valkyrie and Zhang, Xiaohan and Hartmann, Bj{\"o}rn},
  booktitle={Proceedings of the 25th annual ACM symposium on User interface software and technology},
  pages={579--588},
  year={2012}
}

@inproceedings{savage2014series,
  title={A series of tubes: adding interactivity to 3D prints using internal pipes},
  author={Savage, Valkyrie and Schmidt, Ryan and Grossman, Tovi and Fitzmaurice, George and Hartmann, Bj{\"o}rn},
  booktitle={Proceedings of the 27th annual ACM symposium on User interface software and technology},
  pages={3--12},
  year={2014}
}

@article{umetani2017surfcuit,
  title={SurfCuit: surface-mounted circuits on 3D prints},
  author={Umetani, Nobuyuki and Schmidt, Ryan},
  journal={IEEE computer graphics and applications},
  volume={37},
  number={3},
  pages={52--60},
  year={2017},
  publisher={IEEE}
}

@inproceedings{yamaoka2019foldtronics,
  title={FoldTronics: Creating 3D objects with integrated electronics using foldable honeycomb structures},
  author={Yamaoka, Junichi and Dogan, Mustafa Doga and Bulovic, Katarina and Saito, Kazuya and Kawahara, Yoshihiro and Kakehi, Yasuaki and Mueller, Stefanie},
  booktitle={Proceedings of the 2019 chi conference on human factors in computing systems},
  pages={1--14},
  year={2019}
}

@inproceedings{groeger2019lasec,
  title={Lasec: Instant fabrication of stretchable circuits using a laser cutter},
  author={Groeger, Daniel and Steimle, J{\"u}rgen},
  booktitle={Proceedings of the 2019 CHI Conference on Human Factors in Computing Systems},
  pages={1--14},
  year={2019}
}

@inproceedings{kawahara2013instant,
  title={Instant inkjet circuits: lab-based inkjet printing to support rapid prototyping of UbiComp devices},
  author={Kawahara, Yoshihiro and Hodges, Steve and Cook, Benjamin S and Zhang, Cheng and Abowd, Gregory D},
  booktitle={Proceedings of the 2013 ACM international joint conference on Pervasive and ubiquitous computing},
  pages={363--372},
  year={2013}
}

@inproceedings{schmitz2019trilaterate,
  title={./trilaterate: A Fabrication Pipeline to Design and 3D Print Hover-, Touch-, and Force-Sensitive Objects},
  author={Schmitz, Martin and Stitz, Martin and M{\"u}ller, Florian and Funk, Markus and M{\"u}hlh{\"a}user, Max},
  booktitle={Proceedings of the 2019 CHI Conference on Human Factors in Computing Systems},
  pages={1--13},
  year={2019}
}

@inproceedings{schmitz2017flexibles,
  title={Flexibles: deformation-aware 3D-printed tangibles for capacitive touchscreens},
  author={Schmitz, Martin and Steimle, J{\"u}rgen and Huber, Jochen and Dezfuli, Niloofar and M{\"u}hlh{\"a}user, Max},
  booktitle={Proceedings of the 2017 CHI Conference on Human Factors in Computing Systems},
  pages={1001--1014},
  year={2017}
}

@inproceedings{swaminathan2019fiberwire,
  title={Fiberwire: Embedding electronic function into 3d printed mechanically strong, lightweight carbon fiber composite objects},
  author={Swaminathan, Saiganesh and Ozutemiz, Kadri Bugra and Majidi, Carmel and Hudson, Scott E},
  booktitle={Proceedings of the 2019 CHI Conference on Human Factors in Computing Systems},
  pages={1--11},
  year={2019}
}

@inproceedings{lambrichts2020diy,
  title={DIY Fabrication of High Performance Multi-Layered Flexible PCBs},
  author={Lambrichts, Mannu and Tijerina, Jose Maria and De Weyer, Tom and Ramakers, Raf},
  booktitle={Proceedings of the Fourteenth International Conference on Tangible, Embedded, and Embodied Interaction},
  pages={565--571},
  year={2020}
}

@article{balliu2018selective,
  title={Selective laser sintering of inkjet-printed silver nanoparticle inks on paper substrates to achieve highly conductive patterns},
  author={Balliu, Enkeleda and Andersson, Henrik and Engholm, Magnus and {\"O}hlund, Thomas and Nilsson, Hans-Erik and Olin, H{\aa}kan},
  journal={Scientific reports},
  volume={8},
  number={1},
  pages={10408},
  year={2018},
  publisher={Nature Publishing Group UK London}
}

@inproceedings{katakura20193d,
  title={A 3D printer head as a robotic manipulator},
  author={Katakura, Shohei and Kuroki, Yuto and Watanabe, Keita},
  booktitle={Proceedings of the 32nd Annual ACM Symposium on User Interface Software and Technology},
  pages={535--548},
  year={2019}
}

@inproceedings{fukuchi2012laser,
  title={Laser cooking: a novel culinary technique for dry heating using a laser cutter and vision technology},
  author={Fukuchi, Kentaro and Jo, Kazuhiro and Tomiyama, Akifumi and Takao, Shunsuke},
  booktitle={Proceedings of the ACM multimedia 2012 workshop on Multimedia for cooking and eating activities},
  pages={55--58},
  year={2012}
}

@inproceedings{saakes2013paccam,
  title={PacCAM: material capture and interactive 2D packing for efficient material usage on CNC cutting machines},
  author={Saakes, Daniel and Cambazard, Thomas and Mitani, Jun and Igarashi, Takeo},
  booktitle={Proceedings of the 26th annual ACM symposium on User interface software and technology},
  pages={441--446},
  year={2013}
}

@inproceedings{weichel2015reform,
  title={ReForm: integrating physical and digital design through bidirectional fabrication},
  author={Weichel, Christian and Hardy, John and Alexander, Jason and Gellersen, Hans},
  booktitle={Proceedings of the 28th Annual ACM Symposium on User Interface Software \& Technology},
  pages={93--102},
  year={2015}
}

@inproceedings{teibrich2015patching,
  title={Patching physical objects},
  author={Teibrich, Alexander and Mueller, Stefanie and Guimbreti{\`e}re, Fran{\c{c}}ois and Kovacs, Robert and Neubert, Stefan and Baudisch, Patrick},
  booktitle={Proceedings of the 28th Annual ACM Symposium on User Interface Software \& Technology},
  pages={83--91},
  year={2015}
}

@inproceedings{mueller2015scotty,
  title={Scotty: Relocating physical objects across distances using destructive scanning, encryption, and 3D printing},
  author={Mueller, Stefanie and Fritzsche, Martin and Kossmann, Jan and Schneider, Maximilian and Striebel, Jonathan and Baudisch, Patrick},
  booktitle={Proceedings of the Ninth International Conference on Tangible, Embedded, and Embodied Interaction},
  pages={233--240},
  year={2015}
}

@inproceedings{gao2015revomaker,
  title={RevoMaker: Enabling multi-directional and functionally-embedded 3D printing using a rotational cuboidal platform},
  author={Gao, Wei and Zhang, Yunbo and Nazzetta, Diogo C and Ramani, Karthik and Cipra, Raymond J},
  booktitle={Proceedings of the 28th annual ACM symposium on user interface software \& technology},
  pages={437--446},
  year={2015}
}

@inproceedings{vasquez2020jubilee,
  title={Jubilee: An extensible machine for multi-tool fabrication},
  author={Vasquez, Joshua and Twigg-Smith, Hannah and Tran O'Leary, Jasper and Peek, Nadya},
  booktitle={Proceedings of the 2020 CHI Conference on Human Factors in Computing Systems},
  pages={1--13},
  year={2020}
}

@inproceedings{wang2016xprint,
  title={xprint: A modularized liquid printer for smart materials deposition},
  author={Wang, Guanyun and Yao, Lining and Wang, Wen and Ou, Jifei and Cheng, Chin-Yi and Ishii, Hiroshi},
  booktitle={Proceedings of the 2016 CHI conference on human factors in computing systems},
  pages={5743--5752},
  year={2016}
}

@article{kawahara2014building,
  title={Building functional prototypes using conductive inkjet printing},
  author={Kawahara, Yoshihiro and Hodges, Steve and Gong, Nan-Wei and Olberding, Simon and Steimle, J{\"u}rgen},
  journal={IEEE Pervasive computing},
  volume={13},
  number={3},
  pages={30--38},
  year={2014},
  publisher={IEEE}
}

@inproceedings{burstyn2015printput,
  title={Printput: Resistive and capacitive input widgets for interactive 3D prints},
  author={Burstyn, Jesse and Fellion, Nicholas and Strohmeier, Paul and Vertegaal, Roel},
  booktitle={IFIP Conference on Human-Computer Interaction},
  pages={332--339},
  year={2015},
  publisher={Springer International Publishing},
  address={Bamberg, Germany},
  organization={Springer}
}

@inproceedings{hodges2014circuitstickers,
author = {Hodges, Steve and Villar, Nicolas and Chen, Nicholas and Chugh, Tushar and Qi, Jie and Nowacka, Diana and Kawahara, Yoshihiro},
title = {Circuit Stickers: Peel-and-Stick Construction of Interactive Electronic Prototypes},
year = {2014},
isbn = {9781450324731},
publisher = {Association for Computing Machinery},
address = {New York, NY, USA},
url = {https://doi.org/10.1145/2556288.2557150},
doi = {10.1145/2556288.2557150},
booktitle = {Proceedings of the SIGCHI Conference on Human Factors in Computing Systems},
pages = {1743–1746},
numpages = {4},
location = {Toronto, Ontario, Canada},
series = {CHI '14}
}

@article{lewis2015device,
  title={Device fabrication: Three-dimensional printed electronics},
  author={Lewis, Jennifer A and Ahn, Bok Y},
  journal={Nature},
  volume={518},
  number={7537},
  pages={42--43},
  year={2015},
  publisher={Nature Publishing Group}
}

@article{chennareddy2017modular,
  title={Modular self-reconfigurable robotic systems: a survey on hardware architectures},
  author={Chennareddy, S and Agrawal, Anita and Karuppiah, Anupama and others},
  journal={Journal of Robotics},
  volume={2017},
  year={2017},
  publisher={Hindawi}
}

@article{haghighat2016fluid,
  title={Fluid-mediated stochastic self-assembly at centimetric and sub-millimetric scales: Design, modeling, and control},
  author={Haghighat, Bahar and Mastrangeli, Massimo and Mermoud, Gr{\'e}gory and Schill, Felix and Martinoli, Alcherio},
  journal={Micromachines},
  volume={7},
  number={8},
  pages={138},
  year={2016},
  publisher={MDPI}
}

@inproceedings{ekblaw2018tesserae,
  title={TESSERAE: Self-assembling shell structures for space exploration},
  author={Ekblaw, Ariel and Paradiso, Joseph},
  booktitle={Proceedings of IASS Annual Symposia},
  volume={2018},
  number={1},
  pages={1--8},
  year={2018},
  organization={International Association for Shell and Spatial Structures (IASS)}
}

@inproceedings{osborne1997cube,
  title={EM-Cube: An architecture for low-cost real-time volume rendering},
  author={Osborne, R{\"a}ndy and Pfister, Hanspeter and Lauer, Hugh and Ohkami, TakaHide and McKenzie, Neil and Gibson, Sarah and Hiatt, Wally},
  booktitle={Proceedings of the ACM SIGGRAPH/EUROGRAPHICS workshop on Graphics hardware},
  pages={131--138},
  year={1997}
}

@article{piranda2013new,
  title={A new concept of planar self-reconfigurable modular robot for conveying microparts},
  author={Piranda, Beno{\^\i}t and Laurent, Guillaume J and Bourgeois, Julien and Cl{\'e}vy, C{\'e}dric and M{\"o}bes, Sebastian and Le Fort-Piat, Nadine},
  journal={Mechatronics},
  volume={23},
  number={7},
  pages={906--915},
  year={2013},
  publisher={Elsevier}
}

@misc{schweikardt2007stickybricks,
  title={Stickybricks: An adhesion-based modular self-reconfigurable robotic system},
  author={Schweikardt, Eric and Sitti, Metin},
  year={2007}
}

@article{pandey2016assembly,
  title={Assembly of a 3D cellular computer using folded E-blocks},
  author={Pandey, Shivendra and Macias, Nicholas J and Ciobanu, Carmen and Yoon, ChangKyu and Teuscher, Christof and Gracias, David H},
  journal={Micromachines},
  volume={7},
  number={5},
  pages={78},
  year={2016},
  publisher={MDPI}
}

@article{yim2014softcubes,
  title={SoftCubes: Stretchable and self-assembling three-dimensional soft modular matter},
  author={Yim, Sehyuk and Sitti, Metin},
  journal={The International Journal of Robotics Research},
  volume={33},
  number={8},
  pages={1083--1097},
  year={2014},
  publisher={SAGE Publications Sage UK: London, England}
}

@inbook{programmablefilament2020,
author = {Takahashi, Haruki and Punpongsanon, Parinya and Kim, Jeeeun},
title = {Programmable Filament: Printed Filaments for Multi-Material 3D Printing},
year = {2020},
isbn = {9781450375146},
publisher = {Association for Computing Machinery},
address = {New York, NY, USA},
url = {https://doi.org/10.1145/3379337.3415863},
booktitle = {Proceedings of the 33rd Annual ACM Symposium on User Interface Software and Technology},
pages = {1209–1221},
numpages = {13}
}

@inproceedings{hapticprint2015,
author = {Torres, Cesar and Campbell, Tim and Kumar, Neil and Paulos, Eric},
title = {HapticPrint: Designing Feel Aesthetics for Digital Fabrication},
year = {2015},
isbn = {9781450337793},
publisher = {Association for Computing Machinery},
address = {New York, NY, USA},
url = {https://doi.org/10.1145/2807442.2807492},
doi = {10.1145/2807442.2807492},
booktitle = {Proceedings of the 28th Annual ACM Symposium on User Interface Software and Technology},
pages = {583–591},
numpages = {9},
location = {Charlotte, NC, USA},
series = {UIST '15}
}

@inproceedings{printedspeakers2014,
author = {Ishiguro, Yoshio and Poupyrev, Ivan},
title = {3D Printed Interactive Speakers},
year = {2014},
isbn = {9781450324731},
publisher = {Association for Computing Machinery},
address = {New York, NY, USA},
url = {https://doi.org/10.1145/2556288.2557046},
doi = {10.1145/2556288.2557046},
booktitle = {Proceedings of the SIGCHI Conference on Human Factors in Computing Systems},
pages = {1733–1742},
numpages = {10},
location = {Toronto, Ontario, Canada},
series = {CHI '14}
}

@inproceedings{lenticular2021,
author = {Zeng, Jiani and Deng, Honghao and Zhu, Yunyi and Wessely, Michael and Kilian, Axel and Mueller, Stefanie},
title = {Lenticular Objects: 3D Printed Objects with Lenticular Lens Surfaces That Can Change Their Appearance Depending on the Viewpoint},
year = {2021},
isbn = {9781450386357},
publisher = {Association for Computing Machinery},
address = {New York, NY, USA},
url = {https://doi.org/10.1145/3472749.3474815},
doi = {10.1145/3472749.3474815},
booktitle = {The 34th Annual ACM Symposium on User Interface Software and Technology},
pages = {1184–1196},
numpages = {13},
location = {Virtual Event, USA},
series = {UIST '21}
}

@inproceedings{bumpahead2015,
author = {Yasu, Kentaro and Katsumoto, Yuichiro},
title = {Bump Ahead: Easy-to-Design Haptic Surface Using Magnet Array},
year = {2015},
isbn = {9781450339254},
publisher = {Association for Computing Machinery},
address = {New York, NY, USA},
url = {https://doi.org/10.1145/2818466.2818478},
doi = {10.1145/2818466.2818478},
booktitle = {SIGGRAPH Asia 2015 Emerging Technologies},
articleno = {3},
numpages = {3},
location = {Kobe, Japan},
series = {SA '15}
}

@inproceedings{fluxpaper2015,
author = {Ogata, Masa and Fukumoto, Masaaki},
title = {FluxPaper: Reinventing Paper with Dynamic Actuation Powered by Magnetic Flux},
year = {2015},
isbn = {9781450331456},
publisher = {Association for Computing Machinery},
address = {New York, NY, USA},
url = {https://doi.org/10.1145/2702123.2702516},
doi = {10.1145/2702123.2702516},
booktitle = {Proceedings of the 33rd Annual ACM Conference on Human Factors in Computing Systems},
pages = {29–38},
numpages = {10},
location = {Seoul, Republic of Korea},
series = {CHI '15}
}

@inproceedings{magnetact2019,
author = {Yasu, Kentaro},
title = {Magnetact: Magnetic-Sheet-Based Haptic Interfaces for Touch Devices},
year = {2019},
isbn = {9781450359702},
publisher = {Association for Computing Machinery},
address = {New York, NY, USA},
url = {https://doi.org/10.1145/3290605.3300470},
doi = {10.1145/3290605.3300470},
booktitle = {Proceedings of the 2019 CHI Conference on Human Factors in Computing Systems},
pages = {1–8},
numpages = {8},
location = {Glasgow, Scotland Uk},
series = {CHI '19}
}

@inbook{magnetactanimals2021,
author = {Yasu, Kentaro and Ishikawa, Masaya},
title = {Magnetact Animals: A Simple Kinetic Toy Kit for a Creative Online Workshop for Children},
year = {2021},
isbn = {9781450380959},
publisher = {Association for Computing Machinery},
address = {New York, NY, USA},
url = {https://doi.org/10.1145/3411763.3451533},
booktitle = {Extended Abstracts of the 2021 CHI Conference on Human Factors in Computing Systems},
articleno = {198},
numpages = {4}
}

@inproceedings{polymagnets,
author = {},
title = {Polymagnets},
year = {},
isbn = {},
publisher = {},
address = {},
url = {http://www.polymagnet.com/polymagnets/},
doi = {},
booktitle = {},
pages = {},
numpages = {},
location = {},
series = {last accessed July 26, 2022}
}

@inproceedings{nisser2021stochastic,
  title={Stochastic Self-Assembly with Magnetically Re-programmable Voxels},
  author={Nisser, Martin and Makaram, Yashaswini and Mueller, Stefanie},
  booktitle={ACM Symposium on Computational Fabrication, Demonstration},
  year={2021}
}

@inproceedings{yang2023compumat,
  title={CompuMat: A Computational Composite Material for Tangible Interaction},
  author={Yang, Xinyi and Nisser, Martin and Mueller, Stefanie},
  booktitle={Proceedings of the Seventeenth International Conference on Tangible, Embedded, and Embodied Interaction},
  pages={1--5},
  year={2023}
}

@inproceedings{niu2023pullupstructs,
  title={PullupStructs: Digital Fabrication for Folding Structures via Pull-up Nets},
  author={Niu, Lauren and Yang, Xinyi and Nisser, Martin and Mueller, Stefanie},
  booktitle={Proceedings of the Seventeenth International Conference on Tangible, Embedded, and Embodied Interaction},
  pages={1--6},
  year={2023}
}

@inproceedings{bhattacharya2024fabrobotics,
  title={FabRobotics: Fusing 3D Printing with Mobile Robots to Advance Fabrication, Robotics, and Interaction},
  author={Bhattacharya, Ramarko and Lindstrom, Jonathan and Taka, Ahmad and Nisser, Martin and Mueller, Stefanie and Nakagaki, Ken},
  booktitle={Proceedings of the Eighteenth International Conference on Tangible, Embedded, and Embodied Interaction},
  pages={1--13},
  year={2024}
}

@inproceedings{nisser2019sequential,
  title={Sequential support: 3d printing dissolvable support material for Time-Dependent mechanisms},
  author={Nisser, Martin and Zhu, Junyi and Chen, Tianye and Bulovic, Katarina and Punpongsanon, Parinya and Mueller, Stefanie},
  booktitle={Proceedings of the Thirteenth International Conference on Tangible, Embedded, and Embodied Interaction},
  pages={669--676},
  year={2019}
}

@inproceedings{haghighat2022approach,
  title={An approach based on particle swarm optimization for inspection of spacecraft hulls by a swarm of miniaturized robots},
  author={Haghighat, Bahar and Boghaert, Johannes and Minsky-Primus, Zev and Ebert, Julia and Liu, Fanghzheng and Nisser, Martin and Ekblaw, Ariel and Nagpal, Radhika},
  booktitle={International Conference on Swarm Intelligence},
  pages={14--27},
  year={2022},
  organization={Springer}
}

@inproceedings{BBB,
  title={From Prisons to Programming: Fostering Self-Efficacy via Virtual Web Design Curricula in Prisons and Jails},
  author={Nisser, Martin and Gaetz, Marisa and Fishberg, Andrew and Soicher, Raechel and Faruqi Faraz and Long, Joshua},
  booktitle={Proceedings of the 2024 CHI Conference on Human Factors in Computing Systems},
  pages={1--13},
  year={2024}
}

@phdthesis{nisser20193d,
  title={3D printing dissolvable support material for time-dependent mechanisms},
  author={Nisser, Martin Martin Eric William},
  year={2019},
  school={Massachusetts Institute of Technology}
}

@inproceedings{nisser2022demonstration,
  title={Demonstration of Mixels: Fabricating Interfaces using Programmable Magnetic Pixels},
  author={Nisser, Martin and Makaram, Yashaswini and Covarrubias, Lucian and Bah, Amadou Yaye and Faruqi, Faraz and Suzuki, Ryo and Mueller, Stefanie},
  booktitle={Adjunct Proceedings of the 35th Annual ACM Symposium on User Interface Software and Technology},
  pages={1--3},
  year={2022}
}

@book{womack2007machine,
  title={The machine that changed the world: The story of lean production--Toyota's secret weapon in the global car wars that is now revolutionizing world industry},
  author={Womack, James P and Jones, Daniel T and Roos, Daniel},
  year={2007},
  publisher={Simon and Schuster}
}

@inproceedings{werkheiser20143d,
  title={3D printing in Zero-G ISS technology demonstration},
  author={Werkheiser, Mary J and Dunn, Jason and Snyder, Michael P and Edmunson, Jennifer and Cooper, Kennethe and Johnston, Mallory M},
  booktitle={AIAA SPACE 2014 conference and exposition},
  pages={4470},
  year={2014}
}

@inproceedings{faruqi2023style2fab,
  title={Style2Fab: Functionality-Aware Segmentation for Fabricating Personalized 3D Models with Generative AI},
  author={Faruqi, Faraz and Katary, Ahmed and Hasic, Tarik and Abdel-Rahman, Amira and Rahman, Nayeemur and Tejedor, Leandra and Leake, Mackenzie and Hofmann, Megan and Mueller, Stefanie},
  booktitle={Proceedings of the 36th Annual ACM Symposium on User Interface Software and Technology},
  pages={1--13},
  year={2023}
}

@misc{WEF,
  title = {World Economic Forum},
  howpublished = {\url{https://www.weforum.org/impact/carbon-footprint-manufacturing-industry/}},
  note = {Accessed: 2024-04-16}
}

@misc{EPA,
  title = {Environmental Protection Agency},
  howpublished = {\url{https://www.epa.gov/transportation-air-pollution-and-climate-change/carbon-pollution-transportation}},
  note = {Accessed: 2024-04-16}
}

@inproceedings{meenan2008pull,
  title={Pull-up Patterned Polyhedra: Platonic Solids for the Classroom},
  author={Meenan, EB and Thomas, BG},
  booktitle={Bridges Leeuwarden: Mathematics, Music, Art, Architecture, Culture},
  pages={109--116},
  year={2008}
}

@inproceedings{takahashi2011optimized,
  title={Optimized topological surgery for unfolding 3d meshes},
  author={Takahashi, Shigeo and Wu, Hsiang-Yun and Saw, Seow Hui and Lin, Chun-Cheng and Yen, Hsu-Chun},
  booktitle={Computer graphics forum},
  volume={30},
  number={7},
  pages={2077--2086},
  year={2011},
  organization={Wiley Online Library}
}

@misc{Netlib,
  title = {Polyhedron Database},
  year = {2022},
  howpublished = {\url{https://netlib.org/polyhedra/}},
  note = {Accessed: 2022-10-13}
}

@inproceedings{reitebuch2019discrete,
  title={Discrete gyroid surface},
  author={Reitebuch, Ulrich and Skrodzki, Martin and Polthier, Konrad},
  booktitle={Proceedings of bridges 2019: Mathematics, art, music, architecture, education, culture},
  pages={461--464},
  year={2019}
}

@book{demaine2007geometric,
  title={Geometric folding algorithms: linkages, origami, polyhedra},
  author={Demaine, Erik D and O'Rourke, Joseph},
  year={2007},
  publisher={Cambridge university press}
}

@article{demaine2005survey,
  title={A survey of folding and unfolding in computational geometry},
  author={Demaine, Erik D and O’Rourke, Joseph},
  journal={Combinatorial and computational geometry},
  volume={52},
  pages={167--211},
  year={2005},
  publisher={Cambridge University Press Cambridge}
}

@article{schlickenrieder1997nets,
  title={Nets of polyhedra},
  author={Schlickenrieder, Wolfram},
  journal={Master's Thesis, Technische Universit{\"a}t Berlin},
  year={1997},
  publisher={Citeseer}
}

@phdthesis{chen2013edge,
  title={Edge-unfolding almost-flat convex polyhedral terrains},
  author={Chen, Yanping and others},
  year={2013},
  school={Massachusetts Institute of Technology}
}

@article{tolley2014self,
  title={Self-folding origami: shape memory composites activated by uniform heating},
  author={Tolley, Michael T and Felton, Samuel M and Miyashita, Shuhei and Aukes, Daniel and Rus, Daniela and Wood, Robert J},
  journal={Smart Materials and Structures},
  volume={23},
  number={9},
  pages={094006},
  year={2014},
  publisher={IOP Publishing}
}

@article{firouzeh2015robogami,
  title={Robogami: A fully integrated low-profile robotic origami},
  author={Firouzeh, Amir and Paik, Jamie},
  journal={Journal of Mechanisms and Robotics},
  volume={7},
  number={2},
  pages={021009},
  year={2015},
  publisher={American Society of Mechanical Engineers}
}

@article{kilian2017string,
  title={String actuated curved folded surfaces},
  author={Kilian, Martin and Monszpart, Aron and Mitra, Niloy J},
  journal={ACM Transactions on Graphics (TOG)},
  volume={36},
  number={3},
  pages={1--13},
  year={2017},
  publisher={ACM New York, NY, USA}
}

@article{song2004motion,
  title={A motion-planning approach to folding: From paper craft to protein folding},
  author={Song, Guang and Amato, Nancy M},
  journal={IEEE Transactions on Robotics and Automation},
  volume={20},
  number={1},
  pages={60--71},
  year={2004},
  publisher={IEEE}
}

@article{biedl2005can,
  title={When can a net fold to a polyhedron?},
  author={Biedl, Therese and Lubiw, Anna and Sun, Julie},
  journal={Computational Geometry},
  volume={31},
  number={3},
  pages={207--218},
  year={2005},
  publisher={Elsevier}
}

@inproceedings{aronov1991nonoverlap,
  title={Nonoverlap of the star unfolding},
  author={Aronov, Boris and O'rourke, Joseph},
  booktitle={Proceedings of the seventh annual symposium on Computational geometry},
  pages={105--114},
  year={1991}
}

@article{bern1999ununfoldable,
  title={Ununfoldable Polyhedra with Convex Faces},
  author={Bern, Marshall and Demaine, Erik D and Eppstein, David and Kuo, Eric and Mantler, Andrea and Snoeyink, Jack},
  journal={arXiv preprint cs/9908003},
  year={1999}
}

@article{overvelde2016three,
  title={A three-dimensional actuated origami-inspired transformable metamaterial with multiple degrees of freedom},
  author={Overvelde, Johannes TB and De Jong, Twan A and Shevchenko, Yanina and Becerra, Sergio A and Whitesides, George M and Weaver, James C and Hoberman, Chuck and Bertoldi, Katia},
  journal={Nature communications},
  volume={7},
  number={1},
  pages={1--8},
  year={2016},
  publisher={Nature Publishing Group}
}

@article{niiyama2015pouch,
  title={Pouch motors: Printable soft actuators integrated with computational design},
  author={Niiyama, Ryuma and Sun, Xu and Sung, Cynthia and An, Byoungkwon and Rus, Daniela and Kim, Sangbae},
  journal={Soft Robotics},
  volume={2},
  number={2},
  pages={59--70},
  year={2015},
  publisher={Mary Ann Liebert, Inc. 140 Huguenot Street, 3rd Floor New Rochelle, NY 10801 USA}
}

@inproceedings{10.1145/3526113.3545695,
author = {Fang, Chiao and Chan, Vivian Hsinyueh and Cheng, Lung-Pan},
title = {Flaticulation: Laser Cutting Joints with Articulated Angles},
year = {2022},
isbn = {9781450393201},
publisher = {Association for Computing Machinery},
address = {New York, NY, USA},
url = {https://doi.org/10.1145/3526113.3545695},
doi = {10.1145/3526113.3545695},
booktitle = {Proceedings of the 35th Annual ACM Symposium on User Interface Software and Technology},
articleno = {7},
numpages = {16},
location = {Bend, OR, USA},
series = {UIST '22}
}

@inproceedings{hildebrand2012crdbrd,
  title={crdbrd: Shape fabrication by sliding planar slices},
  author={Hildebrand, Kristian and Bickel, Bernd and Alexa, Marc},
  booktitle={Computer Graphics Forum},
  volume={31},
  number={2pt3},
  pages={583--592},
  year={2012},
  organization={Wiley Online Library}
}

@inproceedings{abdullah2022hingecore,
  title={HingeCore: Laser-Cut Foamcore for Fast Assembly},
  author={Abdullah, Muhammad and Sommerfeld, Romeo and Sievers, Bjarne and Geier, Leonard and Noack, Jonas and Ding, Marcus and Thieme, Christoph and Seidel, Laurenz and Fritzsche, Lukas and Langenhan, Erik and others},
  booktitle={Proceedings of the 35th Annual ACM Symposium on User Interface Software and Technology},
  pages={1--13},
  year={2022}
}

@inproceedings{tao2021inforigami,
  title={infOrigami: A Computer-aided Design Method for Introducing Traditional Perforated Boneless Lantern Craft to Everyday Interfaces},
  author={Tao, Ye and Chen, Yu and Fang, Jian and Lin, Jinpeng and Geng, Jingchun and Fang, Ziqi and Chen, Cejun and Yang, Cheng and Zhang, Fan and Sun, Lingyun and others},
  booktitle={The Adjunct Publication of the 34th Annual ACM Symposium on User Interface Software and Technology},
  pages={55--59},
  year={2021}
}

@inproceedings{tahouni2020self,
  title={Self-shaping curved folding: A 4D-printing method for fabrication of self-folding curved crease structures},
  author={Tahouni, Yasaman and Cheng, Tiffany and Wood, Dylan and Sachse, Renate and Thierer, Rebecca and Bischoff, Manfred and Menges, Achim},
  booktitle={Symposium on computational fabrication},
  pages={1--11},
  year={2020}
}

@inproceedings{rivera2017stretching,
  title={Stretching the bounds of 3D printing with embedded textiles},
  author={Rivera, Michael L and Moukperian, Melissa and Ashbrook, Daniel and Mankoff, Jennifer and Hudson, Scott E},
  booktitle={Proceedings of the 2017 CHI conference on human factors in computing systems},
  pages={497--508},
  year={2017}
}
